\documentclass{article} 
\usepackage{iclr2026_conference,times}

\usepackage{amsmath,amsfonts,bm}

\def\eqref#1{equation~\ref{#1}}

\def\1{\bm{1}}

\DeclareMathAlphabet{\mathsfit}{\encodingdefault}{\sfdefault}{m}{sl}
\SetMathAlphabet{\mathsfit}{bold}{\encodingdefault}{\sfdefault}{bx}{n}

\usepackage{hyperref}
\usepackage{url}
\usepackage{times}
\usepackage{graphicx}
\usepackage{mathtools}
\usepackage{amsthm}
\usepackage{float}
\usepackage{algorithmic}
\usepackage{multirow}
\usepackage{amsmath,amssymb,amsfonts}
\usepackage{booktabs}
\usepackage{subcaption}
\usepackage{bm}
\usepackage{tikz}
\usetikzlibrary{shapes.geometric,arrows.meta, positioning, calc, tikzmark,fit,backgrounds}
\usepackage{hf-tikz}
\usepackage{xstring}
\usepackage[framemethod=tikz]{mdframed}
\usepackage[ruled,vlined,linesnumbered]{algorithm2e}

\newcommand{\hlshiftx}{0.3em}  
\newcommand{\hlshifty}{2.2ex}  

\newcommand{\blkstart}[1]{\tikzmark{#1-tl}}%
\newcommand{\blkend}[1]{%
  \hfill\tikzmark{#1-br}%
  \begin{tikzpicture}[remember picture,overlay]
    \begin{scope}[blend mode=multiply]
      \node[
        fill=blue!7,
        inner xsep=3pt,
        inner ysep=1.5pt,
        fit={($(pic cs:#1-tl)+(\hlshiftx,\hlshifty)$)
             ($(pic cs:#1-br)+(\hlshiftx,\hlshifty)$)}
      ] {};
    \end{scope}
  \end{tikzpicture}%
}

\newif\ifneurips
\neuripstrue

\title{Discovering state equivalences in UCT search trees by action pruning}

\author{Robin Schmöcker \\
Institute for Information Processing\\
Leibniz University Hannover\\
Hannover, Germany \\
\texttt{schmoecker@tnt.uni-hannover.de} \\
\And
Alexander Dockhorn \\
SDU Metaverse Lab \\
University of Southern Denmark \\
Odense, Denmark \\
\texttt{adoc@mmmi.sdu.dk} \\
\And
Bodo Rosenhahn \\
Institute for Information Processing\\
Leibniz University Hannover\\
Hannover, Germany \\
\texttt{rosenhahn@tnt.uni-hannover.de} \\
}

\iclrfinalcopy 
\begin{document}

\maketitle

\begin{abstract}
One approach to enhance Monte Carlo Tree Search (MCTS) is
to improve its sample efficiency by grouping/abstracting states or state-action pairs and sharing statistics within a group. Though state-action pair abstractions are mostly easy to find in algorithms such as On the Go Abstractions in Upper Confidence bounds applied to Trees (OGA-UCT), nearly no state abstractions are found in either noisy or large action space settings due to constraining conditions. We provide theoretical and empirical evidence for this claim, and we slightly alleviate this state abstraction problem by proposing a weaker state abstraction condition that trades a minor loss in accuracy for finding many more abstractions. We name this technique Ideal Pruning Abstractions in UCT (IPA-UCT), which outperforms OGA-UCT (and any of its derivatives) across a large range of test domains and iteration budgets as experimentally validated. IPA-UCT uses a different abstraction framework from Abstraction of State-Action Pairs (ASAP) which is the one used by OGA-UCT, which we name IPA. Furthermore, we show that both IPA and ASAP are special cases of a more general framework that we call p-ASAP which itself is a special case of the ASASAP framework.
\end{abstract}

\section{Introduction}
\label{sec:intro}
\noindent Despite the fact that machine learning (ML) methods are state-of-the-art in many decision-making tasks such as playing Go, or Dota 2 as demonstrated by AlphaGo \cite{SilverHMGSDSAPL16} and OpenAI Five \cite{dota2openaifive}, they require a resource intensive training phase, an undesired property for some domains.
For example, Game Studios rarely employ ML-based non-player characters because they would have to be costly retrained every time the game and its rules are significantly altered such as in patches or during the development cycle. 
Therefore, research into non-learned methods such as MCTS, which is state-of-the-art in some applications like MahJong \cite{mahjongIS}, still has merit. 

One research area to improve MCTS is to enhance the Upper Confidence Bounds (UCB) during the tree policy by first grouping states and state-action pairs with similar values and then using the groups' aggregate statistics instead of single-node statistics for UCB to ultimately reduce variance. 
However, one key weakness of state-of-the-art abstraction algorithms such as On the Go Abstractions in Upper Confidence bounds applied to Trees (OGA-UCT) \citep{OGAUCT} is that they struggle to find meaningful state abstractions given a reasonable computational budget even when the environment has a moderate action space size and stochastic branching factor as will be later illustrated in Section \ref{sec:oga_finds_no_abs}. Hence, they are essentially action abstractions for 1-step Markov Decision Processes (MDP) \citep{sutton2018reinforcement} that are applied layerwise.

In this paper, we tackle exactly this problem by proposing a novel algorithm that directly aims at finding correct state abstractions that are not detected by Abstraction of State-Action Pairs in UCT (ASAP-UCT) \citep{AnandGMS15} or OGA-UCT to enable the detection of more Q node abstractions to ultimately boost the performance.
The contributions of this paper can be summarized as follows:

 \textbf{1.} Based on a theoretical justification and an empirical analysis we demonstrate the serious drawback of current SOTA approaches of finding sufficient state abstractions.

\textbf{2.} We propose \textbf{I}deal \textbf{P}runing \textbf{A}bstractions in \textbf{UCT} (IPA-UCT), an OGA-UCT modification that detects more state abstractions, improves the MCTS performance by increasing the sample efficiency as more state abstractions lead to more action abstractions, which improve the sample efficiency. This modification only has a minor runtime overhead (see Tab.~\ref{tab:runtimes}).
    
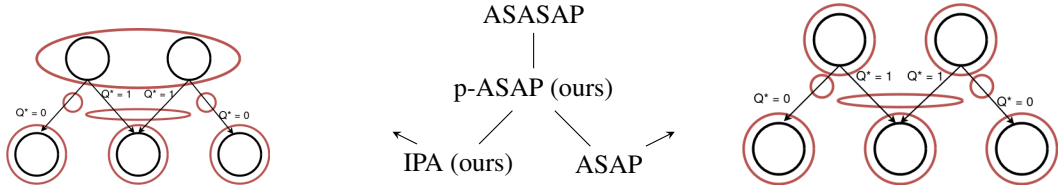
\begin{figure}[ht]
  \centering

  \begin{subfigure}[b]{0.25\textwidth}
    \centering
    \includegraphics[width=\textwidth]{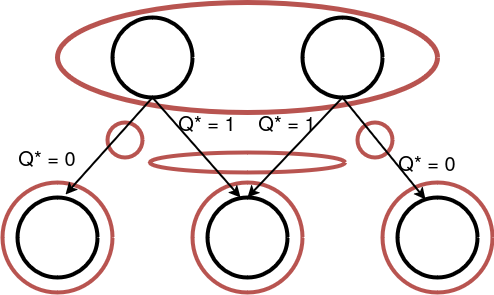}
  \end{subfigure}
  \hfill
  \begin{subfigure}[b]{0.25\textwidth}
    \centering
    \begin{tikzpicture}[>=stealth, node distance=1.5cm]
      \node (A) at (0,3) {ASASAP};
      \node (B) at (0,2) {p-ASAP (ours)};
      \node (C) at (-1,1) {IPA (ours)};
       \node (C2) at (-2,1.5) {};
      \node (D) at (1,1) {ASAP};
      \node (D2) at (2,1.5) {};
      
      \draw (B) -- (A);
      \draw (C) -- (B);
      \draw (D) -- (B);

      \draw[->] (C) -- (C2);
      \draw[->] (D) -- (D2);
    \end{tikzpicture}
  \end{subfigure} \hfill
  \begin{subfigure}[b]{0.3\textwidth}
    \centering
    \includegraphics[width=\textwidth]{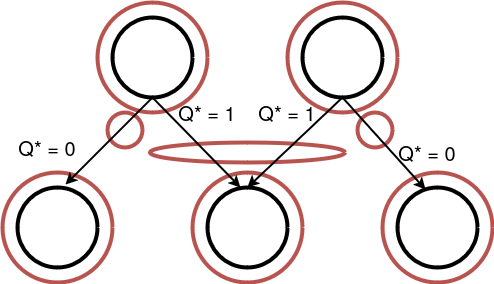}
  \end{subfigure}

  \caption{The hierarchy of abstraction frameworks proposed by us that related Ideal Pruning Abstractions (IPA), Abstraction of State-Action Pairs (ASAP), p(runed)-ASAP, and Alternating State And State-Action Pair Abstractions (ASASAP). The leftmost diagram shows an IPA abstraction on an MDP with 5 states, which are black circles that are connected by deterministic actions that are illustrated with black arrows. The red circles show which actions and states IPA groups/abstracts. The same MDP is shown on the right, but with ASAP abstractions that do not manage to detect the equivalence of the two uppermost states.}
  \label{fig:lattice}
\end{figure}

\textbf{3.} We formulate two new abstraction frameworks, namely, p(runed)-ASAP and \textbf{I}deal \textbf{P}runing \textbf{A}bstractions (IPA) abstractions. Fig.~\ref{fig:lattice} visualizes their hierarchy. In particular, both IPA and ASAP are special cases of p-ASAP and ASASAP \citep{intra}. Though p-ASAP already encompasses both IPA and ASAP, we believe that it further helps understand the core principles behind these abstractions and would help categorize future abstractions.

The paper is structured as follows. Firstly, in \textbf{Section} \ref{sec:foundations}, the theoretical groundwork for this paper is laid. Next, in \textbf{Section} \ref{sec:method}, it is first illustrated why OGA-UCT and ASAP-UCT struggle to find state abstractions, after which in Subsection \ref{sec:vea_abs} we propose our IPA framework and show how to modify OGA-UCT to approximate IPA abstractions in Subsection \ref{sec:vea_uct}. We call this modification IPA-UCT. After having described our methodology, 
the experiment setup is described in \textbf{Section} \ref{sec:experiment_setup}. The experimental results and presented and discussed in \textbf{Section} \ref{sec:experiments} where evaluate and analyze IPA-UCT on various domains to verify its capability to boost performance. At the end, in \textbf{Section} \ref{sec:future_work} we briefly summarise our findings and provide an outlook for future work. 

\section{Foundations of automatic abstractions}
\label{sec:foundations}
\noindent \textbf{Problem model and optimization objective:}
To model sequential decision-making tasks, we use
finite MDPs \citep{sutton2018reinforcement}. In the following, $\Delta(X) \subseteq \mathbb{R}^{|X|}$ denotes the probability simplex of a finite, non-empty set $X$ and $\mathcal{P}(X)$ denotes the power set of $X$.

\textit{Definition:}
    An \textit{MDP} is a 6-tuple $(S,\mu_0,\mathbb{A},\mathbb{P}, R, T)$ where the components are as follows:
    \begin{itemize}
        \item $S \neq \emptyset$ is the finite set of states, $T \subseteq S$ is the (possibly empty) set of terminal states, and  $\mu_0 \in \Delta(S)$ is the probability distribution for the initial state.
        \item $\mathbb{A}\colon S \mapsto A$ maps each state $s$ to the available actions $\emptyset \neq \mathbb{A}(s) \subseteq A$ at state $s$ where $|A| < \infty$.
        \item $\mathbb{P}\colon S \times A \mapsto \Delta(S )$ is the stochastic transition function where we use $\mathbb{P}(s^{\prime} |\: s,a)$ to denote the probability of transitioning from $s \in S$ to $s^{\prime} \in S$ after taking action $a \in \mathbb{A}(s)$ in $s$.
        \item $R \colon S \times A \mapsto \mathbb{R}$ is the reward function.
    \end{itemize}

\noindent  For the remainder of this section, let $M = (S,\mu_0,\mathbb{A},\mathbb{P}, R, T)$ be an MDP and
    \mbox{$P \coloneqq \{(s,a)\: | \: s \in S, a \in \mathbb{A}(s)\}$}
be the set of all state-action pairs. The optimization objective is to find an agent $\pi$ (formally, a mapping $\pi \colon S \mapsto \Delta(A)$)
such that $\pi$ maximizes the expected episode's return where the (discounted) return for of episode $s_0,a_0,r_0, \dots, s_n,a_n,r_n,s_{n+1}$ with $s_{n+1} \in T$ is given by $\gamma^0 r_0 + \ldots + \gamma^n r_n$.

\noindent \textbf{Abstraction frameworks:}
First, the main abstraction framework that IPA will build upon is defined. \cite{AnandGMS15} introduced the so-called Abstractions of State-Action Pairs (ASAP) framework which provides a method to detect value equivalent states and state-action pairs in an MDP.
An example of an ASAP abstraction on a small state graph can be seen in Fig.~\ref{fig:lattice}. The core idea of ASAP is to alternatingly construct a state abstraction (which is simply an equivalence relation over the state space) given a state-action pair abstraction and vice versa.

\textit{ASAP: From state-action pair abstraction to state abstraction:} 
Let $\mathcal{H}^{\prime}_{\text{ASAP}} \subseteq P \times P$ be a state-action pair abstraction, i.e. an equivalence relation over $P$. The corresponding ASAP state abstraction $\mathcal{E} \subseteq S \times S$ is then given by the following:
\begin{equation}
    \begin{aligned}
   & (s_1,s_2) \in \mathcal{E}_{\text{ASAP}}  \iff \\
        &\forall a_1 \in \mathbb{A}(s_1) \, \exists a_2 \in \mathbb{A}(s_2):  
        ((s_1,a_1),(s_2,a_2)) \in \mathcal{H}^{\prime}_{\text{ASAP}} \\
        &\forall a_2 \in \mathbb{A}(s_2) \, \exists a_1 \in \mathbb{A}(s_1):  
        ((s_1,a_1),(s_2,a_2)) \in \mathcal{H}^{\prime}_{\text{ASAP}}.
    \end{aligned}
\end{equation}
In contrast the predecessors of ASAP, like the work by \cite{uctJiang}, two states can be abstracted even if their sets of legal actions differ. However, there is still room to relax this state condition, for example, by having requiring only a subset of the legal action space to have a match. This is the key idea behind IPA which we will be described in Section~\ref{sec:method}.

\textit{ASAP: From state abstraction to state-action pair abstraction}
The second component of ASAP defines how one obtains a state-action pair abstraction $\mathcal{H}_{\text{ASAP}}$ given a state abstraction $\mathcal{E}^{\prime}_{\text{ASAP}}$. Concretely,
any state-action-pair $(s_1,a_1),(s_2,a_2) \in P$ is equivalent i.e. $((s_1,a_1),(s_2,a_2)) \in \mathcal{H}_{\text{ASAP}}$ if and only if the state-action pairs have identical immediate rewards and transition distributions:
\begin{equation}
    \begin{aligned}
         \quad | R(s_1,a_1) - R(s_2,a_2) | 
        & \leq \varepsilon_{\text{a}} \\
        \quad \textrm{and } F \coloneqq \sum \limits_{x \in \mathcal{X}} \bigg| \sum \limits_{s^{\prime} \in x} 
        \mathbb{P}(s^{\prime}|\: s_1,a_1) - \mathbb{P}(s^{\prime}|\: s_2,a_2) \bigg| 
       &  \leq \varepsilon_{\text{t}},
    \end{aligned}
\end{equation}
where $\mathcal{X}$ are the equivalence classes of $\mathcal{E}^{\prime}_{\text{ASAP}}$ and $\varepsilon_{\text{t}} = \varepsilon_{\text{a}} = 0$. 

If one starts with the state abstraction that groups all terminal nodes and repeats these two construction steps until convergence, the final thus-obtained abstraction is called the ASAP abstraction of the given MDP. ASAP abstractions are a special case of ASASAP abstractions \citep{intra} which formalize the working principle of repeatedly constructing a state abstraction from a state-action pair abstraction and vice versa.

\noindent \textbf{Building and using abstractions to enhance search:}
Since, the state and state-action pair spaces can be arbitrarily large, constructing an ASAP on the MDP $M$ is mostly infeasible.
Therefore, ASAP-based abstraction algorithm such as 
ASAP-UCT \citep{AnandGMS15}, OGA-UCT \citep{OGAUCT}, and IPA-UCT (see Section \ref{sec:vea_uct}) construct their abstraction
on the local-layered MDP rooted at the state $s_{\text{d}}$ where the decision has to be made. The local-layered MDP is given as the current MCTS search graph (the MCTS version used here is specified in Section~\ref{sec:mcts}).
  In this simplified setting, one can use dynamic programming to compute the ASAP abstraction since one only needs to have access to the abstraction at layer/depth $i+1$ of the search graph to compute those at depth $i$.
While ASAP-UCT pauses MCTS in regular intervals to construct the ASAP abstraction, the current state of the art that follows this working principle, is OGA-UCT \cite{OGAUCT} which keeps a recency counter for each state-action pair and once its passed, updates tests if its abstraction changes, and if so propagates that change to its parent which might in turn also update their abstraction. This allows one to continuously have access to an up-to-date ASAP abstraction without an enormous runtime overhead.

\textbf{OGA-UCT for multi-agent settings:} In the experiments, we will also evaluate OGA-based algorithms on board games, which are not MDPs as they feature two players. The only modification needed for OGA is to optimize and keep track of the Q values for the player at the turn at the corresponding node.

\textbf{OGA-UCT extensions to high stochasticity settings:} In practice, OGA-UCT is slightly modified to better handle high stochasticity. Two modifications exist with which IPA-UCT will be tested.
Firstly, pruned OGA is identical to OGA-UCT except that 
when building the state-action pair abstractions certain successors are ignored. Concretely, for a state-action pair with $n$ successors with respective probabilities $p_1,\dots,p_n$ those with $p_i < \alpha \cdot \max \{p_1,\dots,p_n\},\ \alpha \in [0,1]$ are ignored. 
Furthermore, there is $(\varepsilon_{\text{a}},\varepsilon_{\text{t}})$-OGA \citep{ogacad} which allows $\varepsilon_{\text{a}},\varepsilon_{\text{t}}$ values greater than zero. \cite{ogacad} describes the implementation details as larger than zero values do not necessarily induce an equivalence relation, i.e. a non-overlapping partition.

\textbf{RSTATE-OGA}: Later, when the different OGA variants are experimentally investigated, one ablation that will also be conducted is to test the performance of random state abstractions to ensure that any performance gains due to the usage of abstractions are better than if random abstractions were used. OGA-UCT that uses random state abstractions is called RSTATE-OGA and functions as follows. Whenever a state node $\mathcal{S}$ is visited for the $K$-th time and its current abstract node consists only of itself, then with the probability $p_{\text{abs}} \in [0,1]$, $\mathcal{S}$'s abstract node is changed with uniform probability to any of the abstract nodes of the same depth. Initially, at creation, any Q node is its own abstract node.

\textbf{Abstraction usage:} A state-action pair abstraction can also be viewed as a partition over $P$. During the tree policy, instead of using the statistics (i.e. cumulative returns and visits) of a single node, the aggregate statistics of the node's group are used to ultimately reduce variance. This is the key abstraction usage mechanism that all methods that are considered in this paper use, e.g. AS-UCT \citep{uctJiang} (predecessor of ASAP-UCT), ASAP-UCT, OGA-UCT, $(\varepsilon_{\text{a}},\varepsilon_{\text{t}})$-OGA, pruned OGA, and IPA-UCT.

\noindent \textbf{Other automatic abstraction algorithms:}
The literature on abstraction algorithms includes abstractions for Sparse Sampling trees \citep{HostetlerFD15}, abstractions of the transition function, \citep{SokotaHAK21,YoonFGK08,YoonFG07,saisubramanian2017optimizing}, purely statistical-based abstractions \citep{aupo} which do not require an equality check operator like OGA, or abstractions that deliberately group states or state-action pairs with differing $V^*$ or $Q^*$ values \citep{kvda}. Another area of research is the abstraction usage, which might involve dynamically abandoning the abstraction \citep{EMCTSXu,ogacad} or defining an intra-abstraction policy \cite{intra}.
Abstractions have also been investigated for other problem settings, such as continuous and/or partially observable problems \citep{HoergerKKY24}, learning-based methods, such as learning and planning on abstract models \citep{OzairLRAOV21,KwakHKLZ24,ChitnisSKKL20}.
An in-depth overview of the non-learning-based abstraction literature has been created by \cite{SchmockerD25}.

\section{Method}
\label{sec:method}
In this Section, we will be introducing our novel IPA-UCT algorithm by first showing that ASAP struggles with finding state abstractions. Then we will be introducing a new abstraction framework called IPA, which IPA-UCT tries to approximate. Lastly, we illustrate on a concrete example how the IPA framework detects state equivalences that ASAP does not.

\noindent \textbf{Why ASAP finds (nearly) no state abstractions:}
\label{sec:oga_finds_no_abs}
In the following, we will first give theoretical arguments why ASAP finds few state abstractions, only after which we show experimental evidence to support these claims.

\noindent \textit{Theory}: Let us consider a simplified model where two states $s_1,s_2$ with $n$ and $l$ actions respectively are given. Furthermore, assume that each of $s_1$'s and $s_2$'s actions are assigned to an abstract Q node from a pool of $m$ abstract Q nodes with uniform probability. Using elementary combinatorial arguments, the probability $p_{\text{abs}}$ of $s_1$ and $s_2$ being abstracted according to the ASAP framework can be exactly denoted and then upper bounded by
\begin{equation}
    p_{\text{abs}} = \frac{\sum\limits_{k=1}^{c \coloneqq \min \{n,l,m\}}
    \binom{m}{k} f(n,k) f(l,k)
    }{m^{n+l}} \leq \left(\frac{2c}{m}\right)^{n+l}
\end{equation}
where $f(n,k)$ is the number of surjections from a set of $n$ elements to a set of $k \leq n$ elements (proof is provided in the supplementary materials Section \ref{sec:proof}). This shows that once there is a critical amount of possible abstractions $m$, then the probability decays at least exponentially in the number of actions $n$ and $l$. The method IPA that we propose won't depend on $m$.

\noindent \textit{Empirical results}: Aside from these theoretical arguments, we empirically measured the abstraction rate for OGA-UCT. The measurements can be seen in the supplementary materials in Tab.\ref{tab:abs_rates} in the OGA column with $\varepsilon_{\text{a}} = \varepsilon_{\text{t}} = 0$. Clearly, with a few exceptions, nearly no state abstractions are built despite the fact that at least value-equivalent states have to exist in some environments due to symmetry reasons such as Game of Life and SysAdmin. There are two notable exceptions, namely Crossing Traffic and Skills Teaching where standard OGA detects a notable number of state abstractions. However, these are arguably trivial: In Skills Teaching, to simulate the student's learning process, every other turn has only a single action, hence these state abstractions are essentially action abstractions. In Crossing Traffic, once the agent has been hit by an obstacle, the game reaches a non-terminal states in which all the agent's actions have no effect. These trivial dead states are detected. 

\noindent \textbf{The p-ASAP and IPA abstraction frameworks:}
\label{sec:vea_abs}
First, we introduce an abstraction framework that we call p(runed)-ASAP of which ASAP is a special case. Given a state-action-pair abstraction $\mathcal{H}$, some action pruning function \mbox{$J \colon S \mapsto \mathcal{P}(A)$} such that \mbox{$J(s) \subseteq \mathbb{A}(s)$} for all $s$, we can define a symmetric and reflexive (but not necessarily transitive) relation $\sim_J$ with $s_1 \sim_J s_2$ if and only if
\begin{align}
    \forall a_1 \in J(s_1)\ \exists a_2 \in \mathbb{A}(s_2) :\quad & ((s_1,a_1),(s_2,a_2)) \in \mathcal{H},
    \label{eq:vea1}
    \\
    \forall a_2 \in J(s_2)\ \exists a_1 \in \mathbb{A}(s_1) :\quad & ((s_1,a_1),(s_2,a_2)) \in \mathcal{H}.
    \label{eq:vea2}
\end{align}
If $\sim_J$ is an equivalence relation, then we call the abstraction using $f(\mathcal{H}) = \sim_J$ a p-ASAP abstraction.

ASAP is obtained from p-ASAP by using \mbox{$J_{\text{ASAP}}(s) = \mathbb{A}(s)$} for all $s$. As illustrated in Section \ref{sec:oga_finds_no_abs}, the ASAP framework is practically unable to detect any state abstractions, hence, the goal is to find a $J$ that does as much pruning as possible whilst keeping value-invariance of the resulting abstraction. This is already guaranteed by ASAP because when all actions have an equivalent match, then it is also the case for the optimal action which determines the $V^*$ value.

If one chooses \mbox{$J^*(s)$} to be the set of optimal actions (i.e., those with the maximal $Q^*$ value), then the maximal number of action pruning is performed whilst ensuring value equivalence of any two abstracted states. Note that $\sim_{J^*}$ is an equivalence relation. We refer to the $p$-ASAP abstraction using $J^*$ as \textbf{I}deal-\textbf{P}runing-\textbf{A}bstractions (IPA) as only those states are grouped that have the same value under optimal play. However, this framework still does not capture all value equivalent states, as two states may be value equivalent but have no ASAP-equivalent actions.

\noindent \textbf{IPA versus ASAP on Navigation:}
\label{subsec:ipa_on_navigation}
In this section, we will demonstrate a motivating example in which the ASAP framework is unable to detect some state abstractions that are encompassed by the IPA framework.
Consider the Navigation instance that is illustrated in Fig.~\ref{fig:nav_example} and whose definition is given in the supplementary materials \ref{sec:problem_descriptions}. Counterintuitively, the optimal policy is to continuously attempt the straight path $3 \to 8 \to 13 \to 18$ which yields an average return of $-3$. Going around cell $8$, either left or right, has a lower average return of $-5$. To decrease the chance of MCTS to take one of these suboptimal paths, it would be of benefit if states 2 and 4 are abstracted as that would imply the actions $3 \to 2$ and $3 \to 4$ could be abstracted too, thus allowing MCTS to average their Q values and therefore decrease this suboptimal-path probability.

However, according to the ASAP framework, states 12 and 14 cannot be abstracted, since the action 14 $\to $ 15, does not have an ASAP-equivalent action in state 12 (this can be checked using that ASAP-equivalent actions must also be value-equivalent). Consequently, states 7,9 and ultimately 2,4 won't be abstracted. On the contrary, the IPA framework does find all of these abstractions, as going from 12 to 13 or from 14 to 13 are the unique optimal actions which are abstracted since they result in the same ground state.

\begin{figure}[htbp]
  \centering
  \begin{tikzpicture}[scale=0.65]
    \fill[gray] (0,0) rectangle (1,1);
    \fill[gray] (4,0) rectangle (5,1);
    
    \fill[gray] (0,3) rectangle (1,4);
    \fill[gray] (1,3) rectangle (2,4);
    \fill[gray] (3,3) rectangle (4,4);
    \fill[gray] (4,3) rectangle (5,4);
    
    \fill[gray] (0,1) rectangle (1,2);
    \fill[gray] (4,1) rectangle (5,2);
    \fill[gray] (4,2) rectangle (5,3);
    
    \fill[gray] (2,1) rectangle (3,2);
    
    \draw (2.5,0.5) circle (0.3cm);
    
    \node at (2.5,3.5) {\Large G};
    
    \foreach \y in {0,...,3} { 
      \foreach \x in {0,...,4} { 
        \pgfmathtruncatemacro{\num}{\x + 1 + 5*\y}
        \node at (\x+0.8,\y+0.8) {\footnotesize \num};
      }
    }
    
    \draw[step=1cm, black] (0,0) grid (5,4);
    \draw[thick] (0,0) rectangle (5,4);
  \end{tikzpicture}
  \caption{A 5$\times$4 Navigation instance to illustrate an example where the IPA framework (our method) detects the value equivalencies of states 2,4 and 7,9 and 12,14 which cannot be done with ASAP. The circle indicates the initial position, G indicates the goal cell, white cells have a reset probability of $0$, and black cells have a reset probability of $0.5$.}
  \label{fig:nav_example}
\end{figure}
\noindent \textbf{IPA-UCT:}
\label{sec:vea_uct}
Next, we will discuss how the IPA abstraction framework can be integrated with OGA-based methods, which we then call \textbf{IPA-UCT}. The full pseudocode for this modification is provided in the supplementary materials in Pseudocode~\ref{alg:ipa:pseudocode}, which highlights the differences to $(\varepsilon_{\text{a}},\varepsilon_{\text{t}})$-OGA in blue.
IPA-UCT will only modify the state abstraction component of OGA; hence, we regard either using the mechanism of pruned OGA or $(\varepsilon_{\text{a}},\varepsilon_{\text{t}})$-OGA for the state-action pair abstraction simply as a parametrization of IPA-UCT. We proceed as follows. Firstly, we will introduce an approximation for $J^*$. Since this approximation $J_{\text{UCB}}$ does not induce an equivalence relation, we will show how we can transform it into one such that it can be incorporated into OGA in the supplementary materials in Section \ref{sec:vea_jhat_to_equiv}. Our idea is to approximate $J^*(s)$ using current search tree information. The approximation $J_{\text{UCB}}$ for $J^*$  for a state $s$ is 
\begin{equation}
    J_{\text{UCB}}(s) = \{a \in \mathbb{A}(s)\: | \: \text{UCB}(a) \geq Q_{\text{max}}\}
\end{equation}
where $Q_{\text{max}} = \max\limits_{a \in \mathbb{A}(s)} Q(s,a)$ is the maximum Q value statistic at node $s$ and UCB$(a)$ denotes the current UCB value of action $a$ in state $s$. The idea is that, mostly, one cannot tell what the optimal action is; however, given enough visits, one can oftentimes exclude some actions which are almost certainly not optimal. We parametrize this technique by some $\lambda_{\text{p}} \in \mathbb{R} \cup \{ \mathbb{\infty} \}$ which is used as the exploration constant for the UCB values that are used for this pruning procedure.
If one chooses $\lambda_{\text{p}} = 0$, then only those actions are kept that have the current maximum Q value. If one selects $\lambda_{\text{p}} = \infty$, then no pruning takes place, hence $J_{\text{UCB}} = J_{\text{ASAP}}$. Hence, $\lambda_{\text{p}}$ controls the riskiness when building the state abstractions. As will be later shown, the best performances are reached with non-trivial $\lambda_{\text{p}}$ values that have an optimal tradeoff between finding additional correct state abstractions that are built at the cost of faulty new ones.
Since the state abstractions now do no longer exclusively depend on the abstractions of the Q nodes, we introduce a recency counter for state nodes as well. The recency count's threshold is set for simplicity to the same value that is used for the Q nodes. As with the Q nodes, whenever that recency counter reaches the threshold, we update the state abstraction along with the current value for $J_{\text{UCB}}$. 

Though heuristical in nature, IPA-UCT has the same soundness guarantee as OGA-UCT as specified in the following theorem, which is proven in the supplementary materials in Section \ref{sec:proof1}.

\textit{Soundness theorem:} The abstraction on IPA-UCT's search tree will become sound (i.e. group only states with the same $V^*$ value and state-action pairs with the same $Q^*$ value) almost surely in the iteration limit  when using OGA-UCT as the state-action pair abstraction mechanism.

\section{Experiment setup}
\label{sec:experiment_setup}
In this section, we describe the general experiment setup. Any deviations from this setup will be explicitly mentioned.

\textbf{Parameters:}
Originally, OGA \citep{OGAUCT} used the absolute value of the abstract Q value as the exploration constant. However, this technique has been improved by the dynamic, scale-independent exploration factor Global-Std \citep{demcts}. The Global-Std exploration constant 
has the form $C \cdot \sigma$ where $\sigma$ is the standard deviation of the Q values of all nodes in the search tree and $C \in \mathbb{R}^+$ is some fixed parameter. Furthermore, we always use $K=3$ as the recency counter, which was proposed by Anand et al. \citep{OGAUCT}.

\textbf{Problem models:}
For this paper, we ran our experiments on a variety of MDPs, all of which are either from the International Probabilistic Planning Conference \citep{grzes2014ippc}, are well-known board games, or are commonly used in the abstraction algorithm literature \citep{AnandGMS15,OGAUCT,HostetlerFD15,YoonFGK08,uctJiang}.
All experiments were run on the finite-horizon versions of the considered MDPs with a default horizon of 50 steps and 100 for the board games with a planning horizon of 50 and a discount factor $\gamma=1$. The board games are zero-sum evaluated by inserting standard MCTS with 500 iterations as the opponent. 
If the reader is not familiar with any of the domains that were used for the experiments, a brief description for each MDP is provided in the supplementary materials in Section~\ref{sec:problem_descriptions}.

\textbf{Evaluation:}
Each data point that we denote in the remaining sections of this paper (e.g. agent returns) is the average of at least 2000 runs. Whenever we denote a confidence interval for a data point, then this is always a confidence interval with a confidence level of 99\% provided by $\approx 2.33$ times the standard error. Furthermore, we use a borda-like ranking system to quantify agents' performances; in particular, we use \textit{pairings} and \textit{relative improvement scores}. For details, see supplementary Section \ref{subsec:scors_defs}.

\textbf{Reproducibility:}
For reproducibility, we released our implementation \citep{repo}.  Our code was compiled with g++ version 13.1.0 using the -O3 flag (i.e. aggressive optimization). 

\section{Experiments}
\label{sec:experiments}
First, we compare the overall performances of IPA-UCT, pruned OGA, and $(\varepsilon_{\text{a}},\varepsilon_{\text{t}})$-OGA, and RSTATE-OGA (to ensure any performance gains come from non-trivial sources) by computing their pairings and relative improvement for different iteration budgets obtained from the performance values of all $> 20$ considered environments. The parameters we varied are the following: For all methods, we used $C \in \{0.5,1,2,4,8,16\}$
 For $(\varepsilon_{\text{a}},\varepsilon_{\text{t}})$-OGA, we tested $\varepsilon_{\text{a}} \in \{0,\infty\}$, $\varepsilon_{\text{t}} \in \{0,0.2,0.4,0.8\}$, for pruned OGA we used $\alpha \in \{0,0.1,0.2,0.5,0.75,1.0\}$, and for RSTATE-OGA we used $p_{\text{abs}} \in \{0.1,0.2,0.5,1.0\}$. We varied the pruning constant $\lambda_{\text{p}} \in \{0,0.25,0.5,1,2,4,\infty\}$ where $\lambda_{\text{p}} = \infty$ corresponds to doing no pruning at all i.e. defaulting to standard pruned OGA or $(\varepsilon_{\text{a}},\varepsilon_{\text{t}})$-OGA.
 Each parameter combination was run with 100, 200, 500, and 1000 iterations. 

\begin{figure}[H]
    \centering
    \begin{minipage}[t]{0.49\linewidth}
        \centering
        \includegraphics[width=\linewidth]{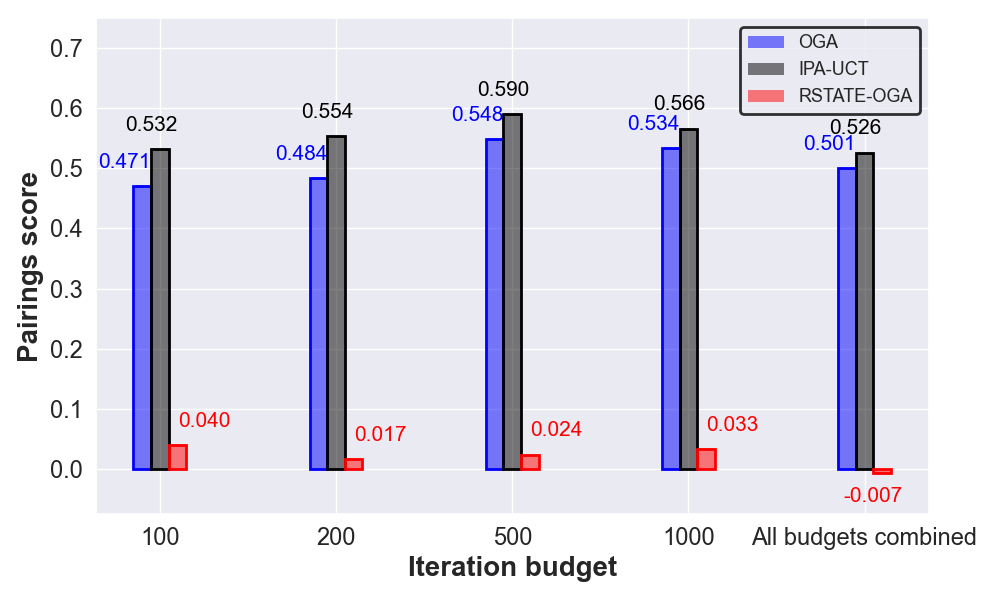}
        \caption{The pairings scores of the best IPA-UCT parameter combination compared to the best parameter combination of RSTATE-OGA, and both pruned OGA and $(\varepsilon_{\text{a}},\varepsilon_{\text{t}})$-OGA (summarized as OGA). The overall generalization performance of IPA-UCT for all iteration settings was achieved using pruned OGA with $\lambda_{\text{p}} = 2$, $C=2$, and $\alpha=0.75$.}
        \label{fig:ipa:pairings}
    \end{minipage}
    \hfill
    \begin{minipage}[t]{0.49\linewidth}
        \centering
        \includegraphics[width=\linewidth]{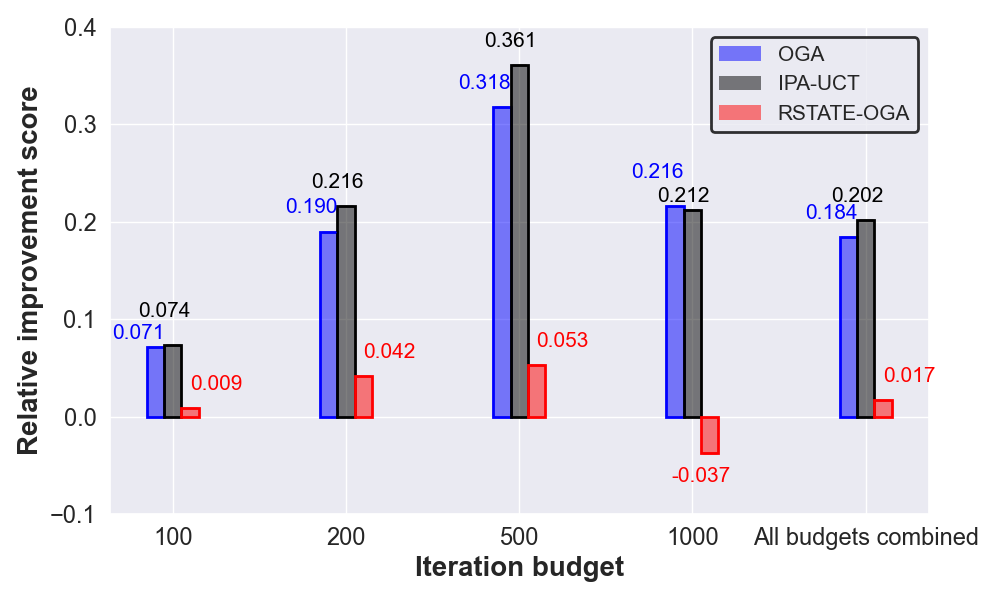}
        \caption{The relative improvement scores of the best IPA-UCT parameter combination compared to the best parameter combination of RSTATE-OGA, and both pruned OGA and $(\varepsilon_{\text{a}},\varepsilon_{\text{t}})$-OGA (summarized as OGA). The overall generalization performance of IPA-UCT for all iteration settings was achieved using $(0,0.2)$-OGA with $\lambda_{\text{p}} = 1$ and $C=1$.}
        \label{fig:ipa:improvs}
    \end{minipage}
\end{figure}

Bar charts ~\ref{fig:ipa:pairings} compare the pairings and relative improvement scores for the best parameter-combinations for each iteration budget. This shows that using IPA-UCT clearly has better generalization capabilities than not using the modified state abstraction that IPA-UCT introduces, with a sweet spot in performance being the 500 iterations setting where an average 5\% performance increase over the best OGA parameter combination can be found. Except for the relative improvement score in the 1000 iterations setting, IPA-UCT attained higher scores in all iteration budgets than standard OGA-based methods.
IPA-UCT can however, only gain a clear advantage over OGA-based techniques in this generalization setting as the per-environment parameter-optimized yield only minor (if any) improvements except for Cooperative Recon, which we visualized and discuss in the supplementary materials Section \ref{subsec:optimized}. Nonetheless, these results show that IPA-UCT can be a valuable drop-in improvement for OGA-based algorithms, offering a clear advantage when one cannot afford to fine-tune parameters per task. In the next section, we discuss how $\lambda_{\text{p}}$ can be chosen.

\begin{figure}[H]
\centering

\begin{minipage}{0.4\textwidth}
\centering
\includegraphics[width=\linewidth]{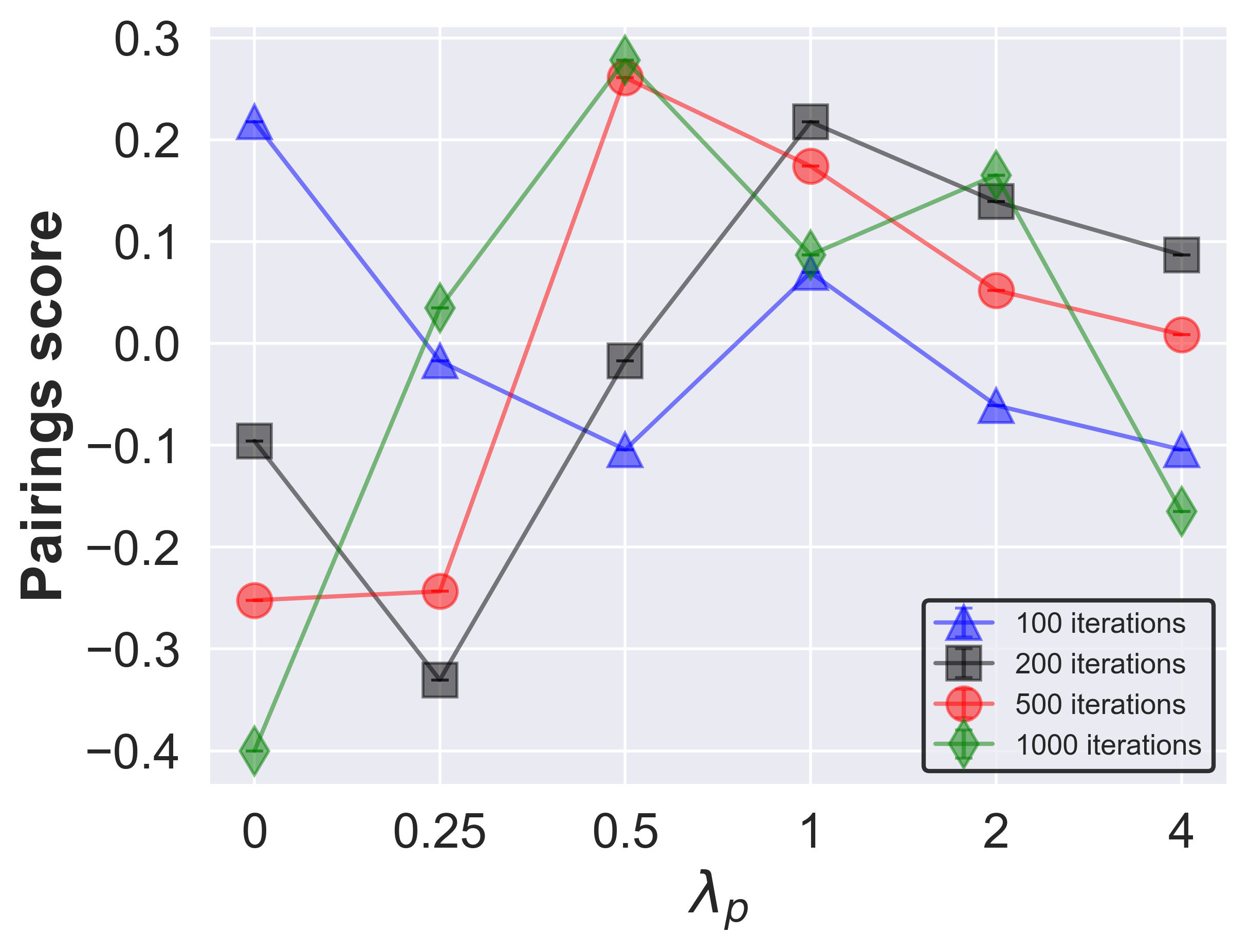}
\caption*{Pairings scores}
\end{minipage}
\hfill
\begin{minipage}{0.4\textwidth}
\centering
\includegraphics[width=\linewidth]{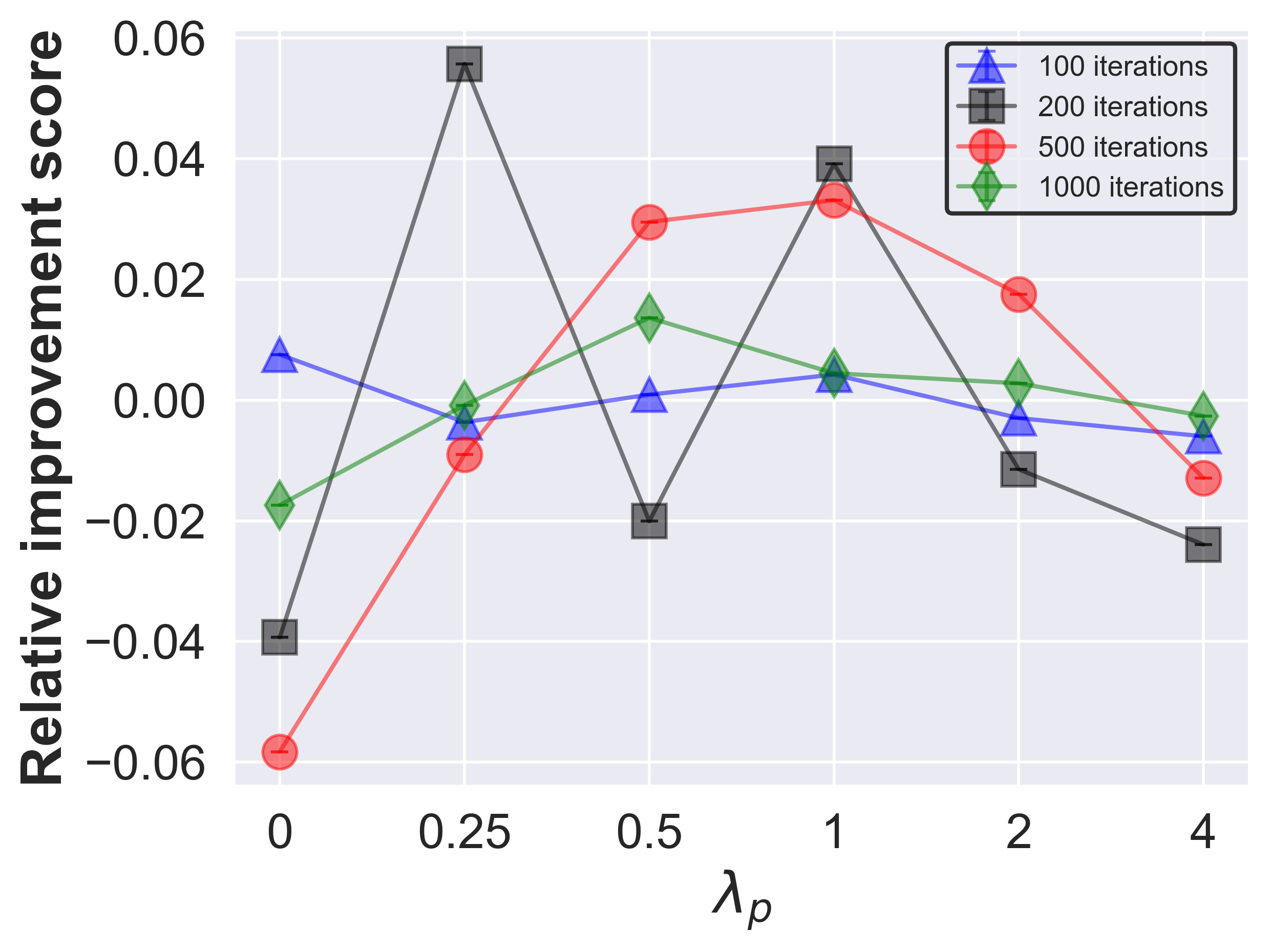}
\caption*{Relative improvement scores}
\end{minipage}

\caption{The pairings and relative improvement scores of different $\lambda_{\text{p}}$ values (when only paired against each other for a fixed iteration budget, i.e. all curves are independent of each other) for different iteration budgets. Though the data points in between $\lambda_{\text{p}}$ weren't measured, we still drew connecting lines for the reader to better differentiate between the course of the scores for the different budgets.}
\label{fig:ipa:ablation_filter}
\end{figure}

\noindent \textbf{Ablation: Performance as a function of} $\boldsymbol{\lambda_{\text{p}}}$\textbf{:}
Next, we investigate the relative performance between the $\lambda_{\text{p}}$ values. Fig.~\ref{fig:ipa:ablation_filter} shows the performance curve when varying $\lambda_{\text{p}}$ for the here-considered iteration budgets in terms of the pairings and relative improvement score for all environments (the performance graphs for each individual environment can be found in the supplementary materials in Fig.~\ref{fig:ipa:optimized_filter_mp}. The following observations can be made:

 \noindent \textbf{1)} First and foremost, all curves feature a clear downwards trend as $\lambda_{\text{p}}$ approaches the largest here-considered value $4$. All curves have a peak at less than $4$. This further validates the positive impact that the UCB-based pruning has on the performance, as the higher the $\lambda_{\text{p}}$ value, the closer IPA-UCT is to standard OGA.

 \noindent \textbf{2)}  For both scores, the curves for 500 and 1000 iterations have a single peak, which is either $\lambda_{\text{p}} = 0.5$ or $\lambda_{\text{p}} = 1$, depending on the score type.

 \noindent \textbf{3)}  Surprisingly, even for small iteration counts of 100 or 200 iterations, there are still clear peaks which are at either $\lambda_{\text{p}} = 0$ or $\lambda_{\text{p}} = 0.25$ (except for the 200 iterations pairings score). This makes sense as in lower iteration budgets, IPA-UCT needs to be more risk-taking in the abstraction building as there aren't enough visits to have confidence in the pruning, hence the $\lambda_{\text{p}}$ peaks are at lower values than for the higher iteration budgets.

\noindent \textbf{Alternative pruning methods:}
Though we found most success with the proposed $J_{\text{UCB}}$ function of pruning actions that relies on the UCB and Q values only, we also conducted preliminary experiments on two different approaches that are described in the following, and whose performances are shown in Tab.~\ref{tab:ipa:alternatives}. However, both approaches performed worse than $J_{\text{UCB}}$, whose downsides we will briefly cover, and why we ultimately presented IPA-UCT instead of these.

 \noindent \textit{Confidence-based pruning}: For this method, we kept track of a confidence interval with confidence level $p_{\text{c}}\in[0,1]$ for each Q value. We then used $J_{\text{conf}}$ which prunes all actions at a state $s$ whose upper confidence bound are lower than the highest lower bound. 
     We call this method \textit{CONF-UCT}.
    In our observation, the Q values were much too noisy to perform any meaningful pruning.
    
 \noindent \textit{Hard pruning}: For this method $J_{\text{top}}$, one only keeps the best $n_{\text{matches}}$ actions when ordered by their current Q value. To avoid the risk of faulty prunings, $J_{\text{top}} =J_{\text{ASAP}}$ if the node has less than $n_{\text{min}}$ visits. This method, which we name \textit{TOPN-UCT}, performed nearly on PAR with $J_{\text{UCB}}$, however, it has two parameters to configure rather than just one. 

\section{Conclusion and Future Work}
\label{sec:future_work}
In this paper, we first defined the IPA abstraction framework and generalized best IPA and ASAP 
by introducing p-ASAP which itself is a special case of ASASAP abstractions \citep{intra}. Next, we showed both empirically and theoretically that OGA-UCT effectively finds no state abstractions. We proposed IPA-UCT to alleviate this issue and showed that IPA-UCT yields a consistent performance improvement over OGA-based methods.

One limitation of IPA-UCT is that there is no single $\lambda_{\text{p}}$ value that performs well for all environment. One avenue for future work is to automatically detect the correct value for $\lambda_{\text{p}}$. While for some environments $\lambda_{\text{p}}=0$ is best, this can be harmful to others. Furthermore, some environments prefer neither $\lambda_{\text{p}}= \infty$ (i.e. no pruning) nor $\lambda_{\text{p}}=0$ but rather some value in between.
Also, IPA-UCT is clearly not optimal in that still many state abstractions, especially those arising due to symmetry (e.g., in Game of Life or SysAdmin), are not detected because as OGA-UCT, IPA-UCT relies on the detection of action abstractions. We believe that this near-exact abstraction that is being built in IPA-UCT and OGA-UCT is not the path forward to resolve this issue, as it requires too many Q nodes to have sampled nearly all their possible outcomes which is mostly infeasible when the stochasticity has more than binary outcomes. A new automatic abstraction paradigm is required.
Lastly, since IPA-UCT is a modification of OGA-UCT is suffers from the same limitation in that a directed acyclic search graph is required for any abstractions to be detected because if no two state-action-pairs result in the same state, then no action abstraction can be built.

\newpage 

\bibliography{references}
\bibliographystyle{iclr2026_conference}

\appendix
\section{Supplementary Materials}
\label{sec:appendix}
\subsection{Proof of soundness theorem}
\label{sec:proof1}
In IPA-UCT using standard OGA-UCT, every state-action pair of the MDP will be expanded, and its visits converge in probability to $\infty$ due to UCB's exploration term. Hence, in the limit, every state-action pair successor will have also been sampled almost surely. Next, assuming that this is the case, one shows inductively, starting from the bottom layer, that the built abstraction will become sound almost surely.

\textbf{Induction start}: The fully expanded search tree's bottom layer contains only terminal states, which are grouped by default. This abstraction is sound as terminal states have a $Q^*$ value of 0.

\textbf{Induction step}: Assume that the state abstraction at layer $L+1$ becomes sound almost surely. Consequently, the state-action pairs' Q-values in layer $L$ will converge in probability to their $Q^*$ values as UCB is used as the tree-policy. Also, independent of the state-action pair's Q-values, their abstractions almost surely become sound as they are built with the standard ASAP rules. The $V^*$ value of any state is defined as the maximum $Q^*$ value of its actions. Since the Q-values of the state-action pairs in layer $L$ converge in probability to their $Q^*$ values, the Q value of any optimal action $a^*$ for any state $s$ at layer $L$ and therefore its UCB value will almost surely be greater than the Q value of any suboptimal action of $s$. Therefore, all the optimal actions of $s$ will almost surely never be pruned simultaneously. Hence, the state abstraction at layer $L$ will also almost surely become sound. \qed

\subsection{Parameter-optimized performances}
\label{subsec:optimized}
\begin{figure}[H]
\centering

\begin{minipage}{0.3\textwidth}
\centering
\includegraphics[width=\linewidth]{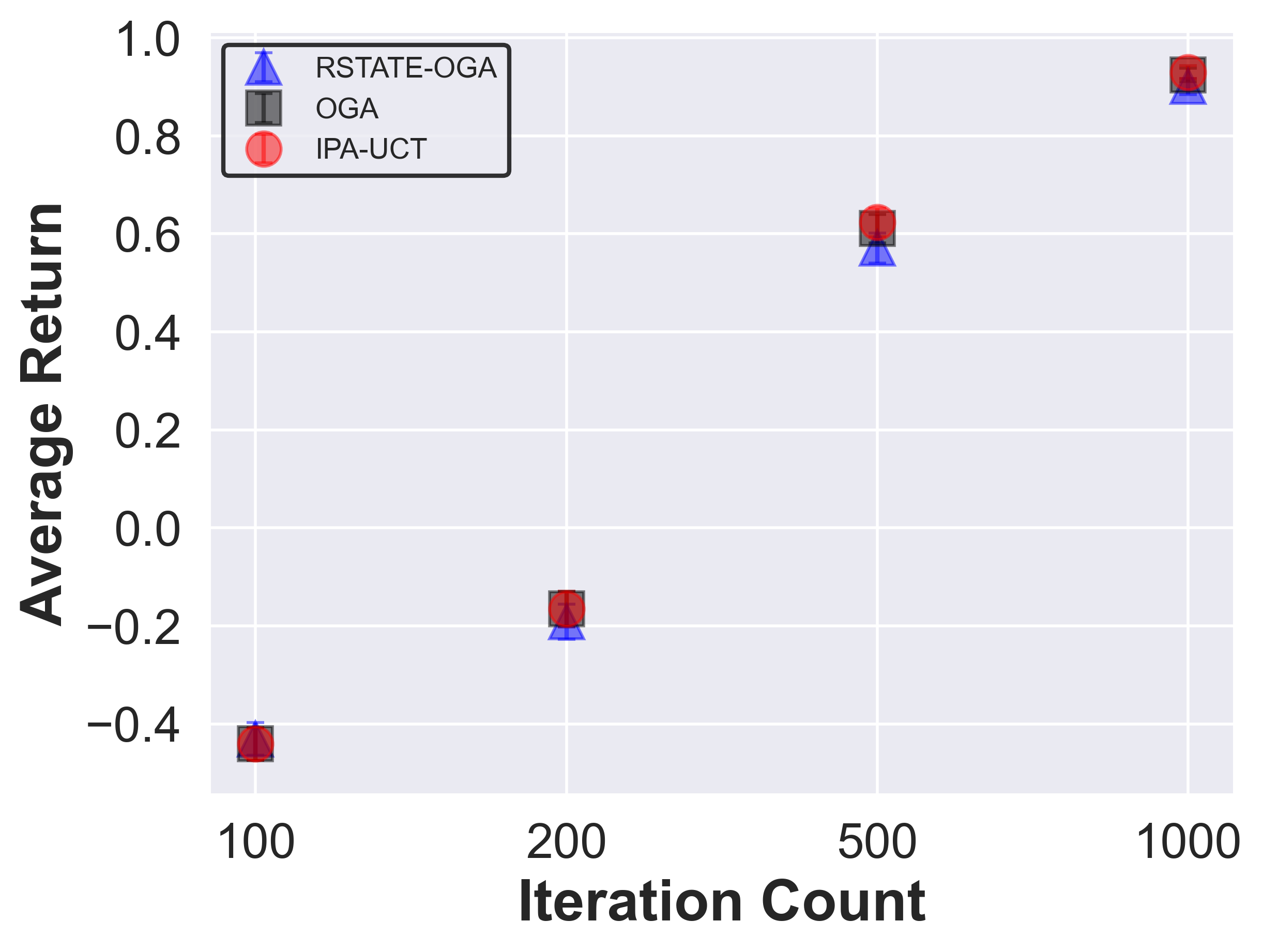}
\caption*{(p) Chess}
\end{minipage}
\hfill
\begin{minipage}{0.3\textwidth}
\centering
\includegraphics[width=\linewidth]{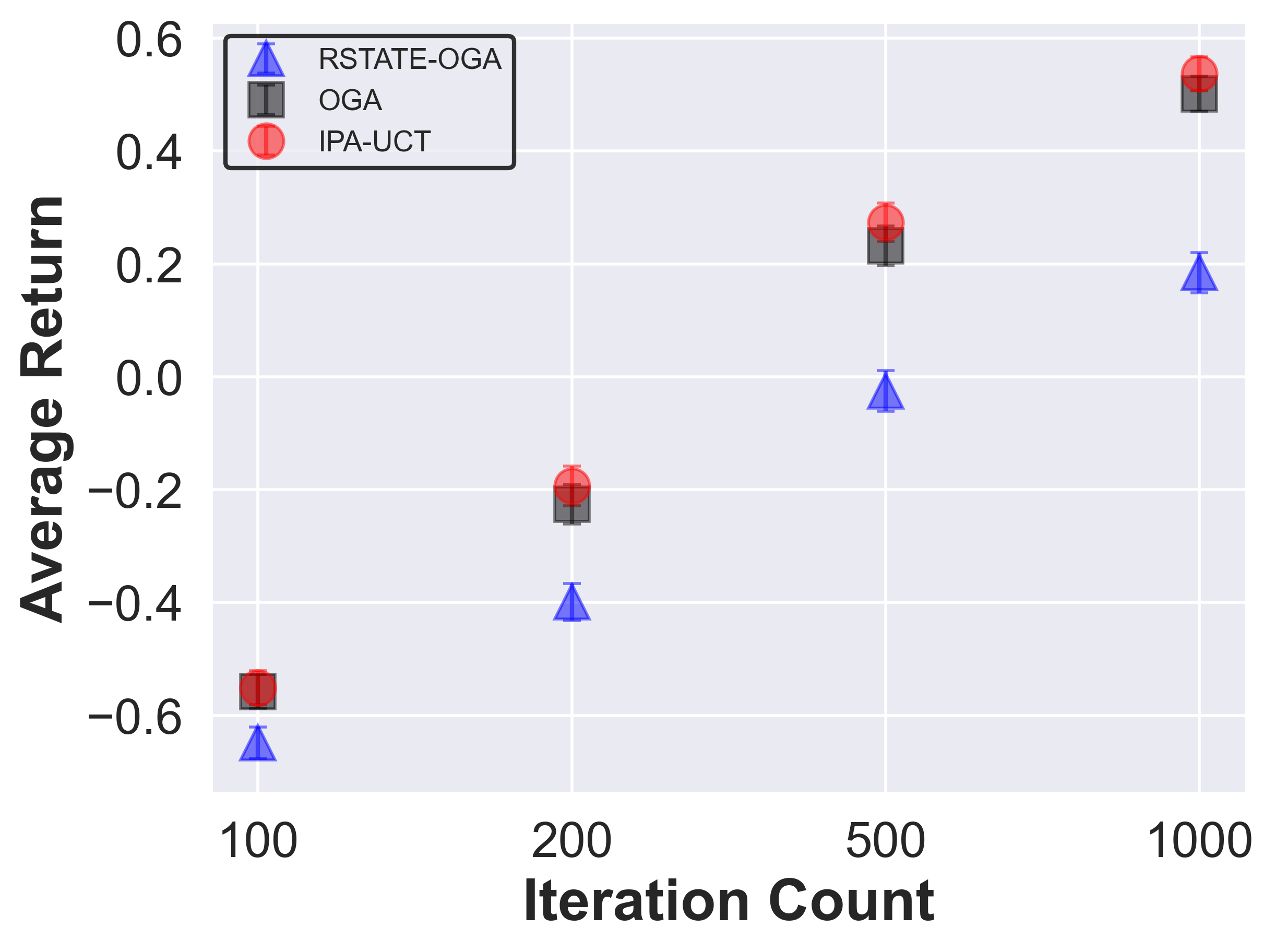}
\caption*{(q) Connect 4}
\end{minipage}
\hfill
\begin{minipage}{0.3\textwidth}
\centering
\includegraphics[width=\linewidth]{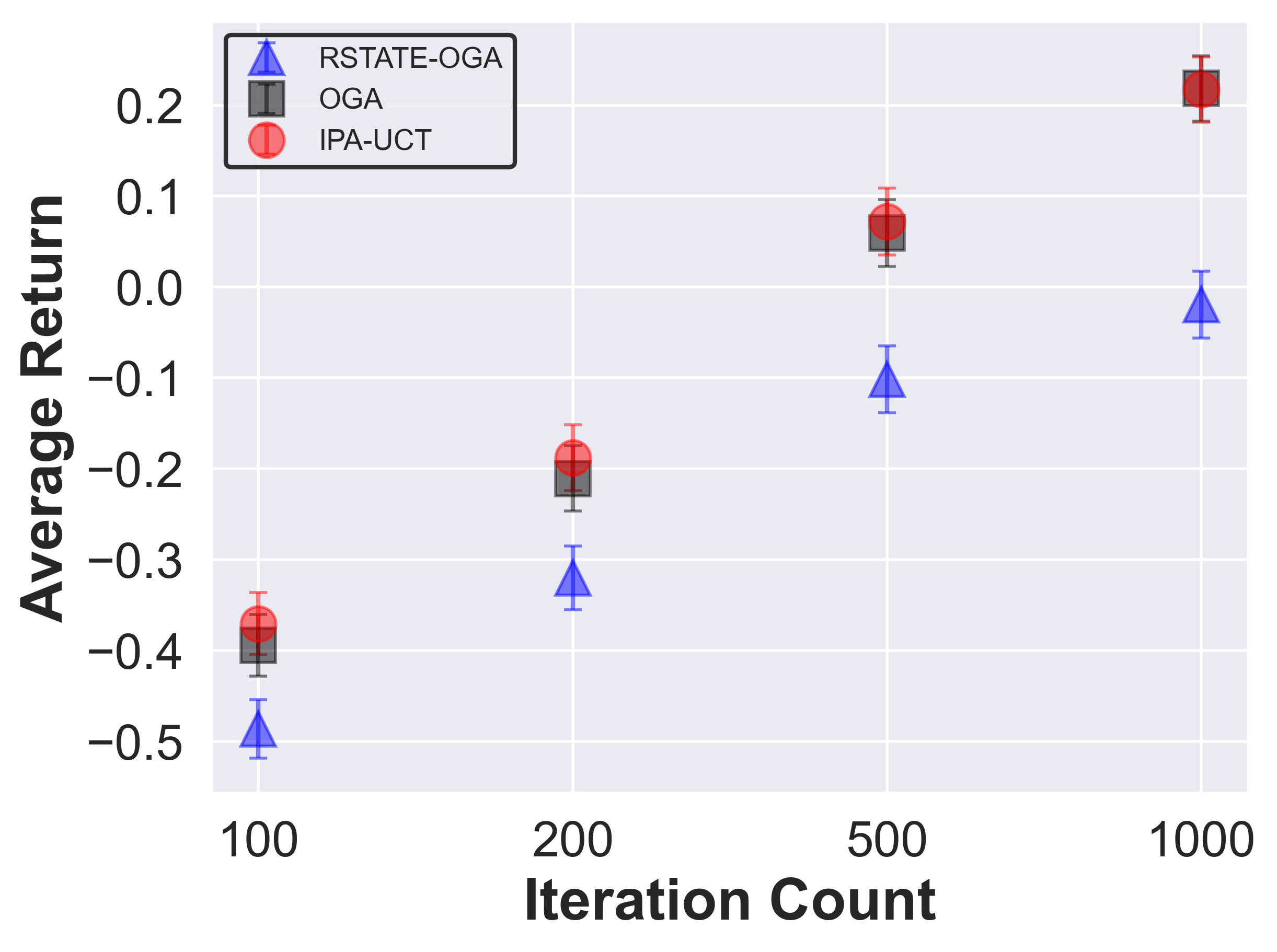}
\caption*{(r) Constrictor}
\end{minipage}
\hfill
\begin{minipage}{0.3\textwidth}
\centering
\includegraphics[width=\linewidth]{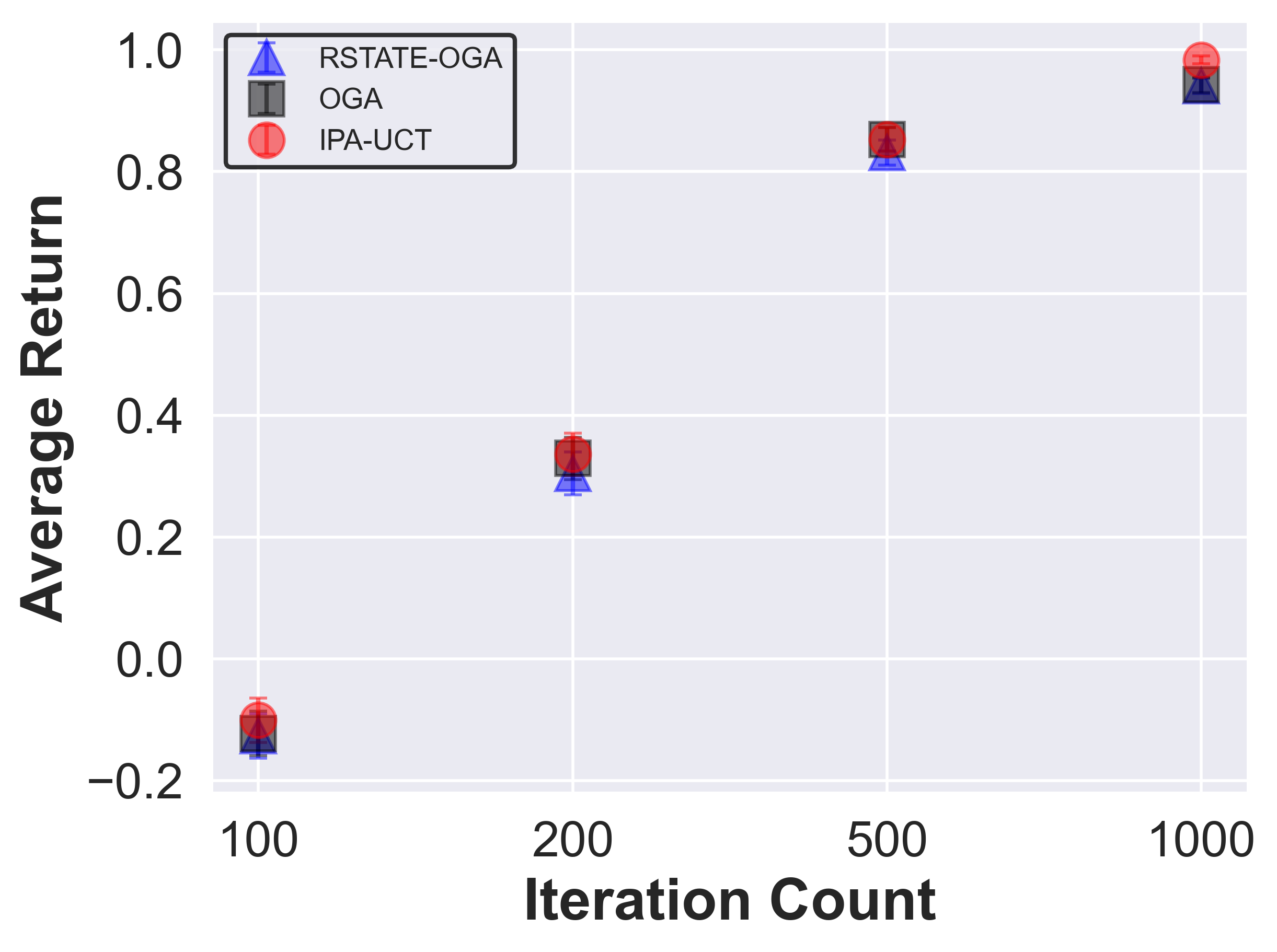}
\caption*{(s) Numbers Race}
\end{minipage}
\hfill
\begin{minipage}{0.3\textwidth}
\centering
\includegraphics[width=\linewidth]{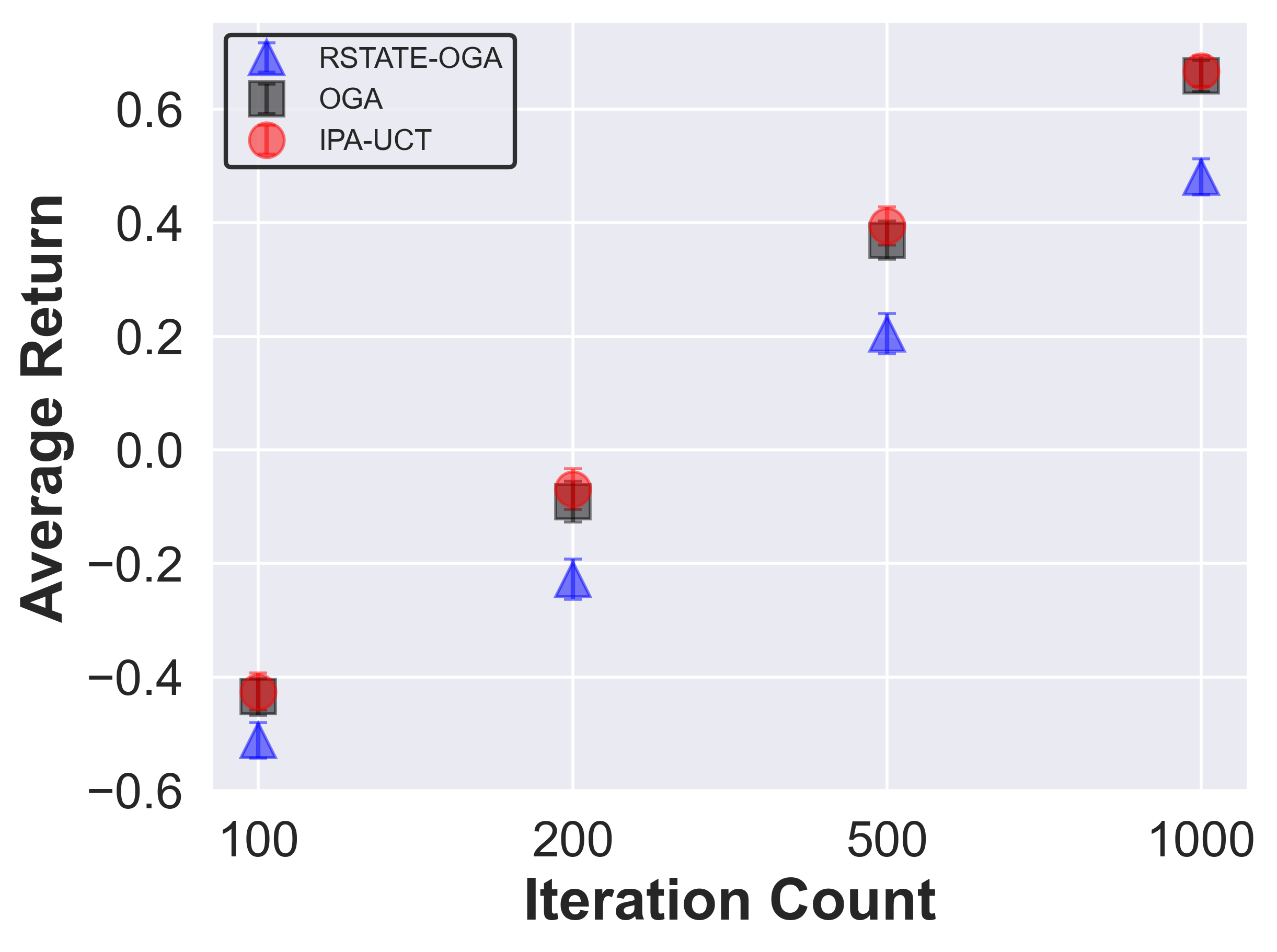}
\caption*{(t) Othello}
\end{minipage}
\hfill
\begin{minipage}{0.3\textwidth}
\centering
\includegraphics[width=\linewidth]{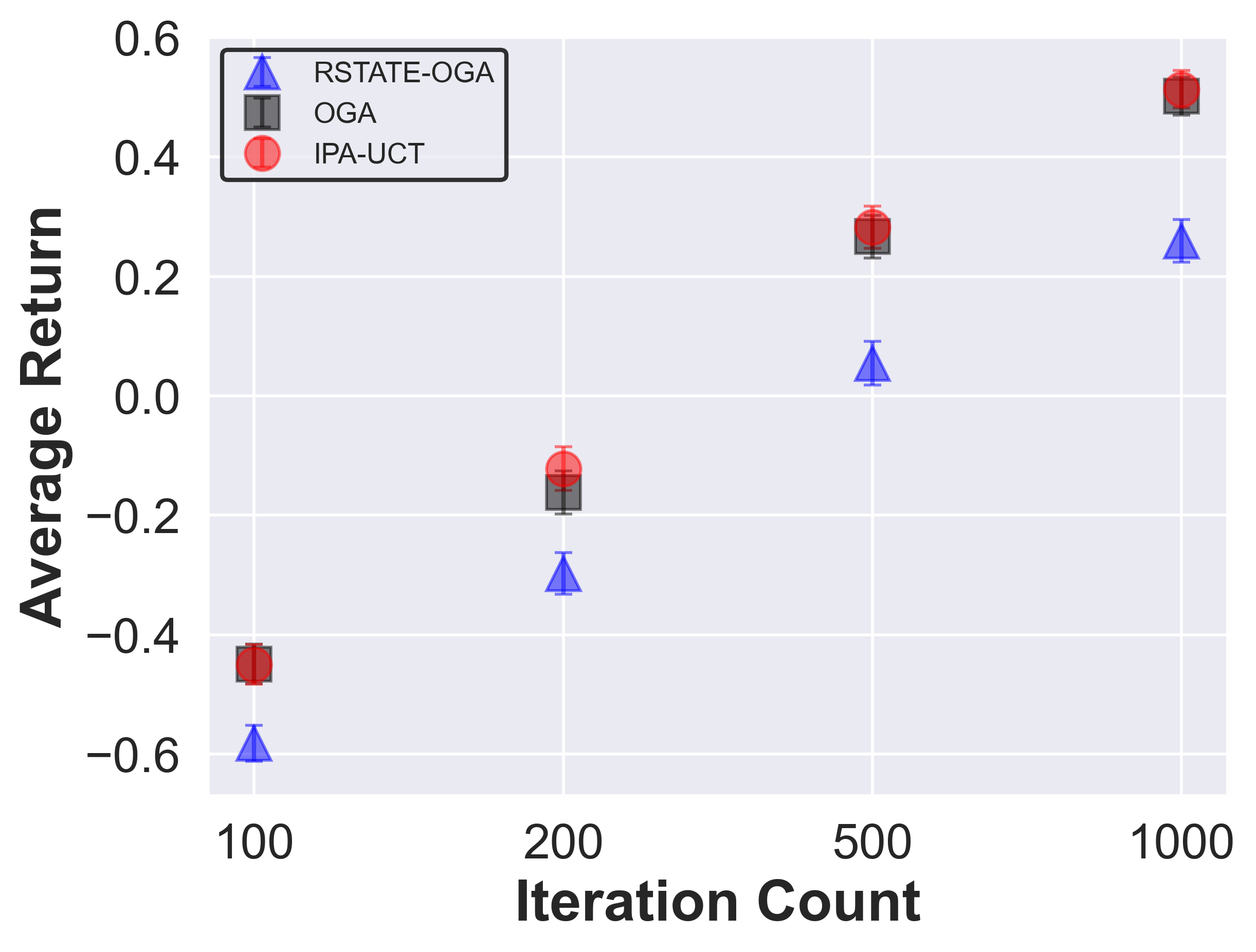}
\caption*{(u) Pylos}
\end{minipage}
\hfill
\begin{minipage}{0.3\textwidth}
\centering
\includegraphics[width=\linewidth]{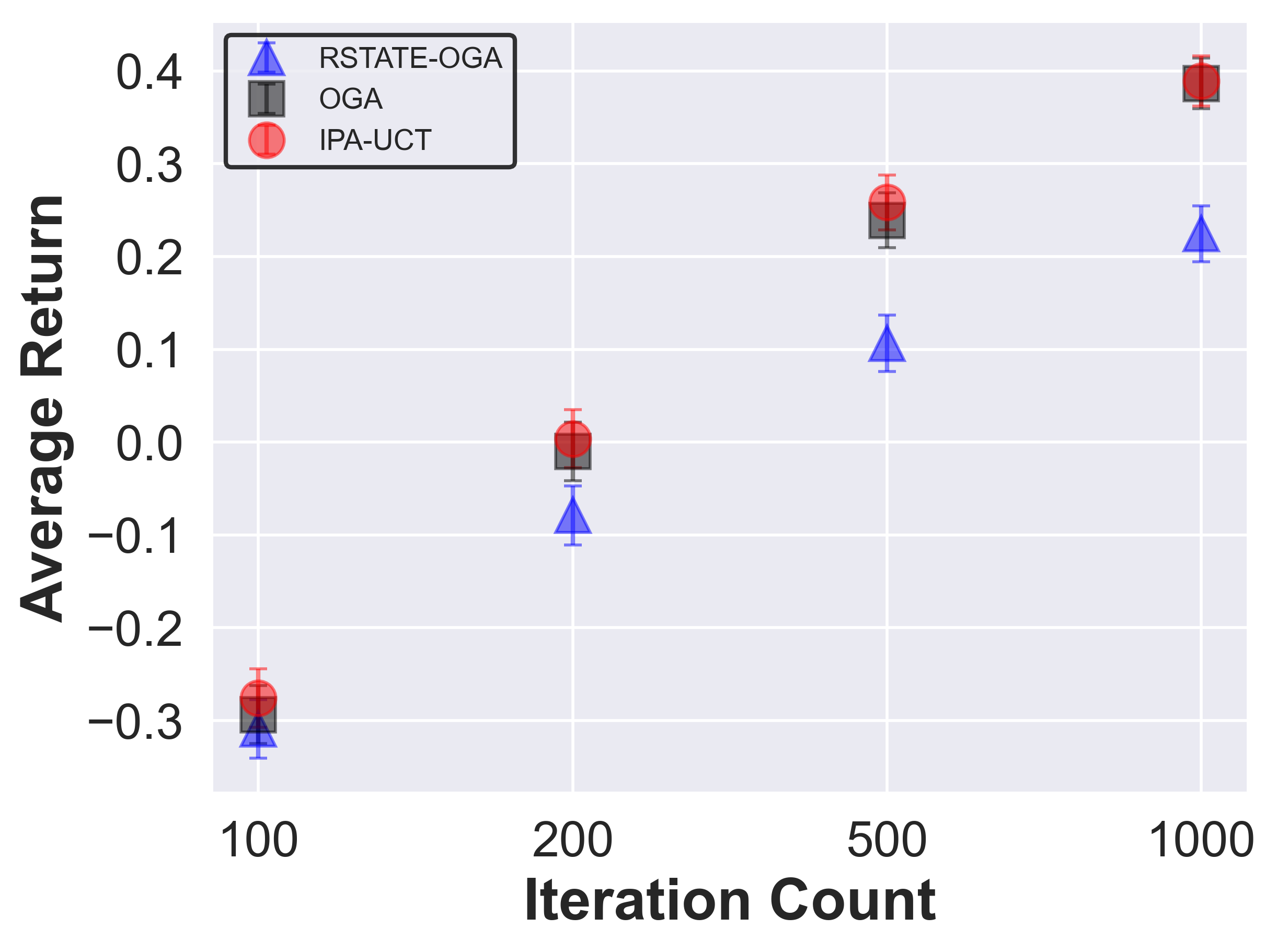}
\caption*{(v) Quarto}
\end{minipage}
\hfill
\begin{minipage}{0.3\textwidth}
\centering
\includegraphics[width=\linewidth]{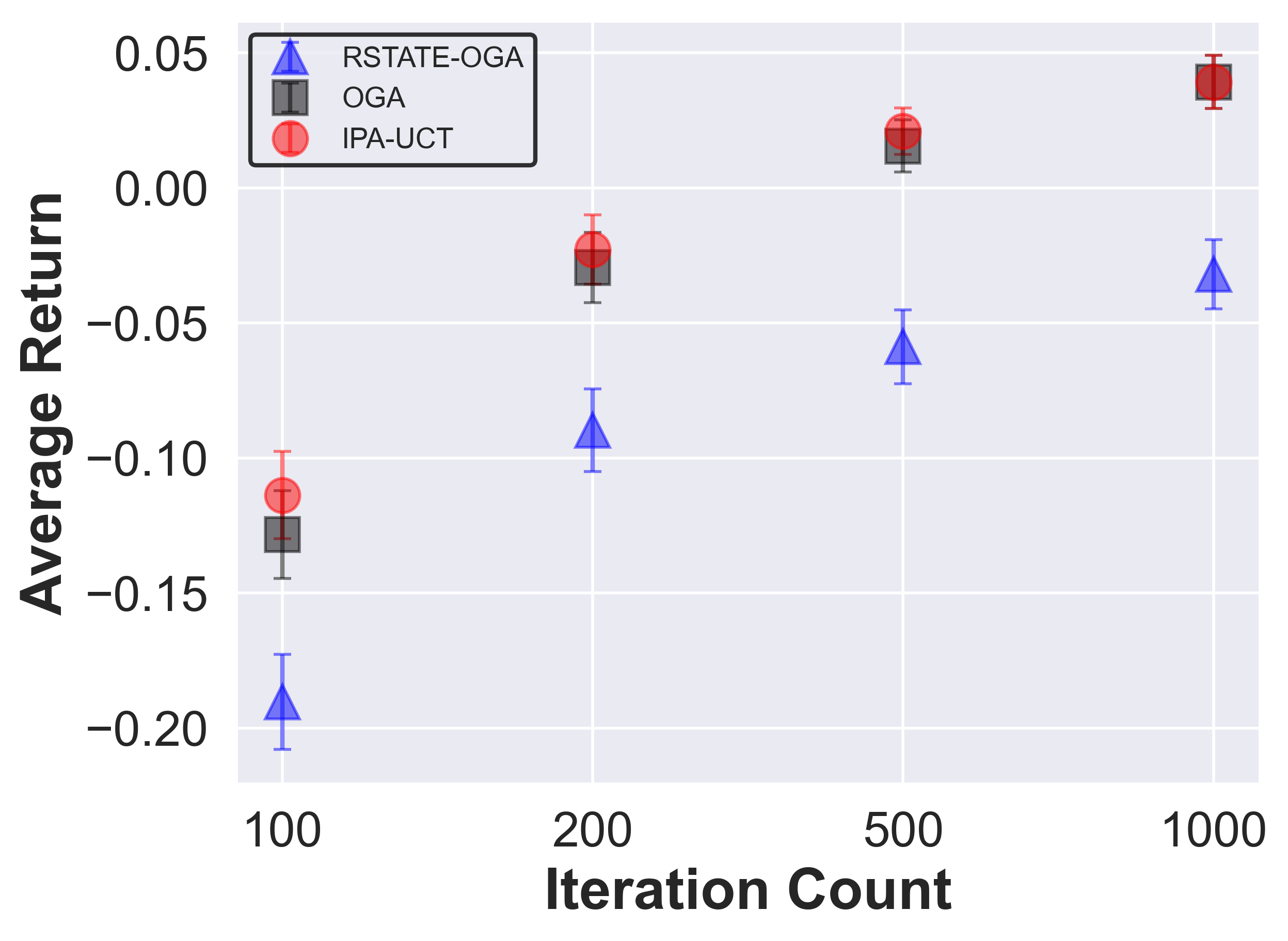}
\caption*{(w) Tic Tac Toe}
\end{minipage}
\hfill
\begin{minipage}{0.3\textwidth}
\centering
\includegraphics[width=\linewidth]{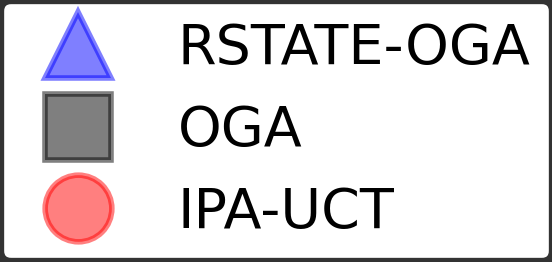}
\caption*{Legend}
\end{minipage}
\hfill
\caption{The performance graphs for all problems of in dependence of the MCTS iteration count of the parameter optimized versions of IPA-UCT versus RSTATE-OGA versus pruned OGA and $(\varepsilon_{\text{a}},\varepsilon_{\text{t}})$-OGA (summarized as OGA). The parameters over which the agents were optimized are identical to those in Section \ref{sec:experiments} except that we used hand picked $\varepsilon_{\text{a}}$ values for each environment which are listed in Tab.~\ref{tab:epsa_values}.}
\label{fig:ipa:optimized_mp}
\end{figure}

Fig.~\ref{fig:ipa:optimized_mp} compares the parameter-optimized performances of IPA-UCT, pruned OGA and $(\varepsilon_{\text{a}},\varepsilon_{\text{t}})$-OGA (summarized simply as OGA) as well as RSTATE-OGA using the same parameters as in the main experimental section \ref{sec:experiments}. 

The key observation that can be made is that IPA-UCT has only limited use in improving the parameter-optimized, as the gains (if any) are only marginal and could be explained due to noise, except for the Cooperative Recon environment where IPA-UCT can a clear advantage. In Manufacturer, Racetrack, Sailing Wind, Tamarisk, Connect 4, Pylos, and Othello, there seems to be a significant, however extremely small gain.

Though IPA-UCT will show more promise in the generalization experiments presented in the main section,  the reasons for the negligible impact of IPA in this setting will be discussed, which can be attributed to three criteria that have to be satisfied for IPA-UCT to have a significant impact, which altogether can be quite rare.

\begin{enumerate}

    \item The domain must have a small action space. For $J_{\text{UCB}}$ to prune an action, it must have a low enough exploration term, which shrinks only with the number of visits. If there are too many actions, the visits will be spread too much. This explains the lack of impact in Academic Advising, Game of Life, or SysAdmin. Of course, using $\lambda_{\text{p}}=0$ does pruning even with no visits; however, we consider performance gains by pruning under such uncertainty as simply lucky. 
    
    \item Ultimately, IPA (as well as ASAP) requires state-action pair abstractions to bootstrap off of. Hence, if almost no action abstractions are found, then IPA cannot detect any state abstractions. Due to its extremely high stochasticity, this is the case for Earth Observation where with reasonable $\varepsilon_{\text{t}}$ values, no action abstractions are found.
    
    \item There must be state equivalences in the first place. Some environments like Sailing Wind or the here-considered Navigation instance feature almost no state-equivalences (for reference, check their corresponding abstraction rate Tab.~\ref{tab:qtable})
\end{enumerate}

\begin{figure}[H]
\centering

\begin{minipage}{0.3\textwidth}
\centering
\includegraphics[width=\linewidth]{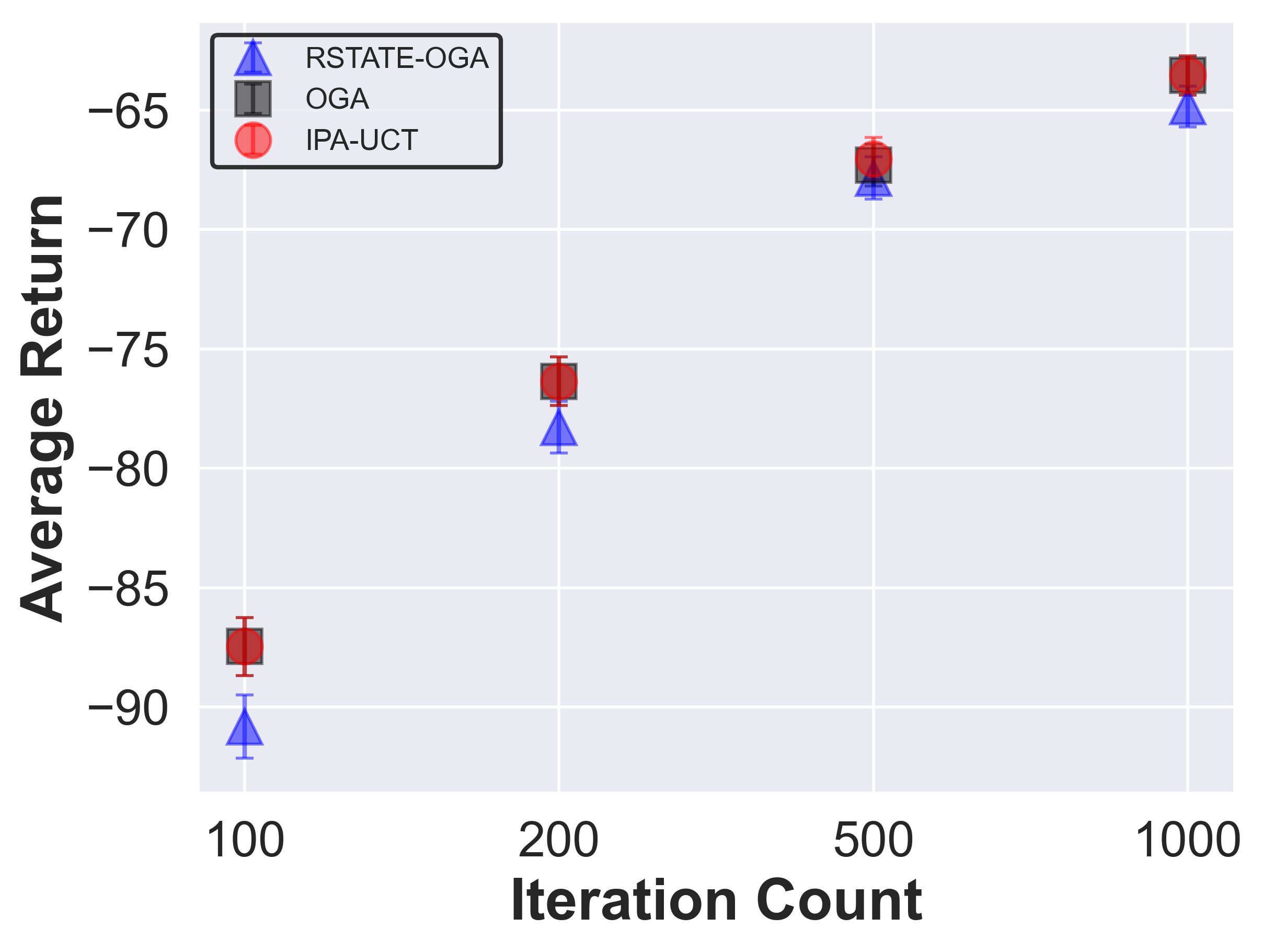}
\caption*{(a) Academic Advising}
\end{minipage}
\hfill
\begin{minipage}{0.3\textwidth}
\centering
\includegraphics[width=\linewidth]{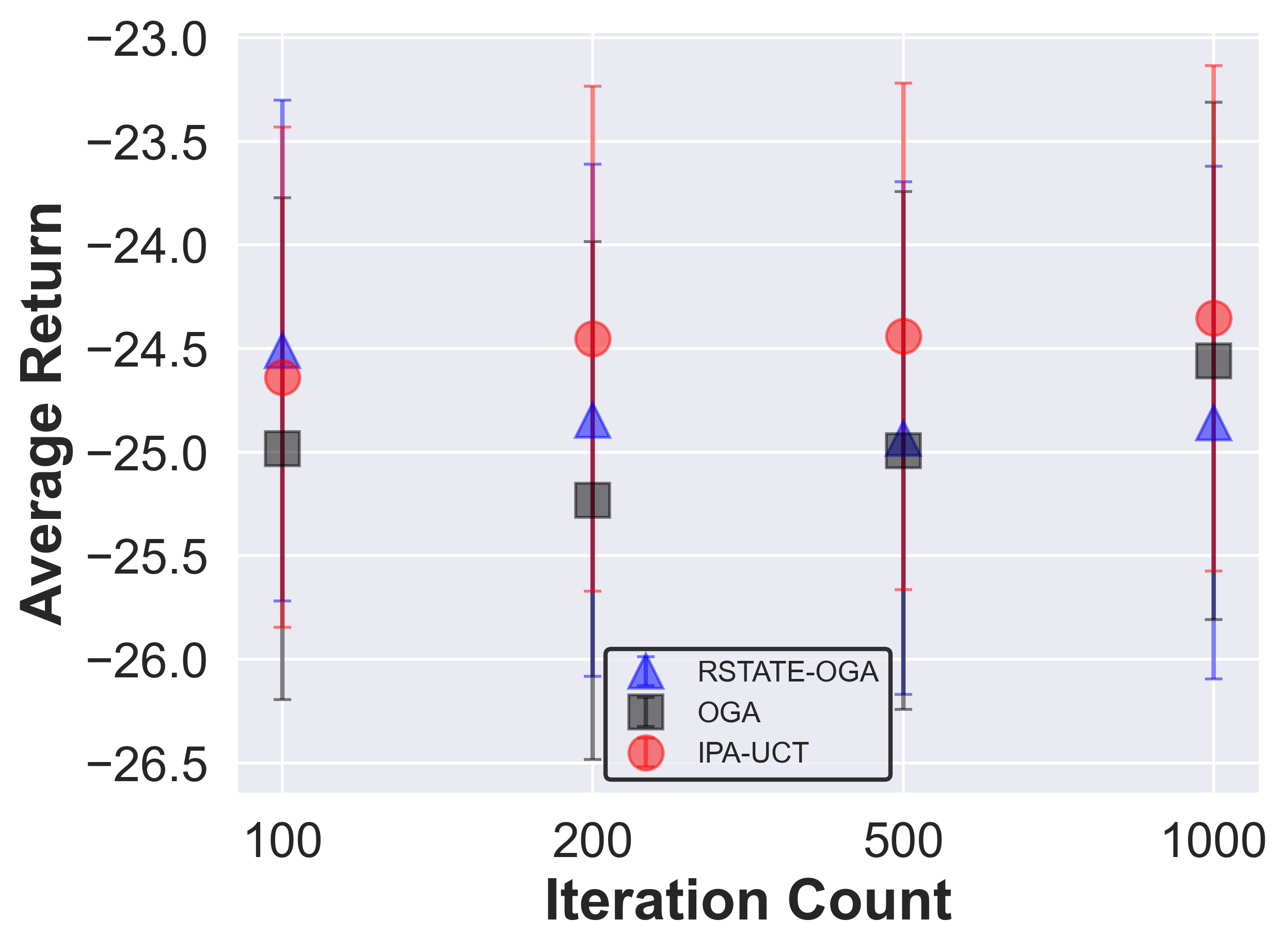}
\caption*{(b) Crossing Traffic}
\end{minipage}
\hfill
\begin{minipage}{0.3\textwidth}
\centering
\includegraphics[width=\linewidth]{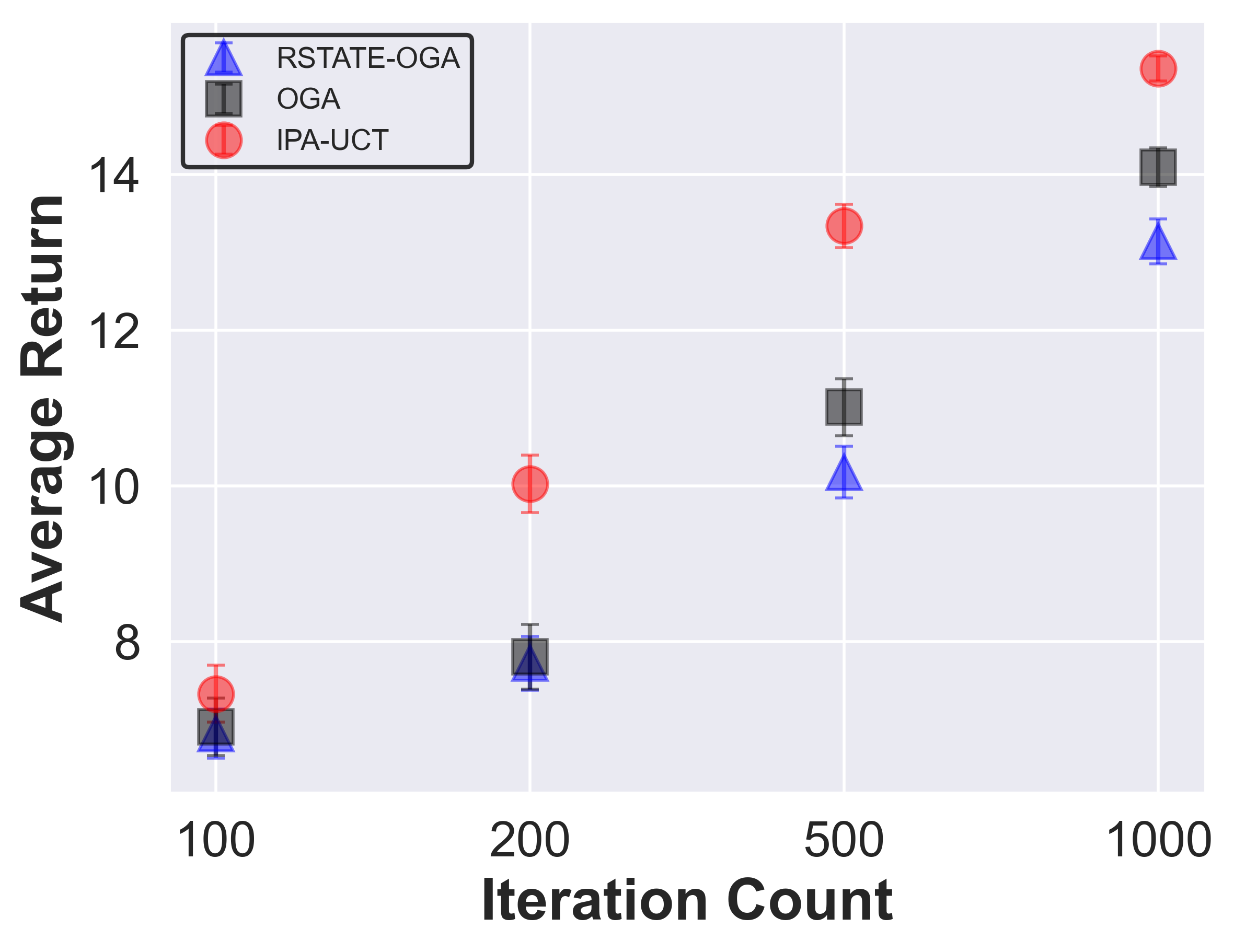}
\caption*{(c) Cooperative Recon}
\end{minipage}
\hfill
\begin{minipage}{0.3\textwidth}
\centering
\includegraphics[width=\linewidth]{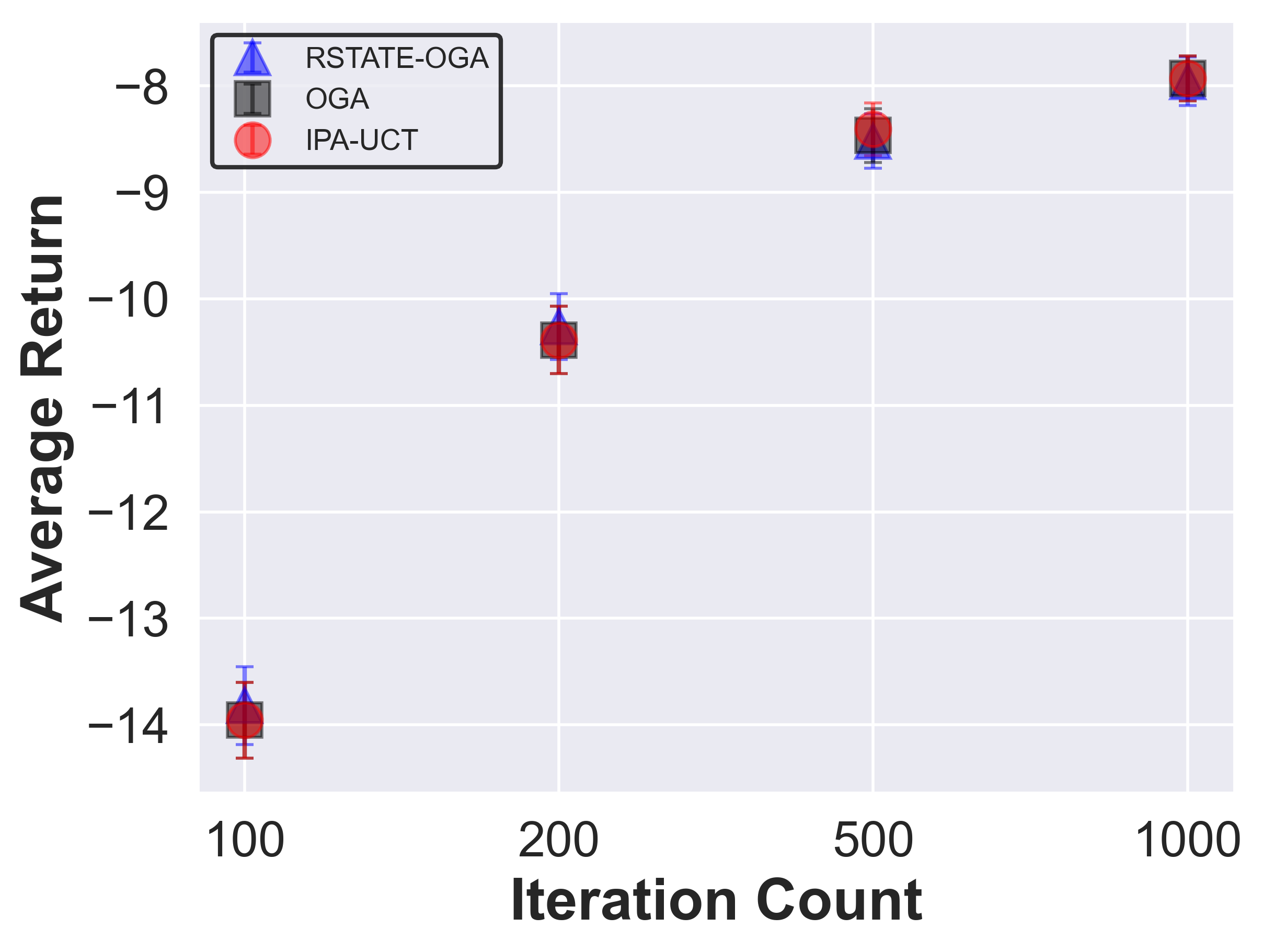}
\caption*{(d) Earth Observation}
\end{minipage}
\hfill
\begin{minipage}{0.3\textwidth}
\centering
\includegraphics[width=\linewidth]{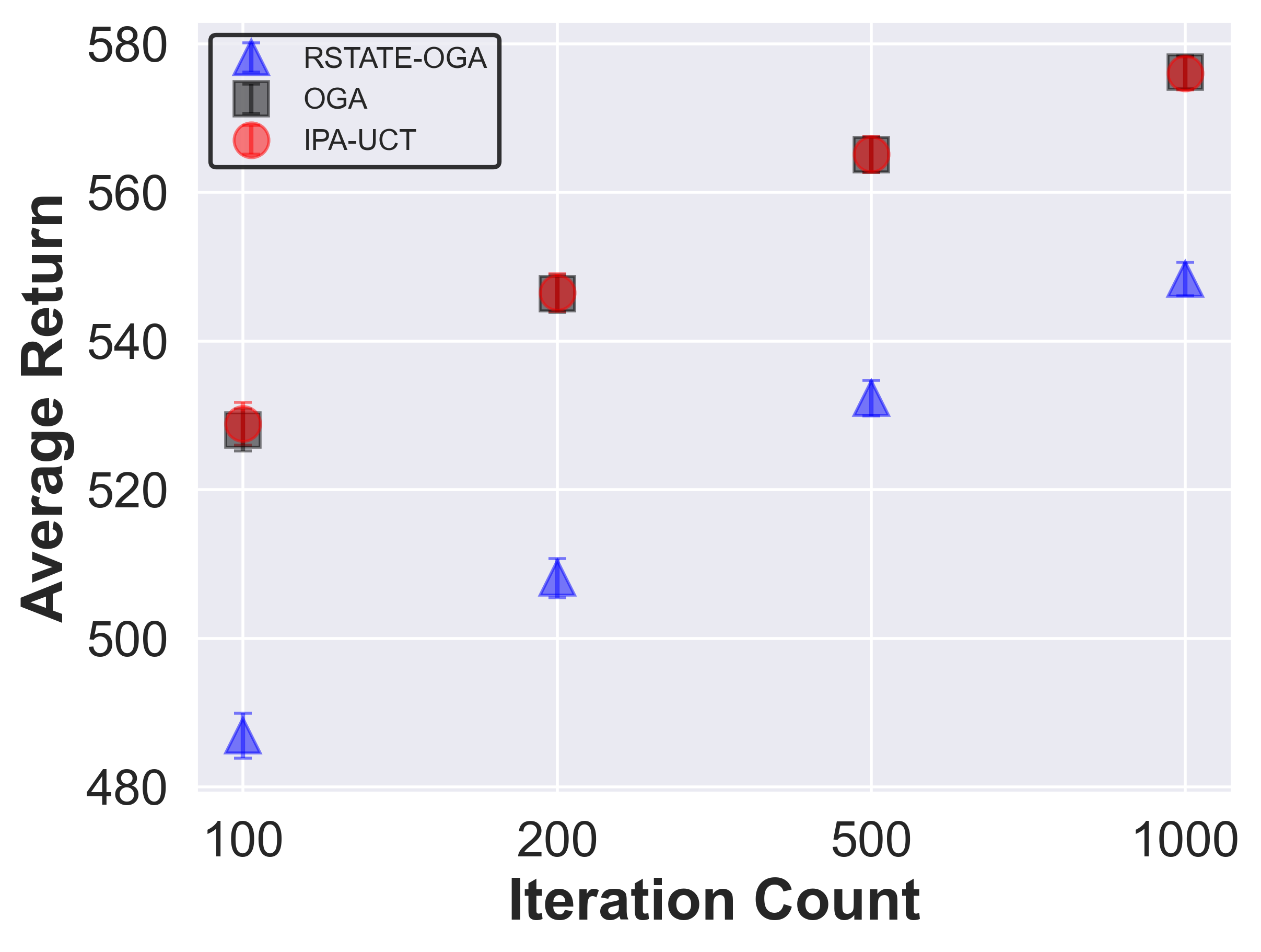}
\caption*{(e) Game of Life}
\end{minipage}
\hfill
\begin{minipage}{0.3\textwidth}
\centering
\includegraphics[width=\linewidth]{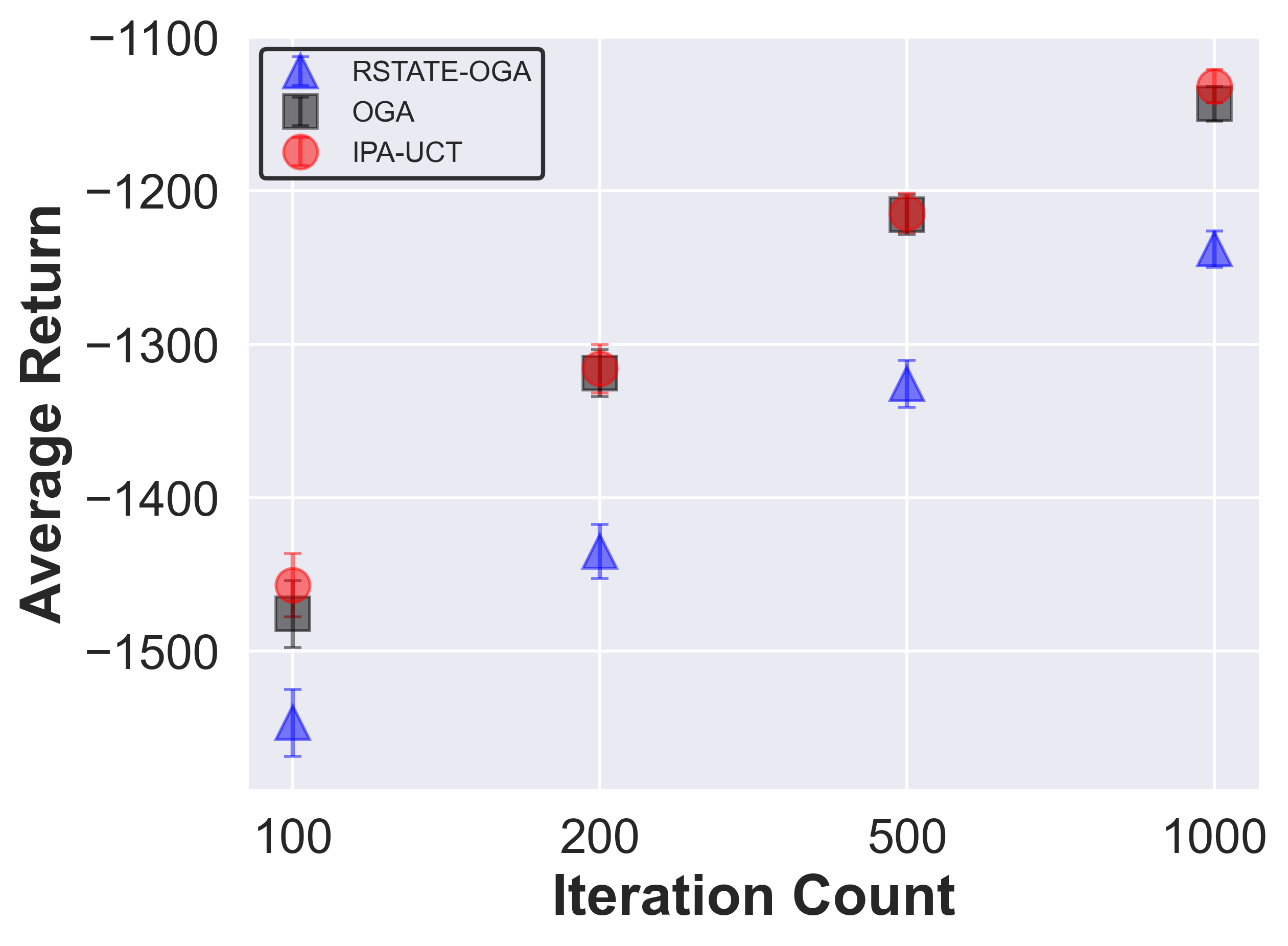}
\caption*{(f) Manufacturer}
\end{minipage}
\hfill
\begin{minipage}{0.3\textwidth}
\centering
\includegraphics[width=\linewidth]{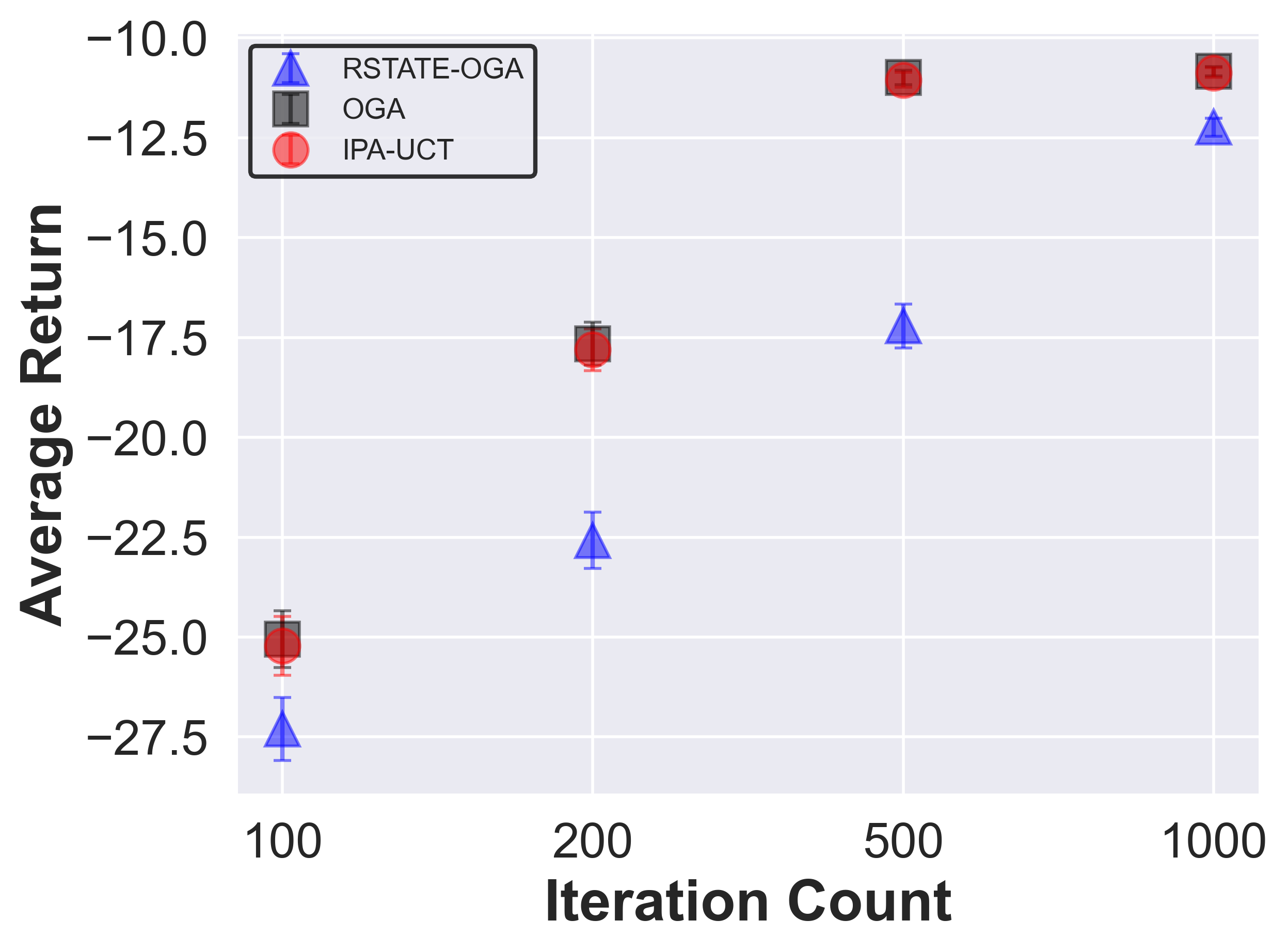}
\caption*{(g) Navigation}
\end{minipage}
\hfill
\begin{minipage}{0.3\textwidth}
\centering
\includegraphics[width=\linewidth]{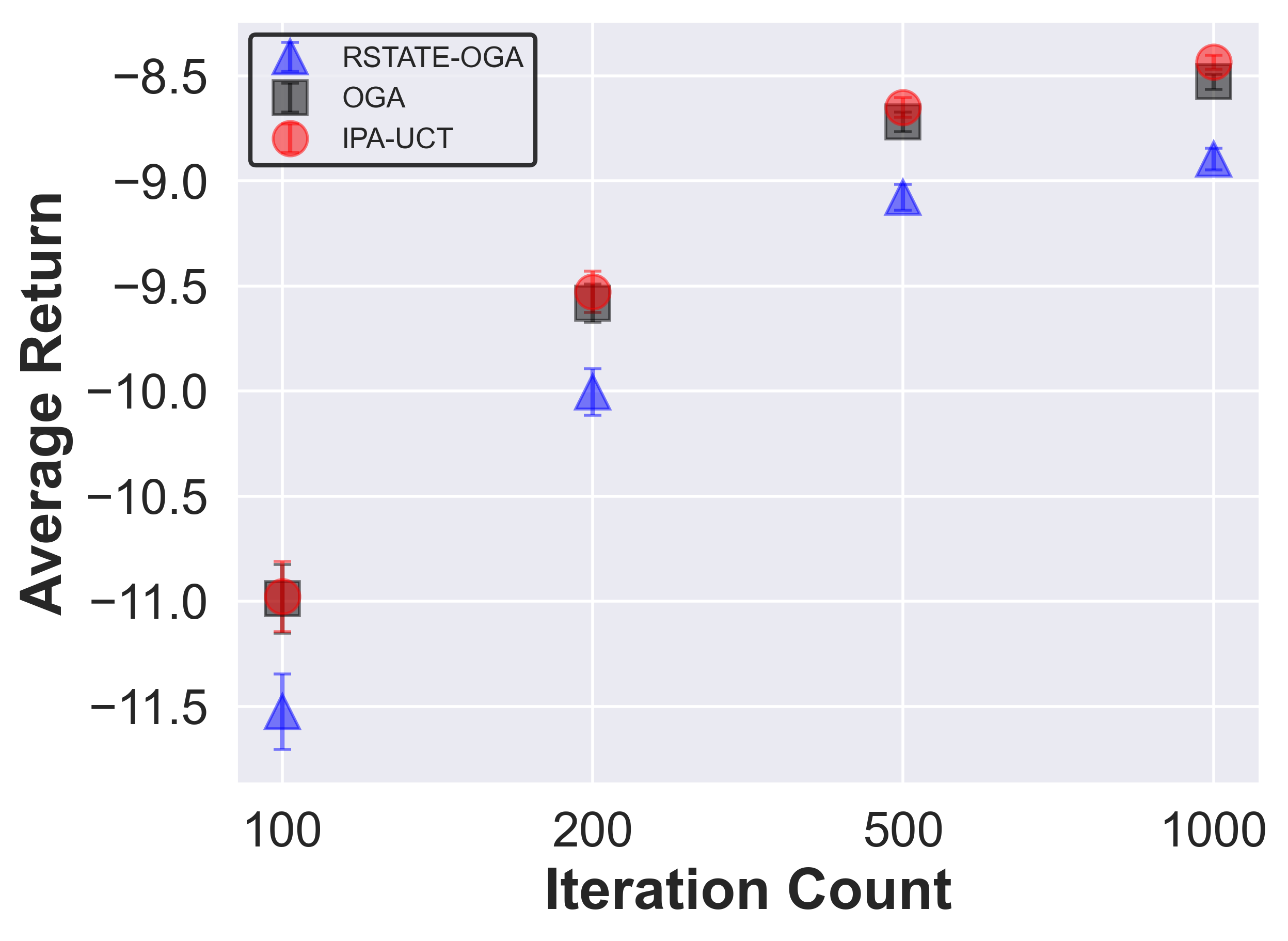}
\caption*{(h) Racetrack}
\end{minipage}
\hfill
\begin{minipage}{0.3\textwidth}
\centering
\includegraphics[width=\linewidth]{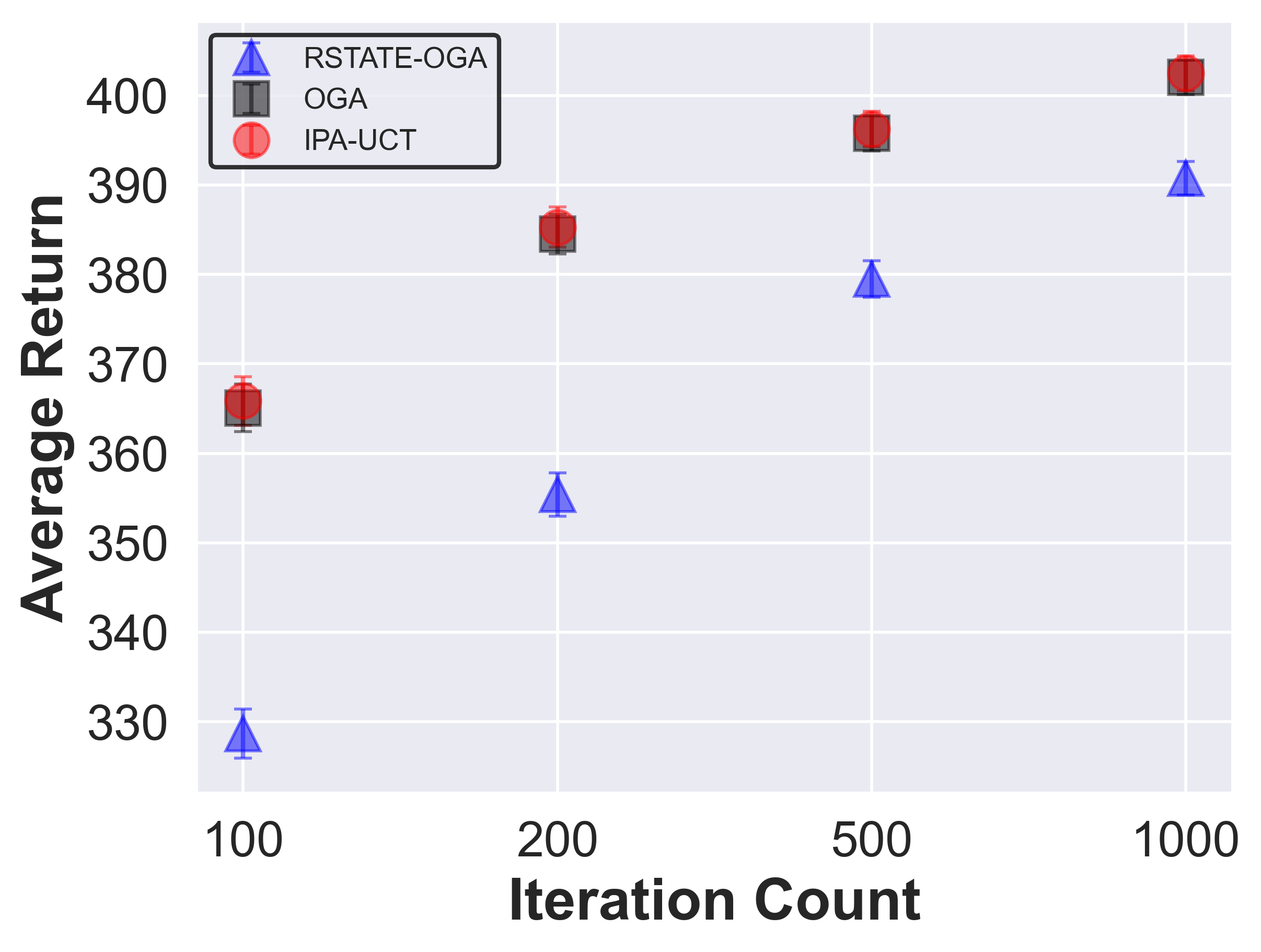}
\caption*{(i) SysAdmin}
\end{minipage}
\hfill
\begin{minipage}{0.3\textwidth}
\centering
\includegraphics[width=\linewidth]{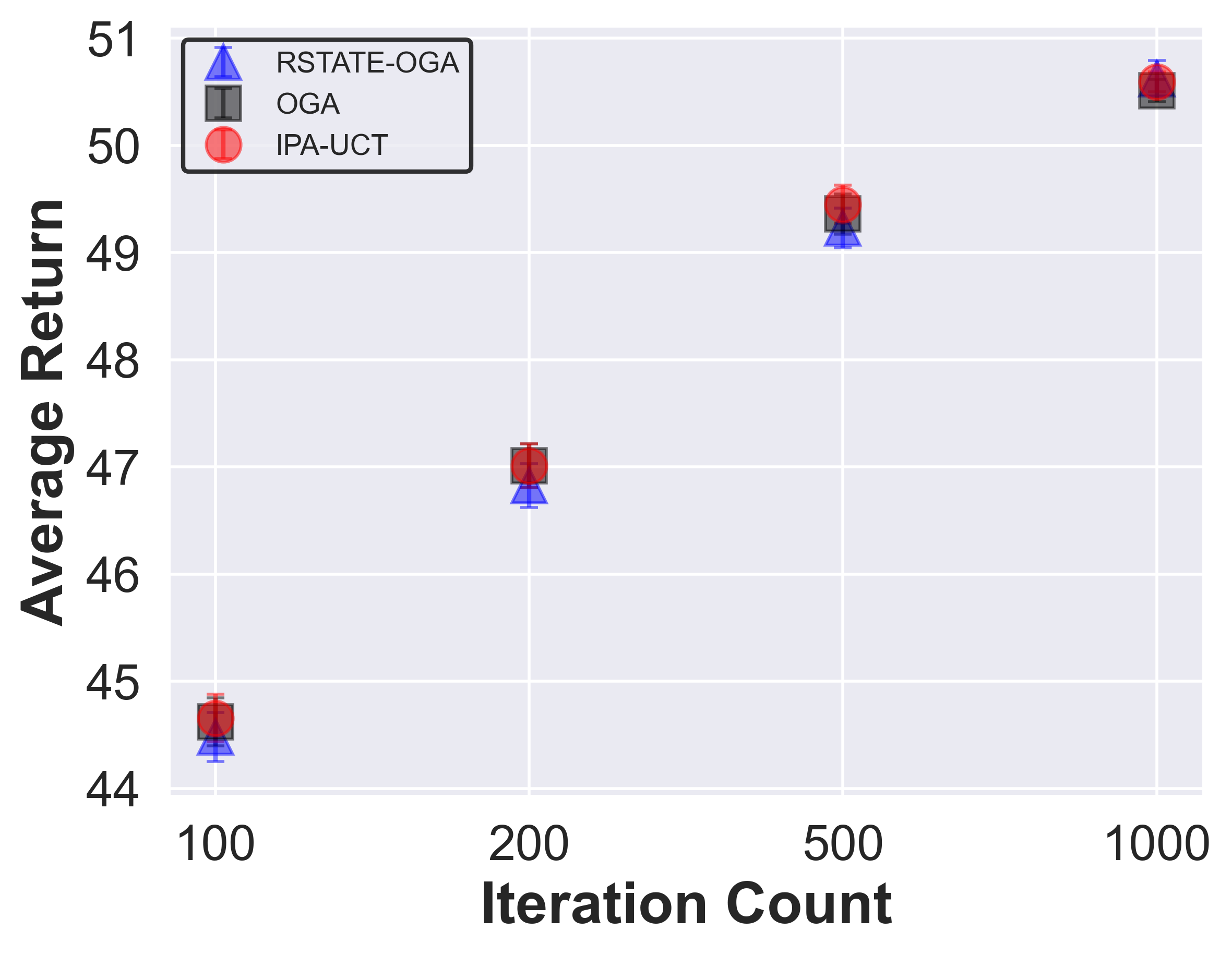}
\caption*{(j) Saving}
\end{minipage}
\hfill
\begin{minipage}{0.3\textwidth}
\centering
\includegraphics[width=\linewidth]{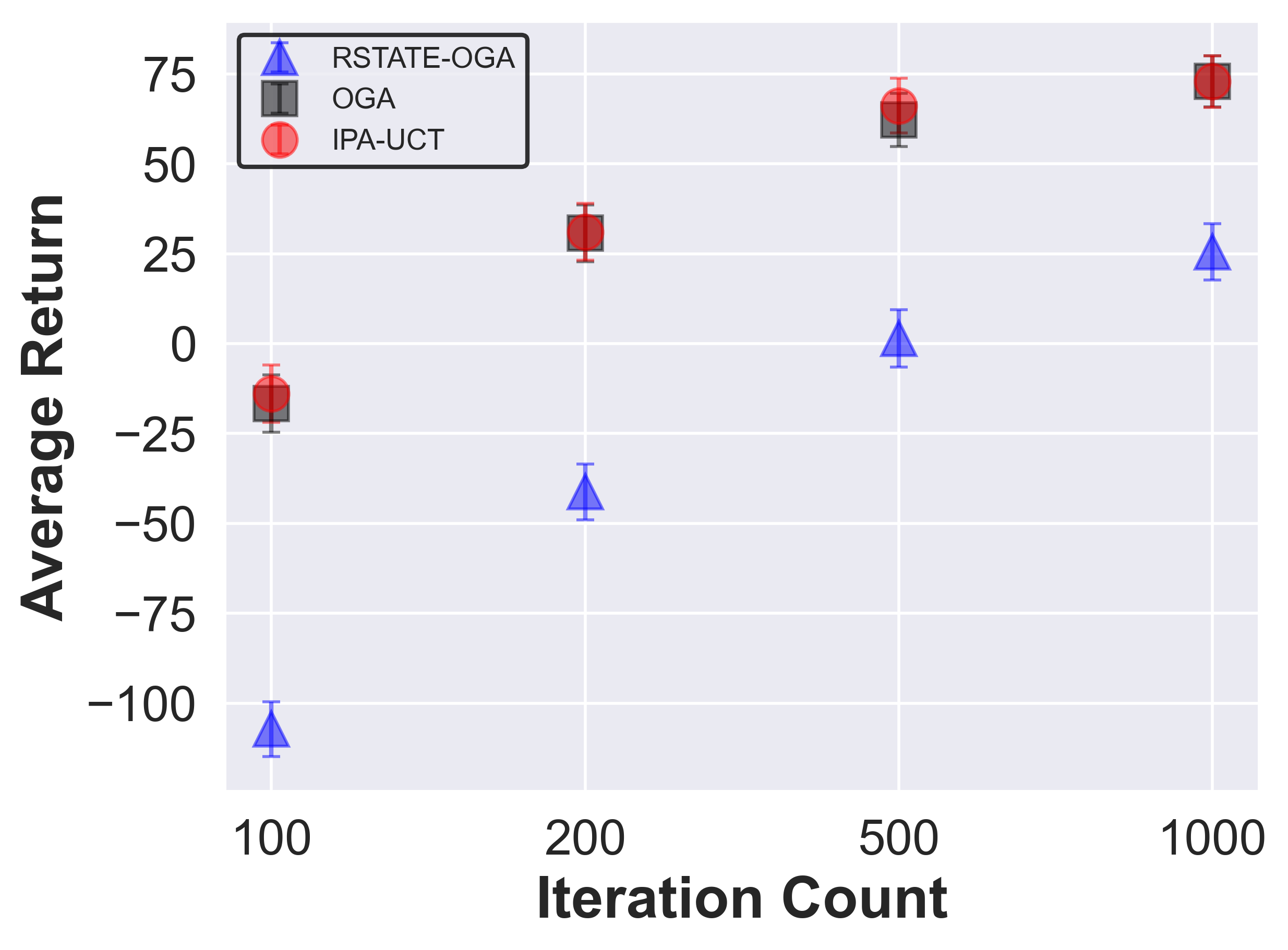}
\caption*{(k) Skill Teaching}
\end{minipage}
\hfill
\begin{minipage}{0.3\textwidth}
\centering
\includegraphics[width=\linewidth]{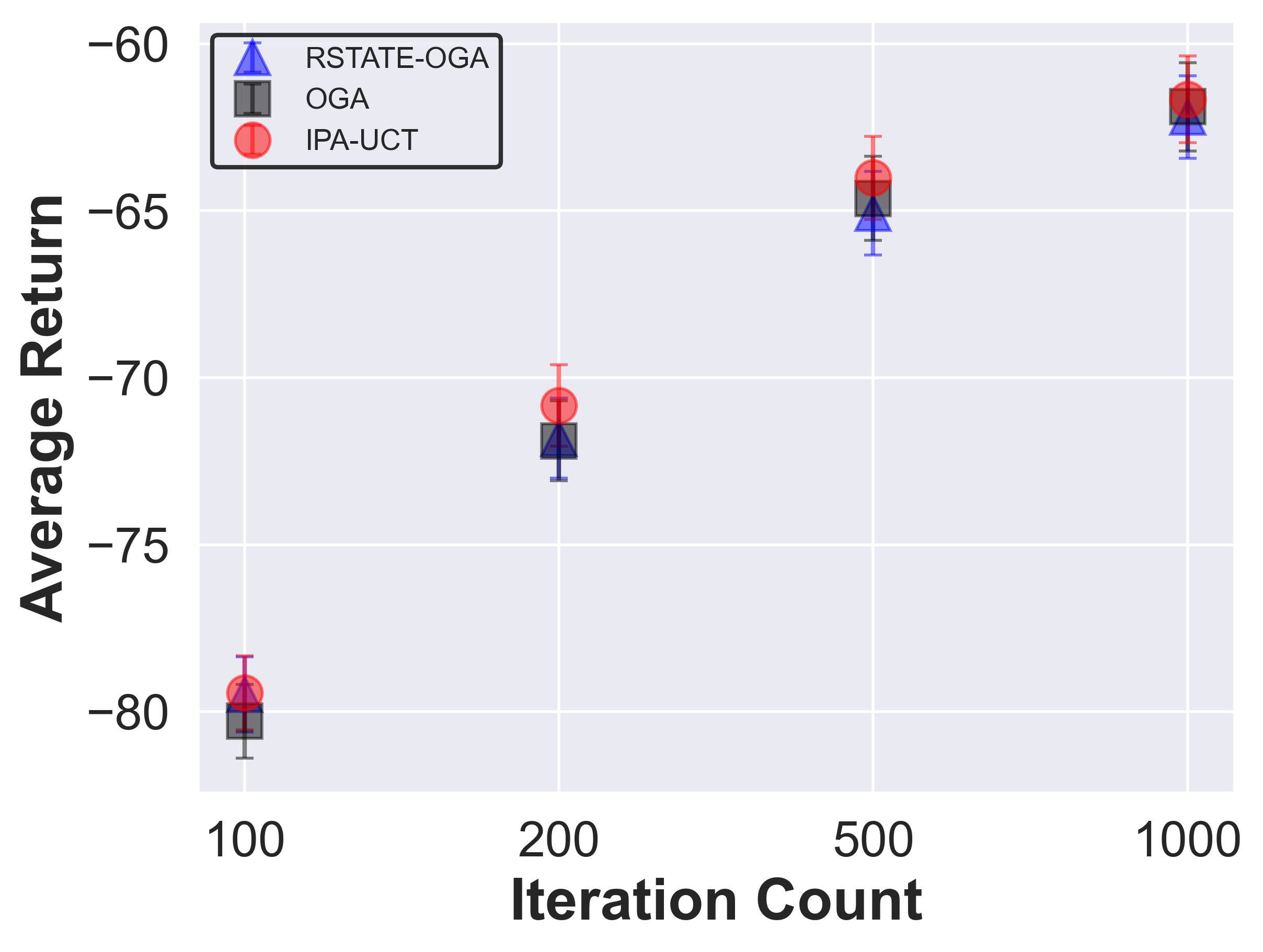}
\caption*{(l) Sailing Wind}
\end{minipage}
\hfill
\begin{minipage}{0.3\textwidth}
\centering
\includegraphics[width=\linewidth]{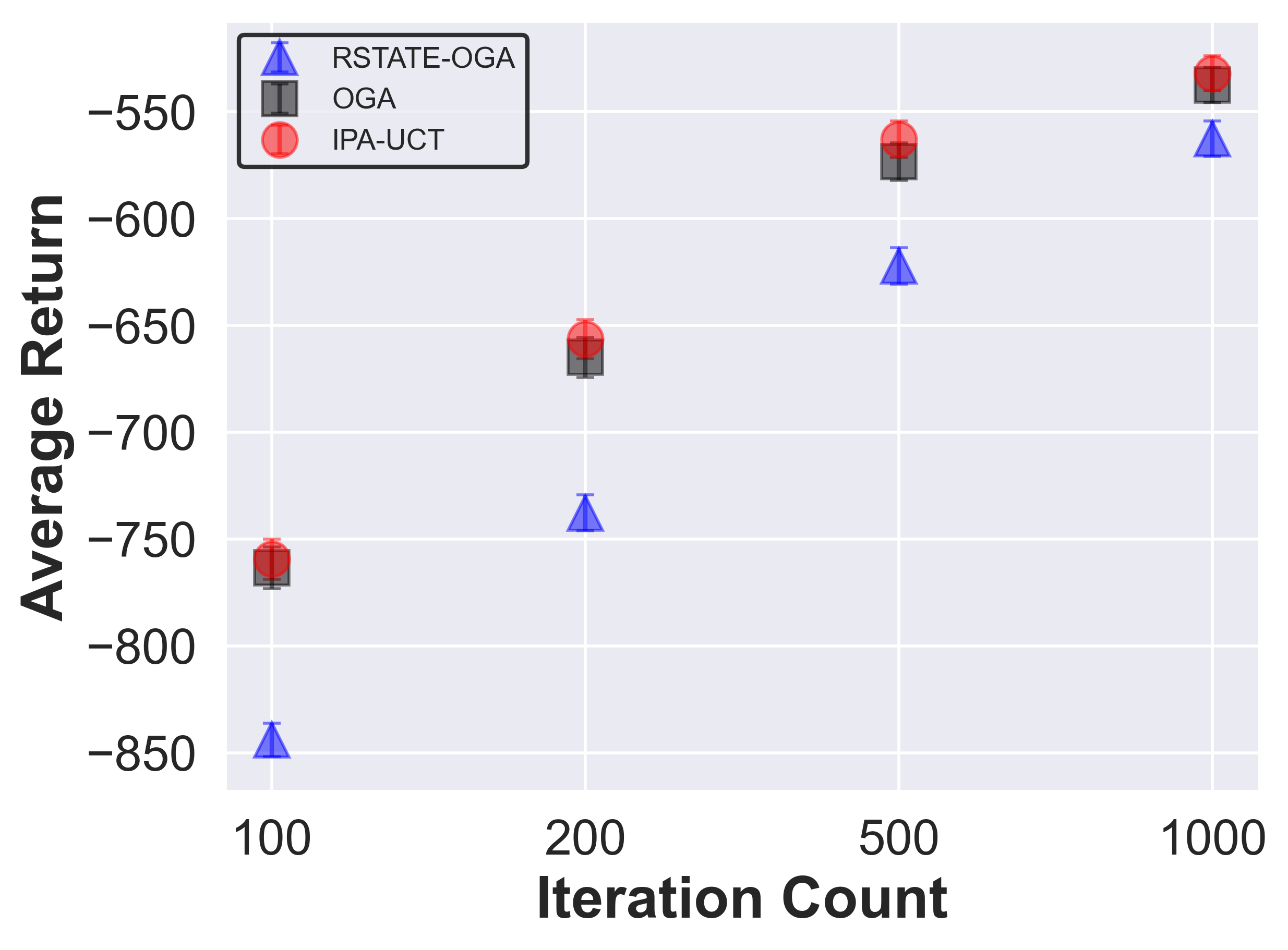}
\caption*{(m) Tamarisk}
\end{minipage}
\hfill
\begin{minipage}{0.3\textwidth}
\centering
\includegraphics[width=\linewidth]{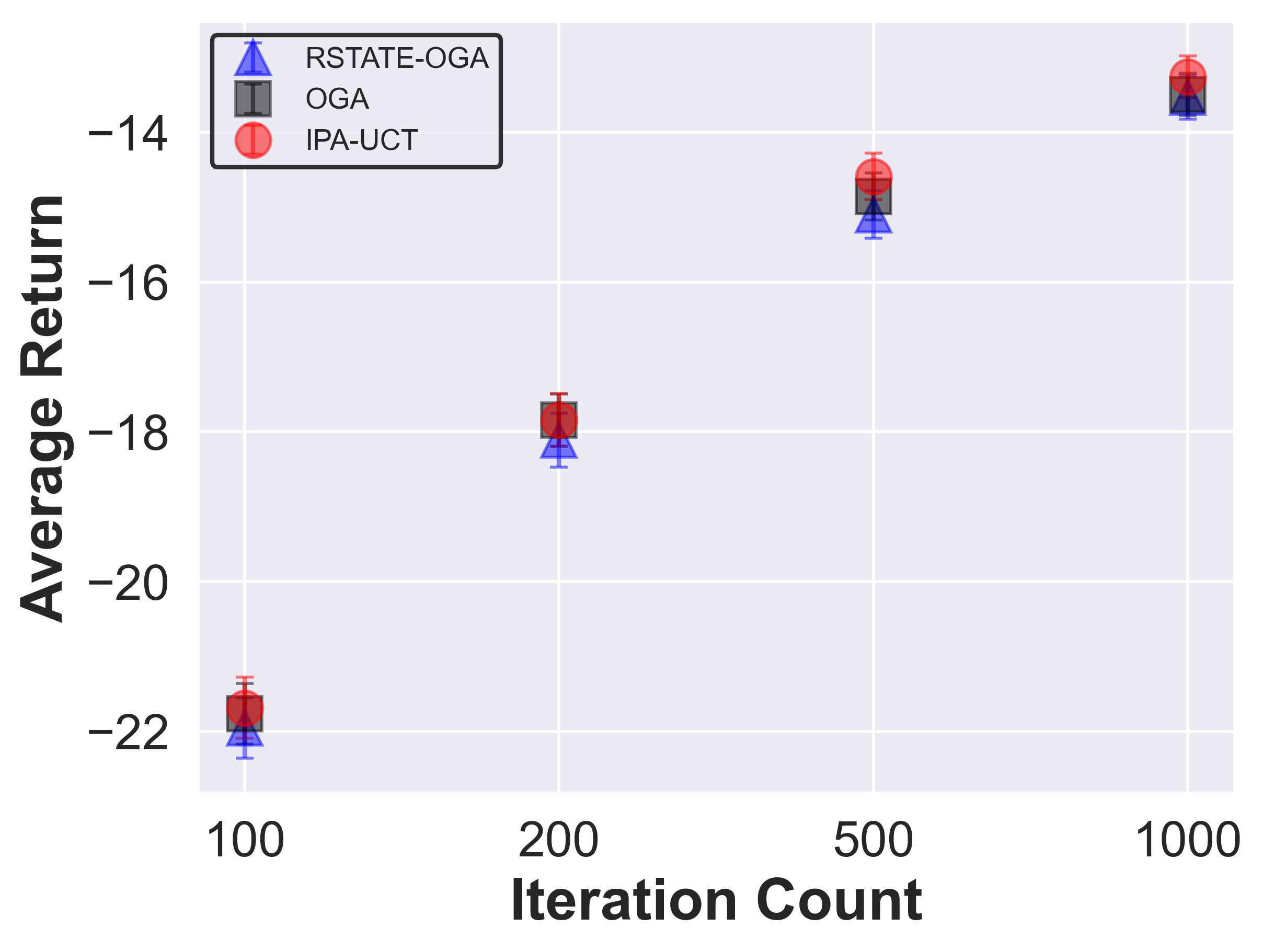}
\caption*{(n) Traffic}
\hfill
\end{minipage}
\begin{minipage}{0.3\textwidth}
\centering
\includegraphics[width=\linewidth]{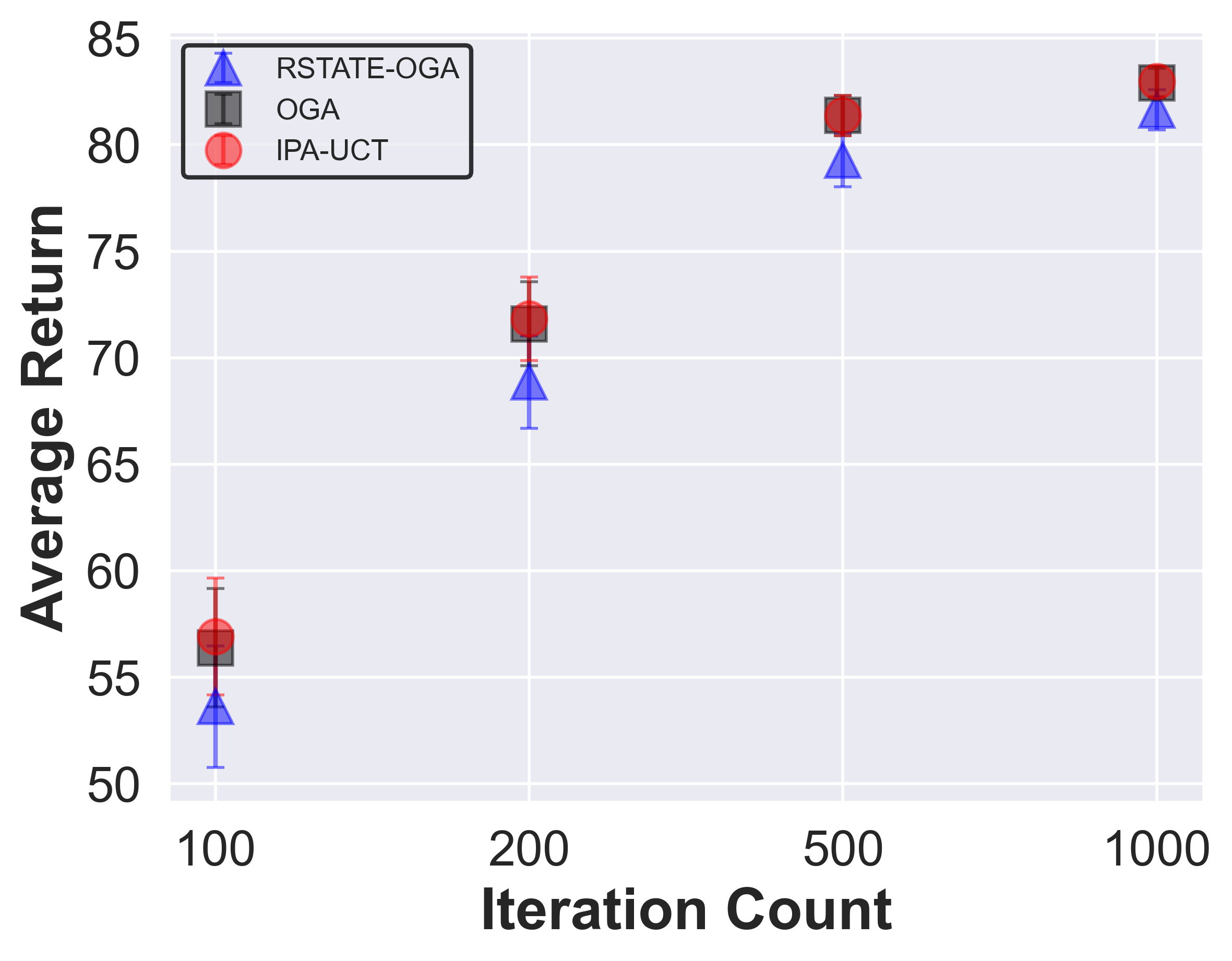}
\caption*{(o) Triangle Tireworld}
\end{minipage}

\label{fig:ipa:optimized_singleplayer}
\end{figure}

\begin{table}[H]
\caption{A list of the environment-specific $\varepsilon_{\text{a}}$ values that were used for the parameter-optimized experiments that used $(\varepsilon_{\text{a}},\varepsilon_{\text{t}})$-OGA. All single-agent domains that are not explicitly listed here use the default values $\varepsilon_{\text{a}} \in \{0,1,2,\infty\}$ and all two-player domains use the values $\varepsilon_{\text{a}} \in \{0,\infty\}$. The values were chosen to be equal to rewards (except 0 and $\infty$) that occur in these environments, to avoid the effect of different $\varepsilon_{\text{a}}$ inducing identical behavior.}
\label{tab:epsa_values}
\centering
\begin{tabular}{l l}
\hline
\textbf{Environment} & $\varepsilon_{\text{a}}$ values \\
\hline
Academic Advising & 0, $\infty$ \\
Cooperative Recon & 0, 0.5, 1.0, $\infty$ \\
Crossing Traffic & 0, $\infty$ \\
Manufacturer & 0, 10, 20, $\infty$ \\
Skill Teaching & 0, 2, 3, $\infty$ \\
Tamarisk & 0, 0.5, 1.0, $\infty$ \\
Default (Single Agent) & 0, 1, 2, $\infty$ \\
Default (Multi Agent) & 0, $\infty$ \\
\hline
\end{tabular}
\end{table}

\subsection{Parameter-optimized performance split by $\lambda_{\text{p}}$ values}

\begin{figure}[H]
\centering

\begin{minipage}{0.3\textwidth}
\centering
\includegraphics[width=\linewidth]{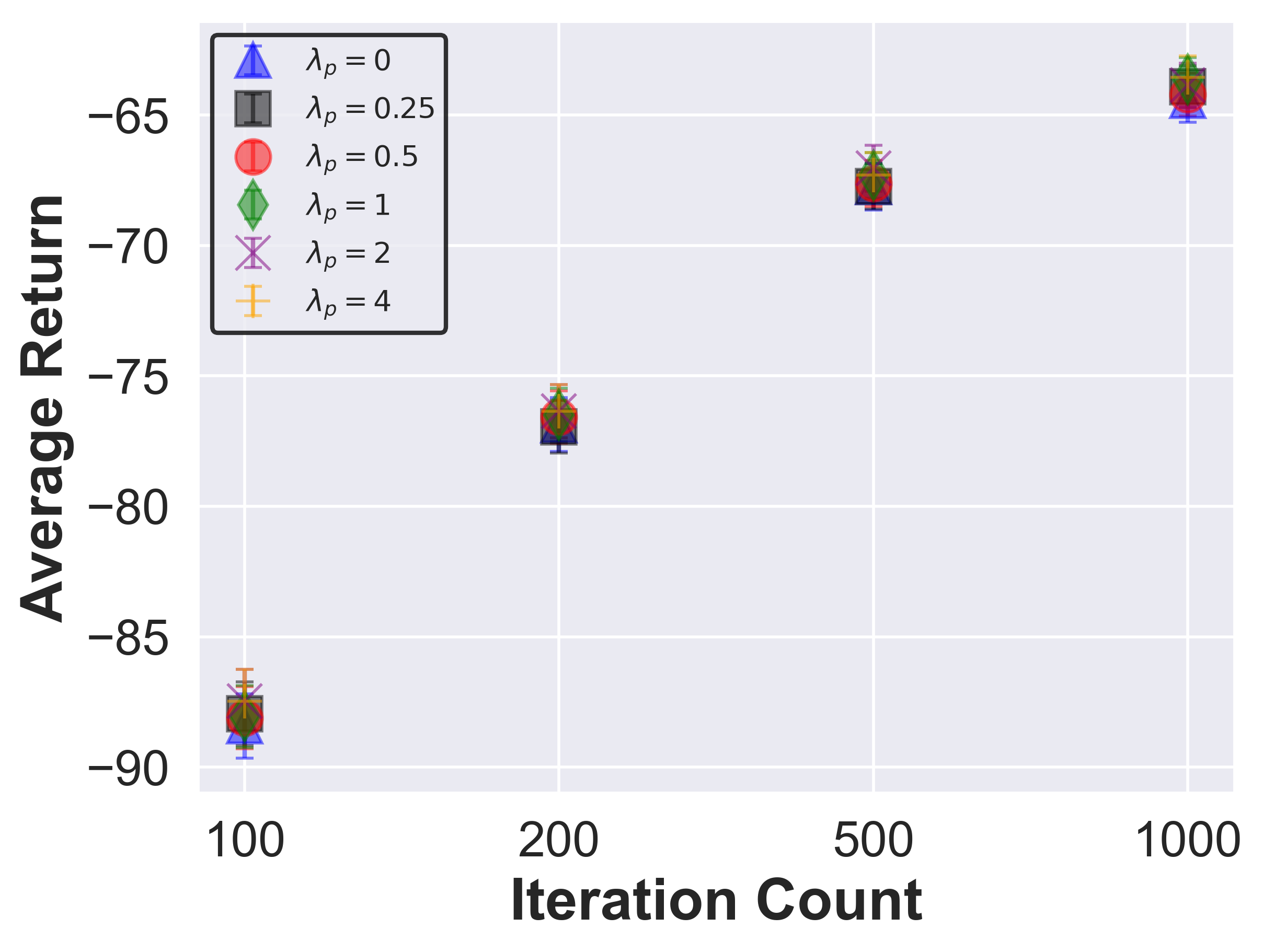}
\caption*{(a) Academic Advising}
\end{minipage}
\hfill
\begin{minipage}{0.3\textwidth}
\centering
\includegraphics[width=\linewidth]{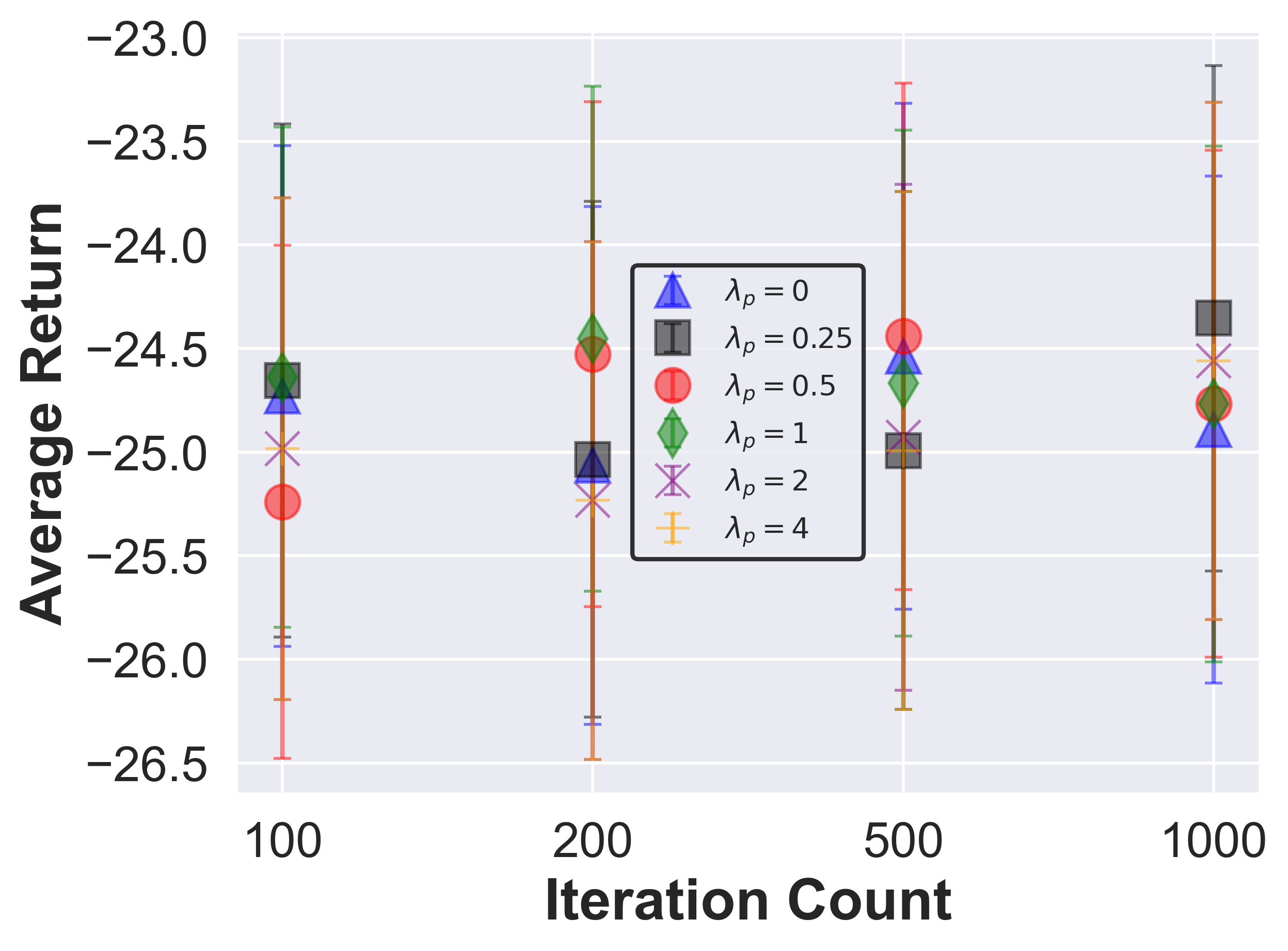}
\caption*{(b) Crossing Traffic}
\end{minipage}
\hfill
\begin{minipage}{0.3\textwidth}
\centering
\includegraphics[width=\linewidth]{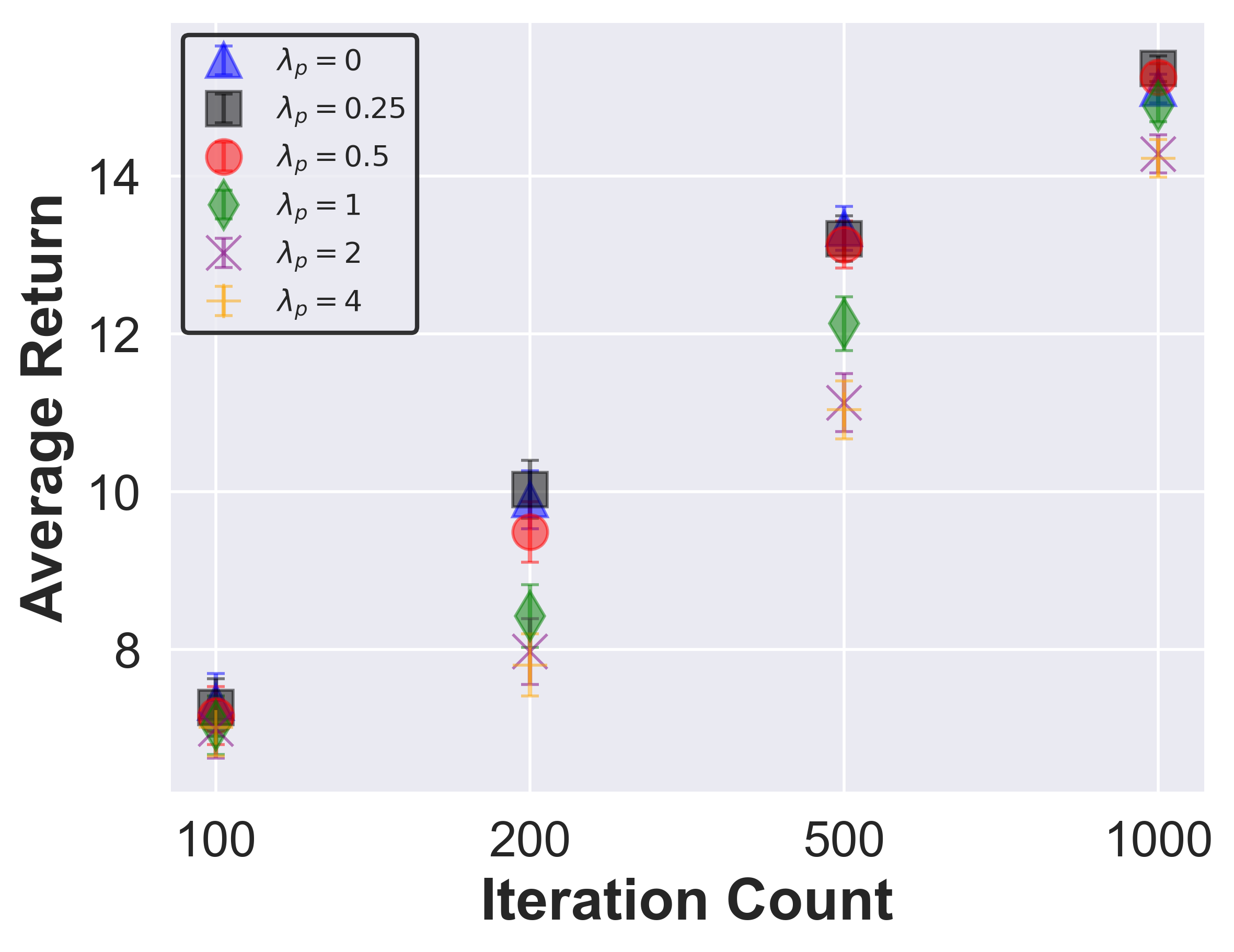}
\caption*{(c) Cooperative Recon}
\end{minipage}
\hfill
\begin{minipage}{0.3\textwidth}
\centering
\includegraphics[width=\linewidth]{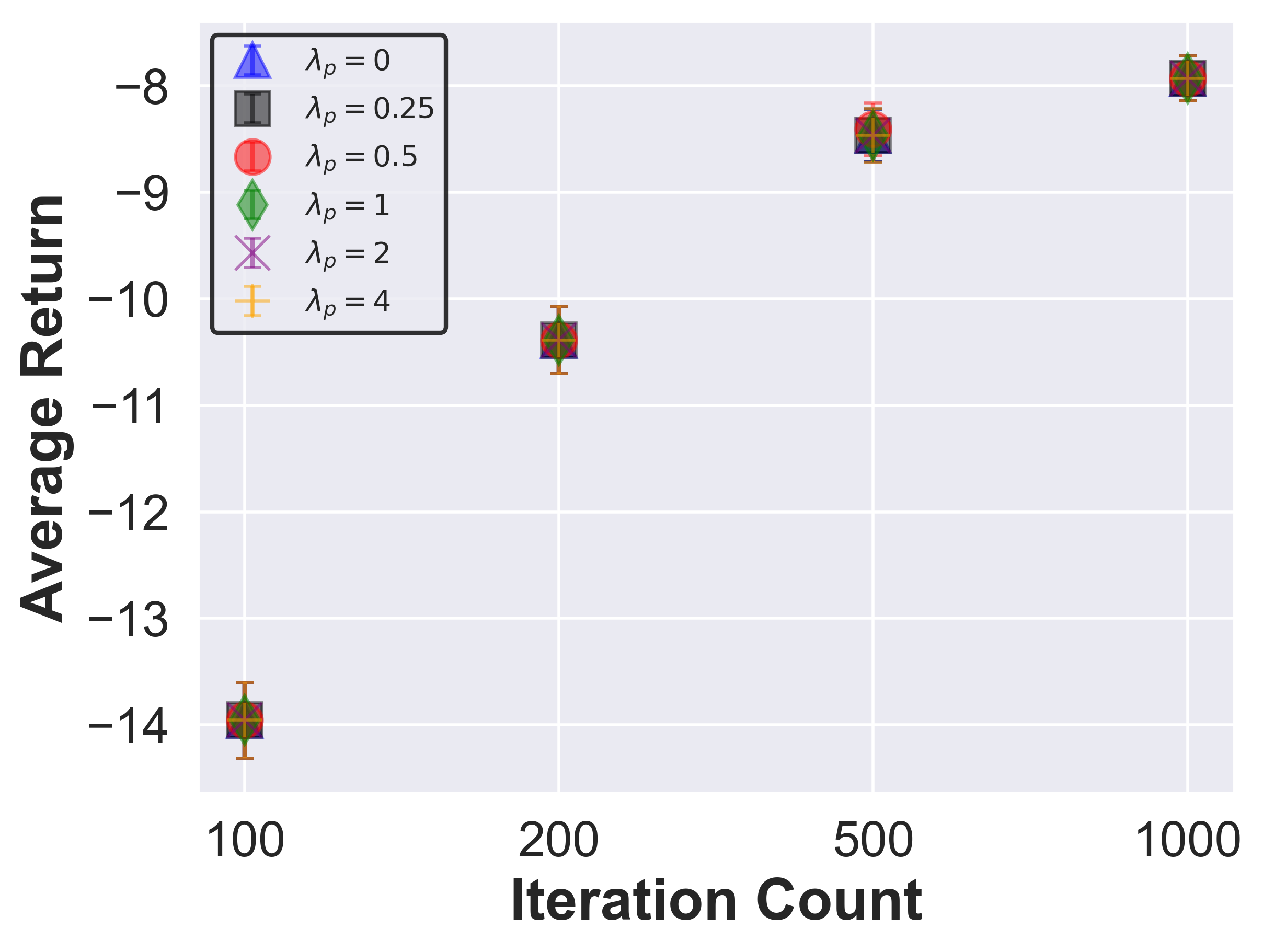}
\caption*{(d) Earth Observation}
\end{minipage}
\hfill
\begin{minipage}{0.3\textwidth}
\centering
\includegraphics[width=\linewidth]{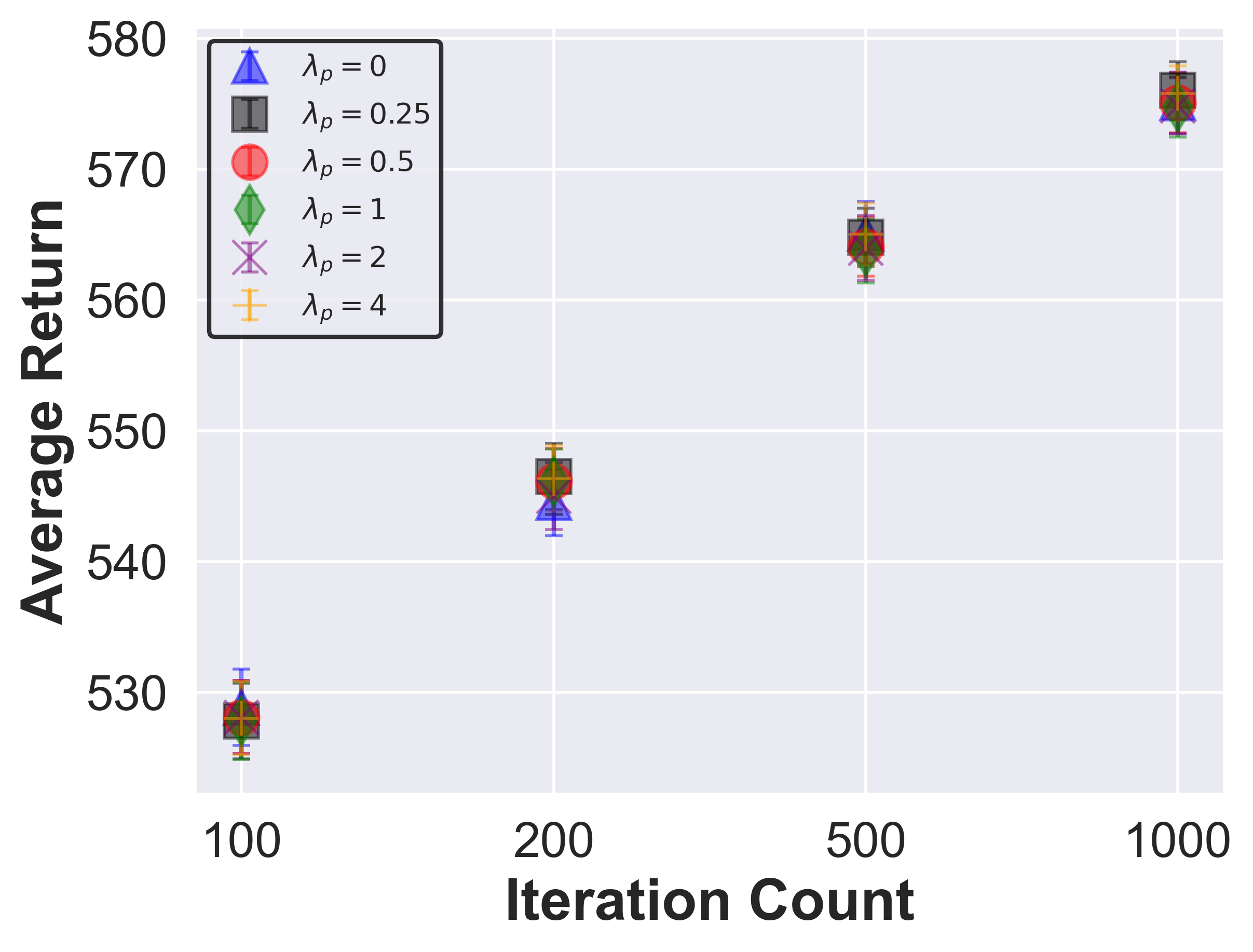}
\caption*{(e) Game of Life}
\end{minipage}
\hfill
\begin{minipage}{0.3\textwidth}
\centering
\includegraphics[width=\linewidth]{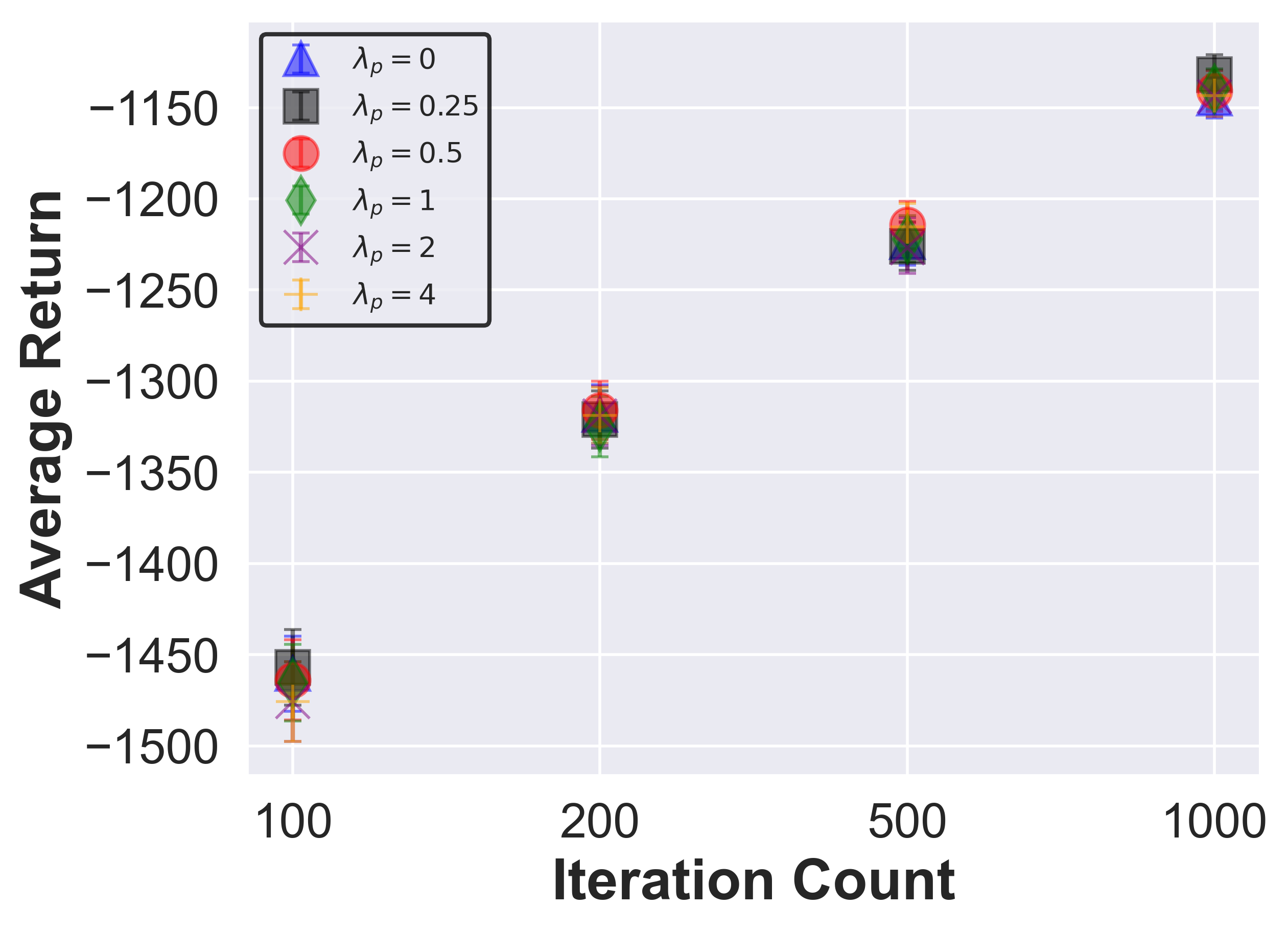}
\caption*{(f) Manufacturer}
\end{minipage}
\hfill
\begin{minipage}{0.3\textwidth}
\centering
\includegraphics[width=\linewidth]{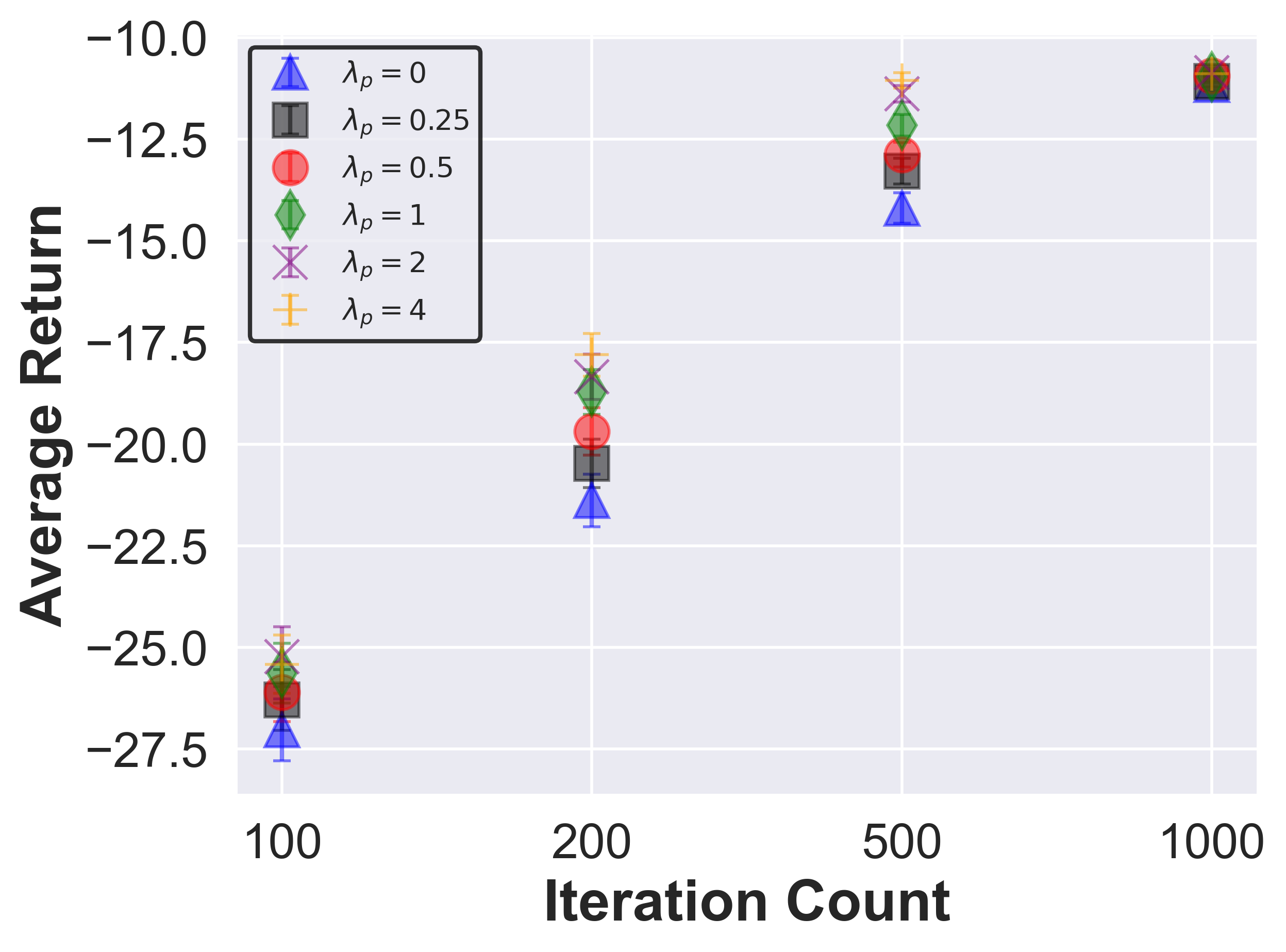}
\caption*{(g) Navigation}
\end{minipage}
\hfill
\begin{minipage}{0.3\textwidth}
\centering
\includegraphics[width=\linewidth]{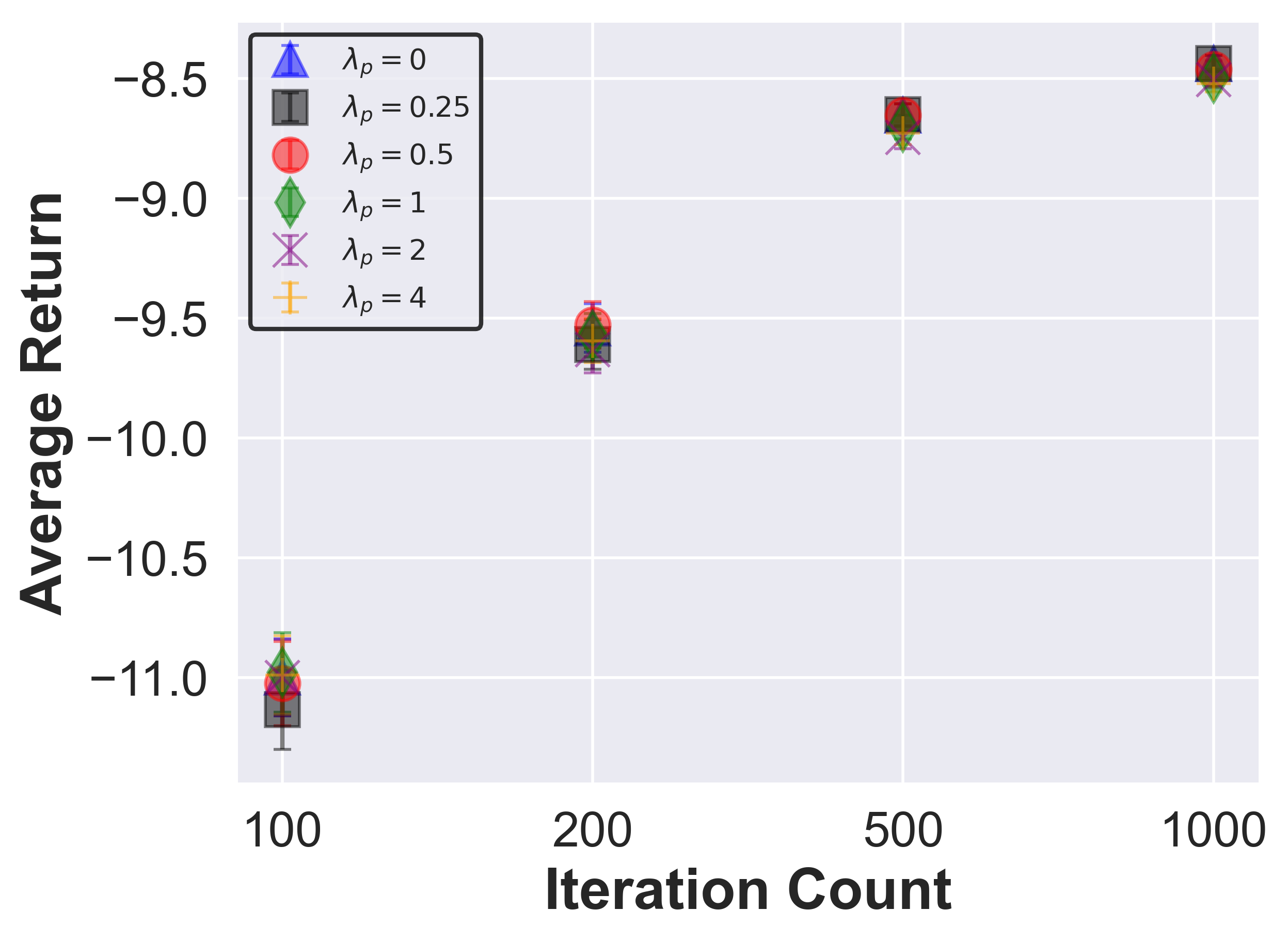}
\caption*{(h) Racetrack}
\end{minipage}
\hfill
\begin{minipage}{0.3\textwidth}
\centering
\includegraphics[width=\linewidth]{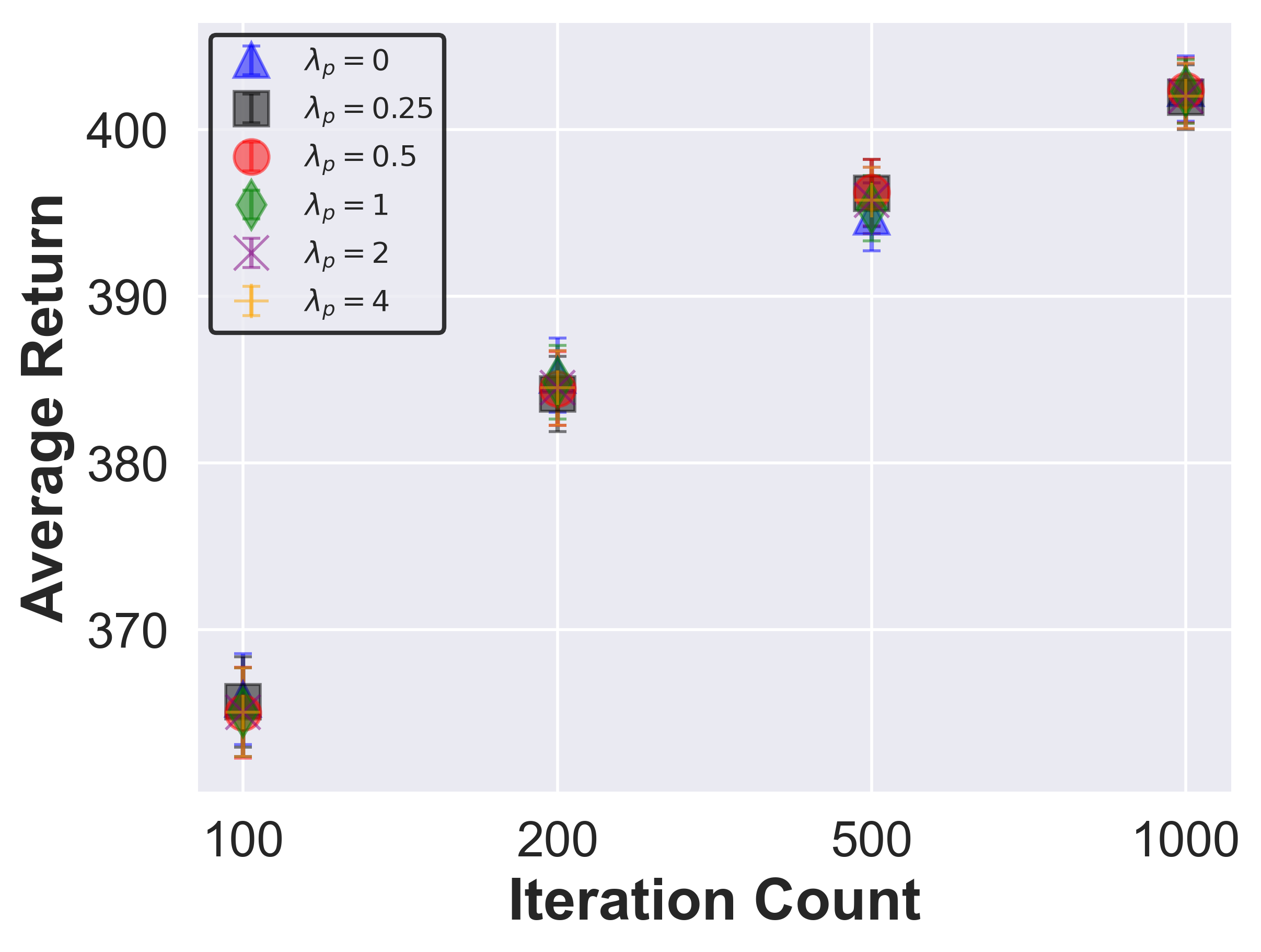}
\caption*{(i) SysAdmin}
\end{minipage}
\hfill
\begin{minipage}{0.3\textwidth}
\centering
\includegraphics[width=\linewidth]{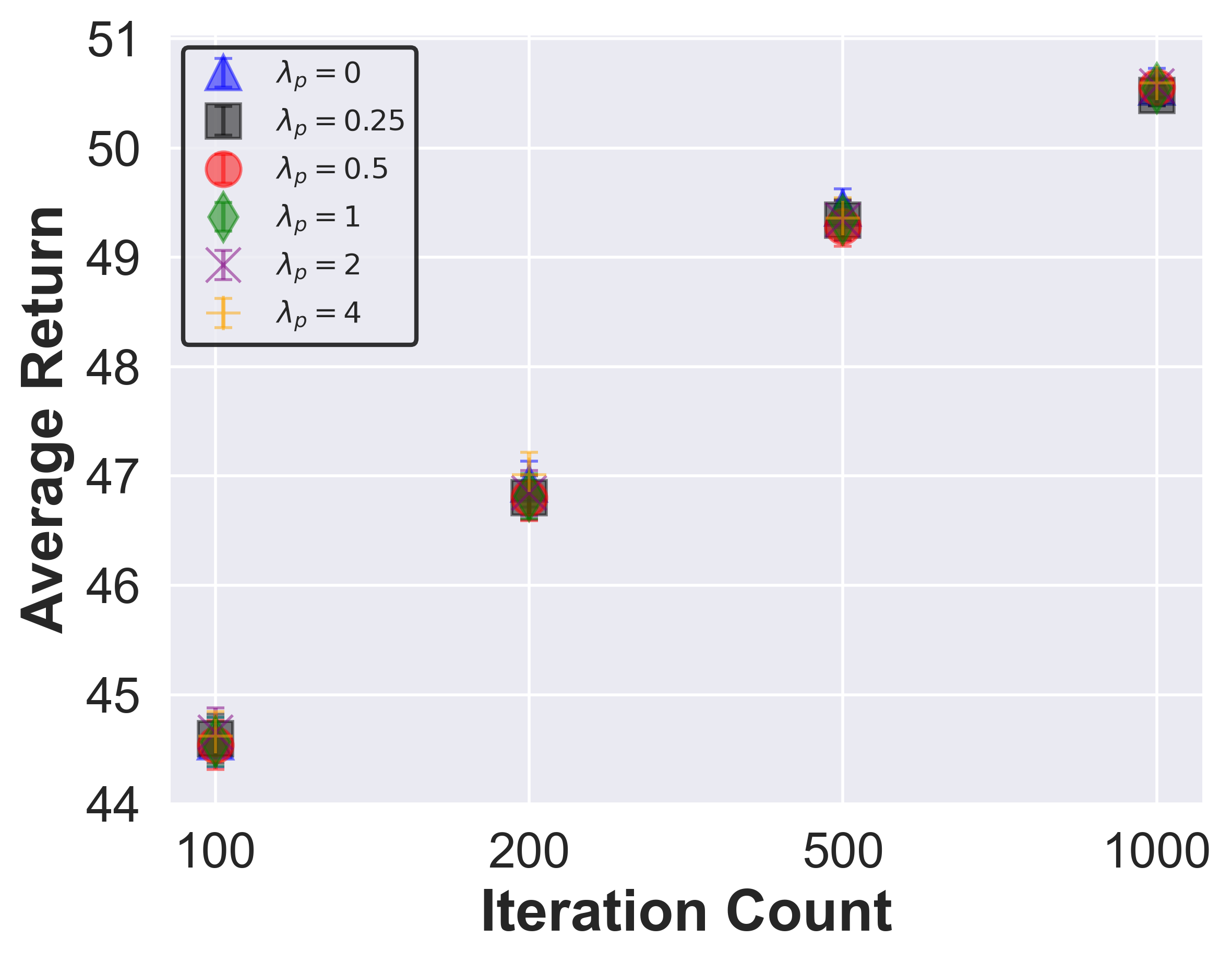}
\caption*{(j) Saving}
\end{minipage}
\hfill
\begin{minipage}{0.3\textwidth}
\centering
\includegraphics[width=\linewidth]{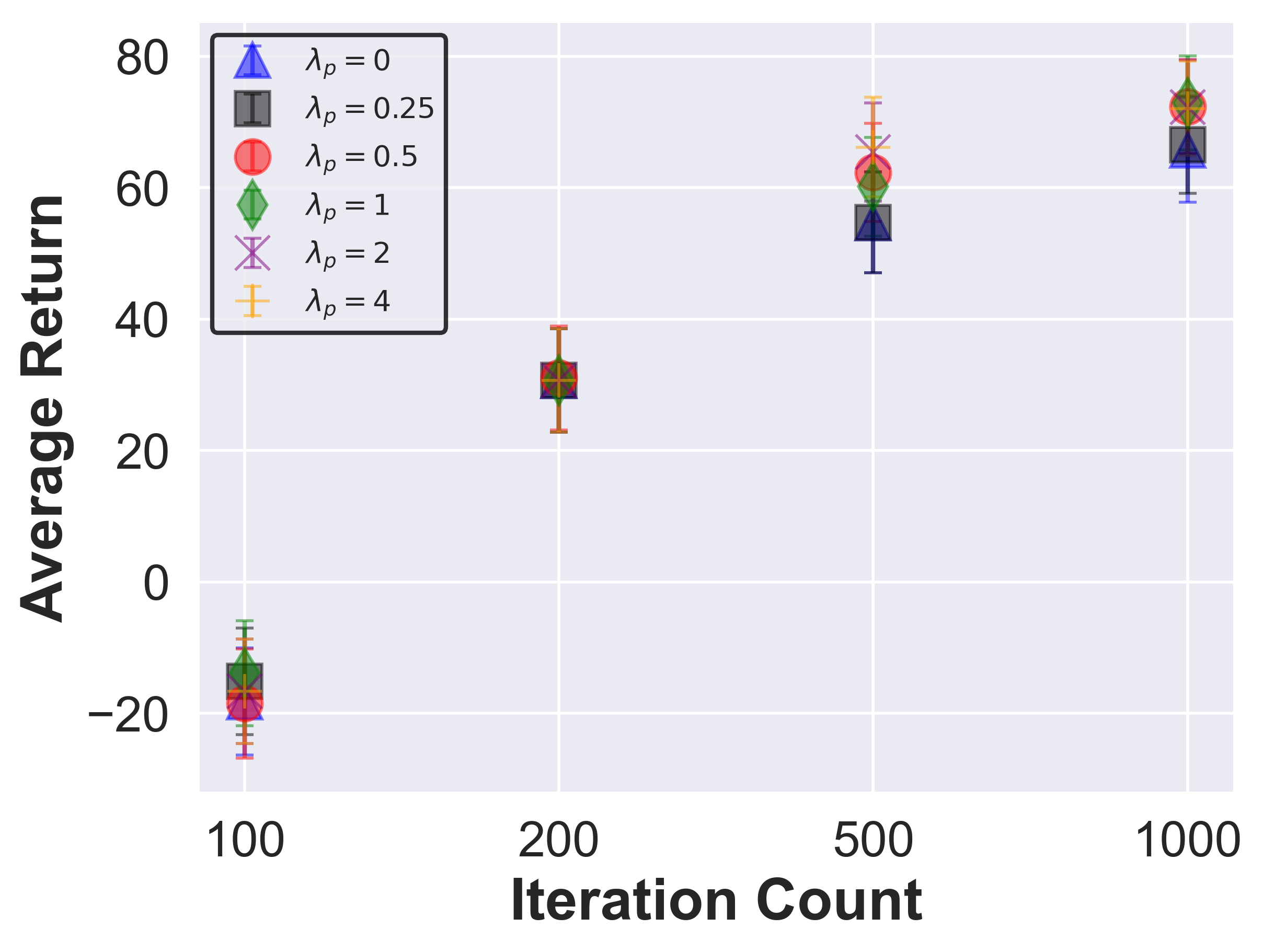}
\caption*{(k) Skill Teaching}
\end{minipage}
\hfill
\begin{minipage}{0.3\textwidth}
\centering
\includegraphics[width=\linewidth]{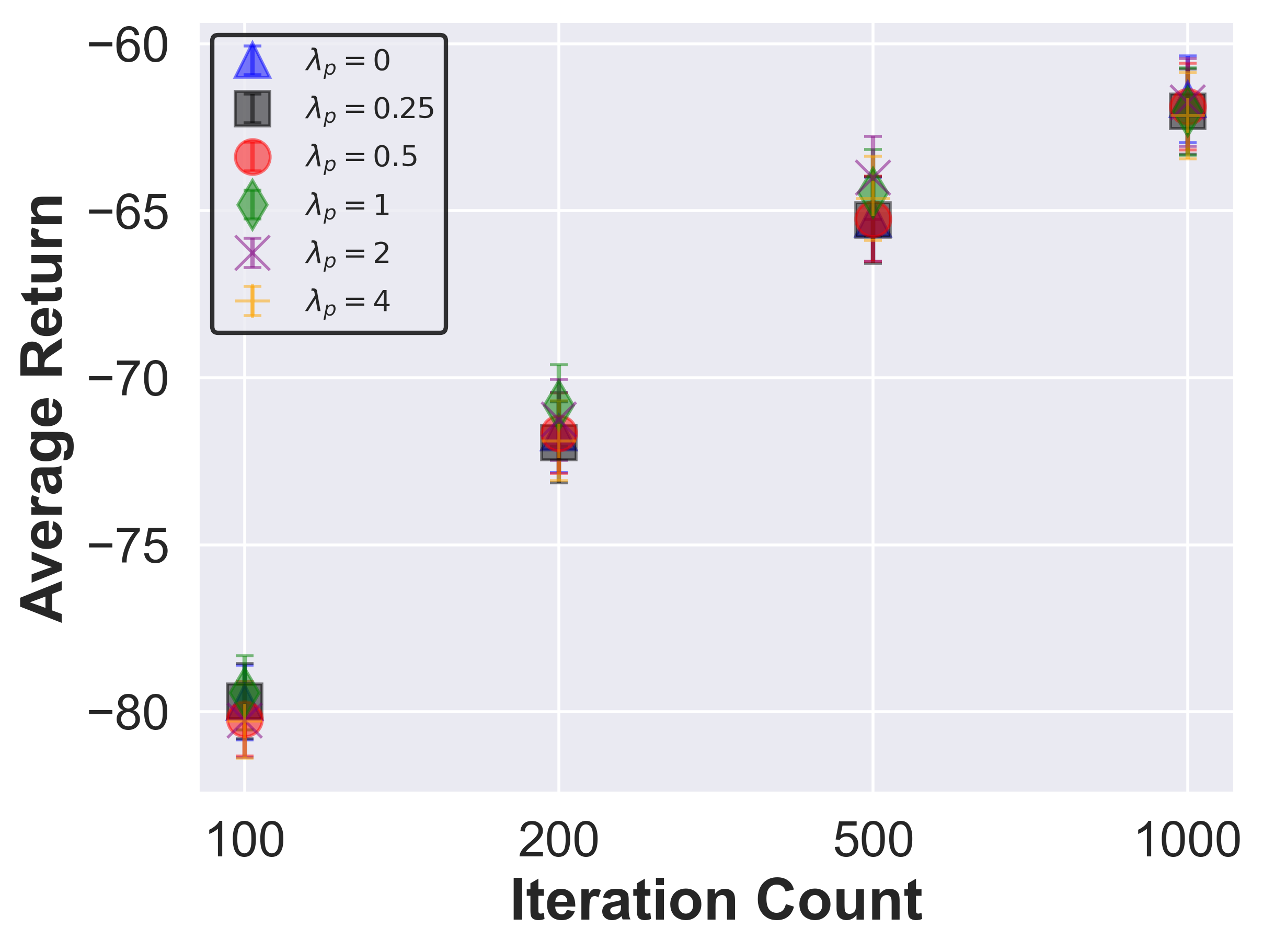}
\caption*{(l) Sailing Wind}
\end{minipage}
\hfill
\begin{minipage}{0.3\textwidth}
\centering
\includegraphics[width=\linewidth]{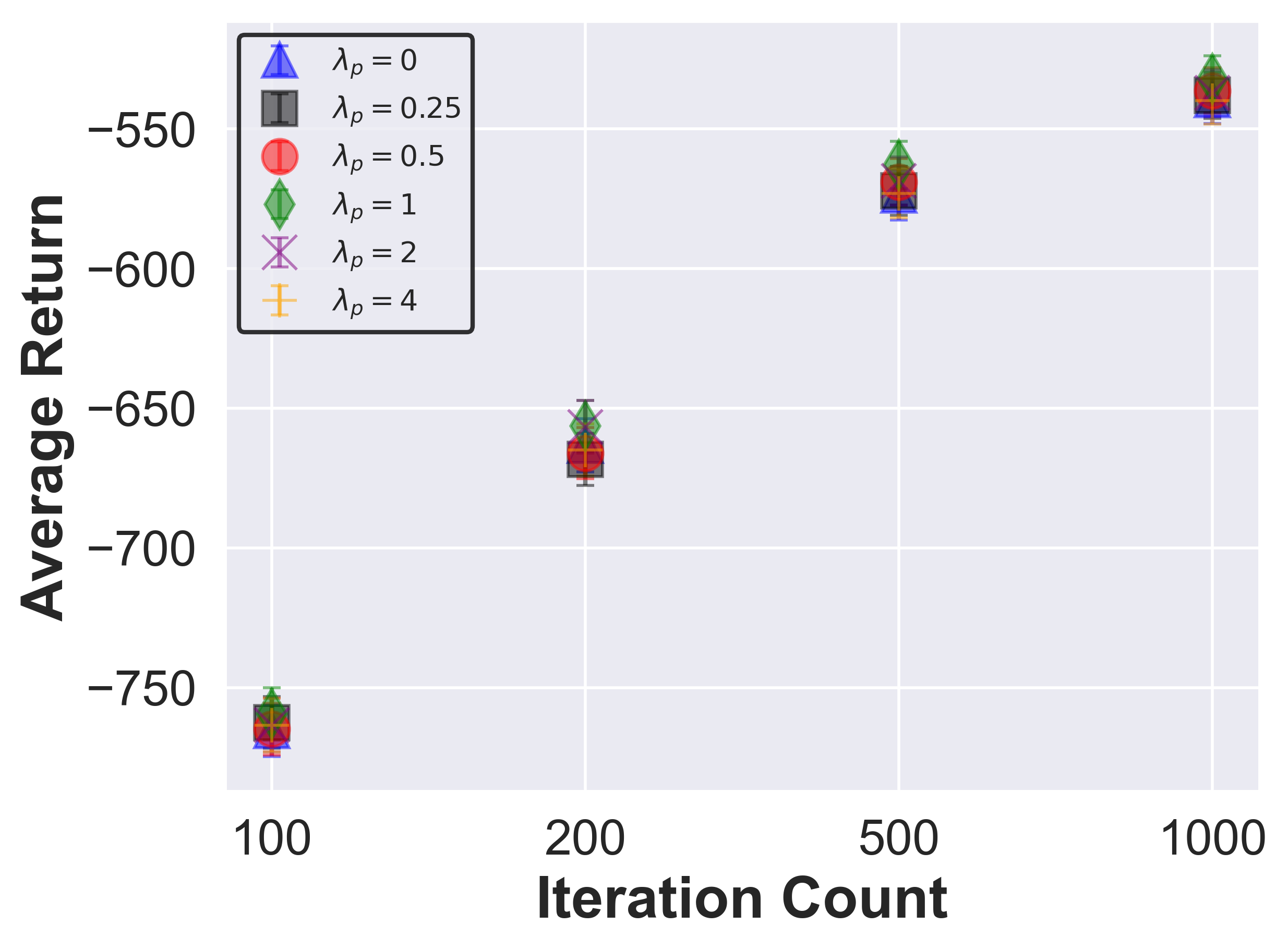}
\caption*{(m) Tamarisk}
\end{minipage}
\hfill
\begin{minipage}{0.3\textwidth}
\centering
\includegraphics[width=\linewidth]{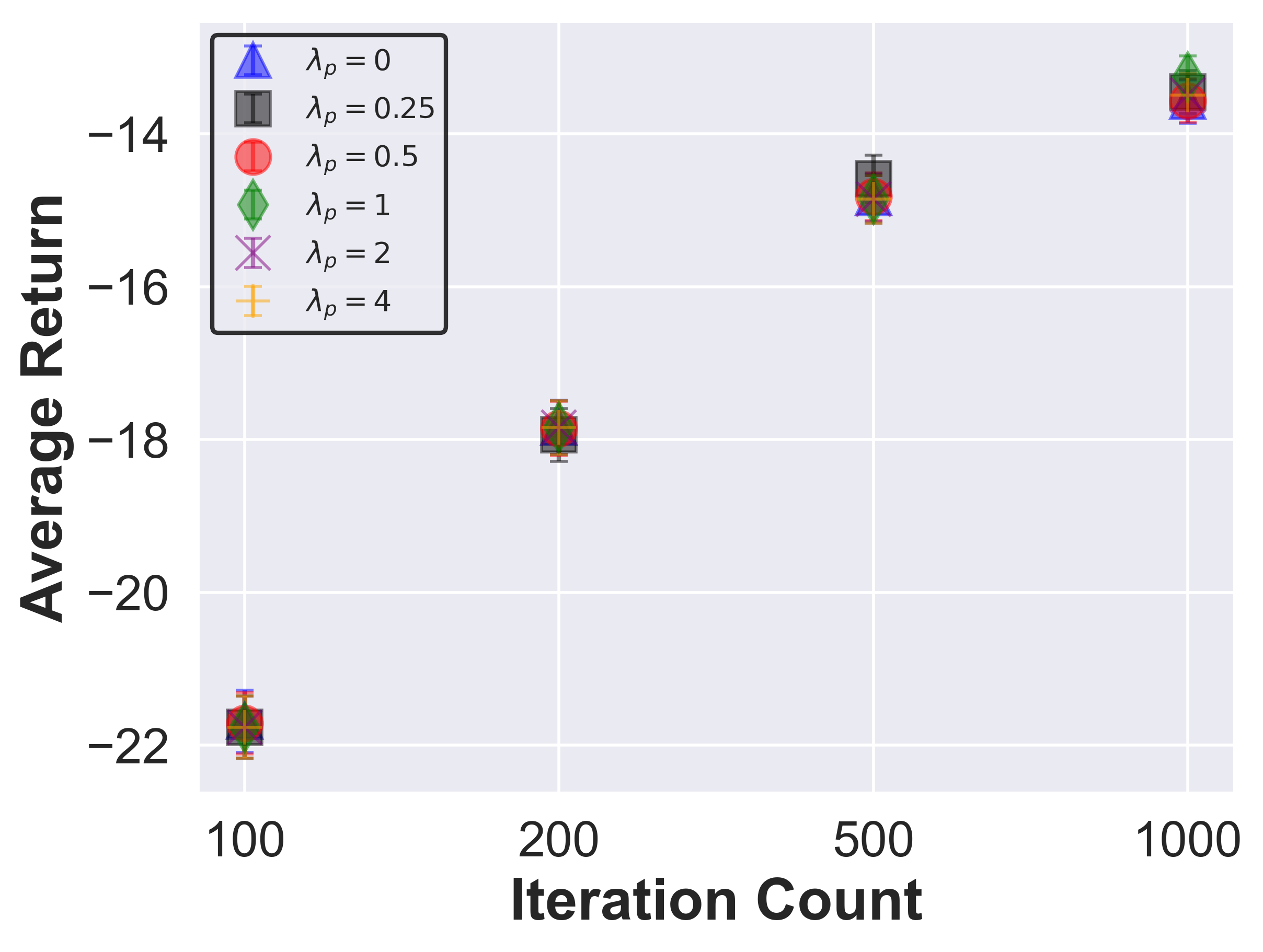}
\caption*{(n) Traffic}
\hfill
\end{minipage}
\begin{minipage}{0.3\textwidth}
\centering
\includegraphics[width=\linewidth]{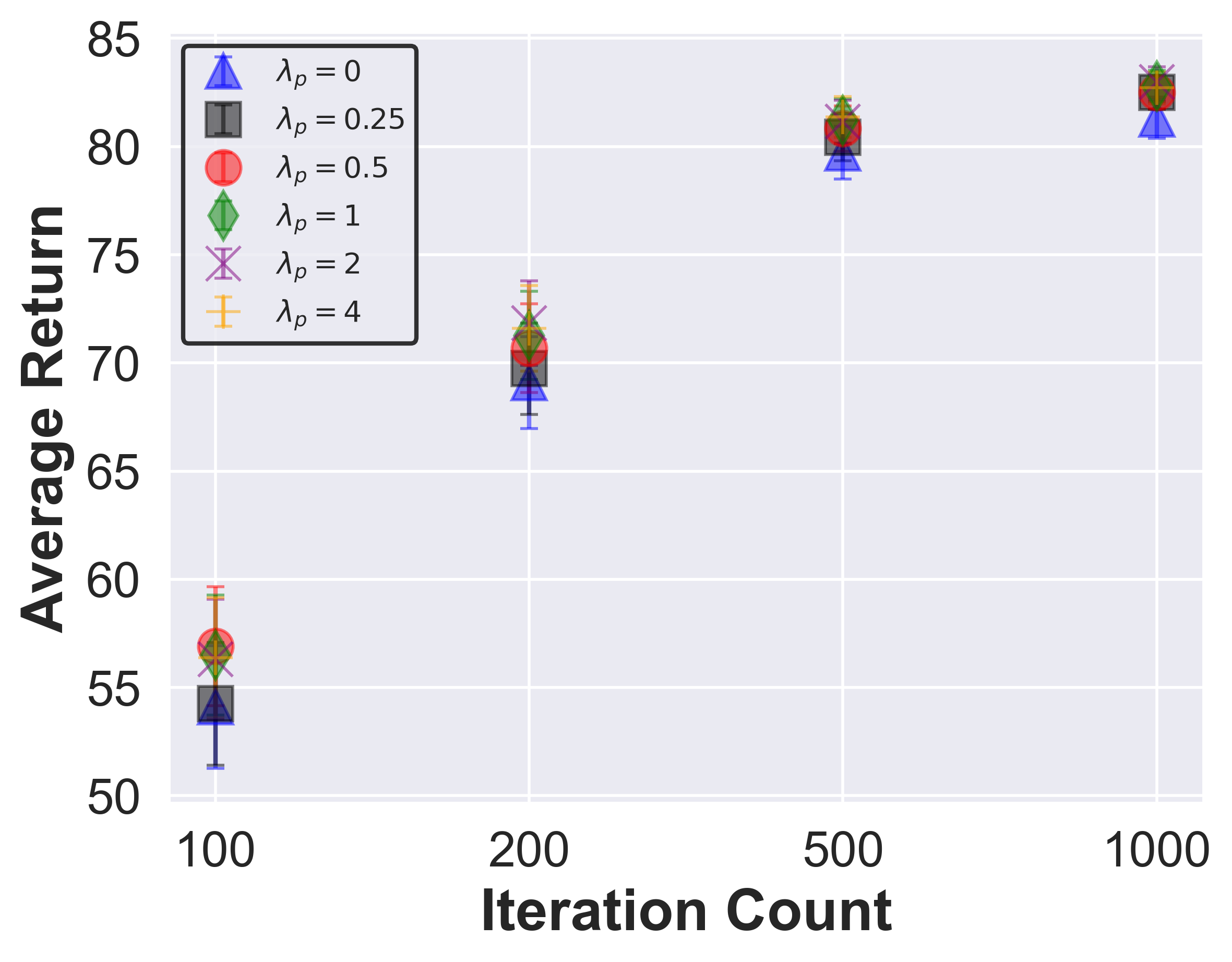}
\caption*{(o) Triangle Tireworld}
\end{minipage}

\label{fig:ipa:optimized_filter}
\end{figure}

\begin{figure}[H]
\centering

\begin{minipage}{0.3\textwidth}
\centering
\includegraphics[width=\linewidth]{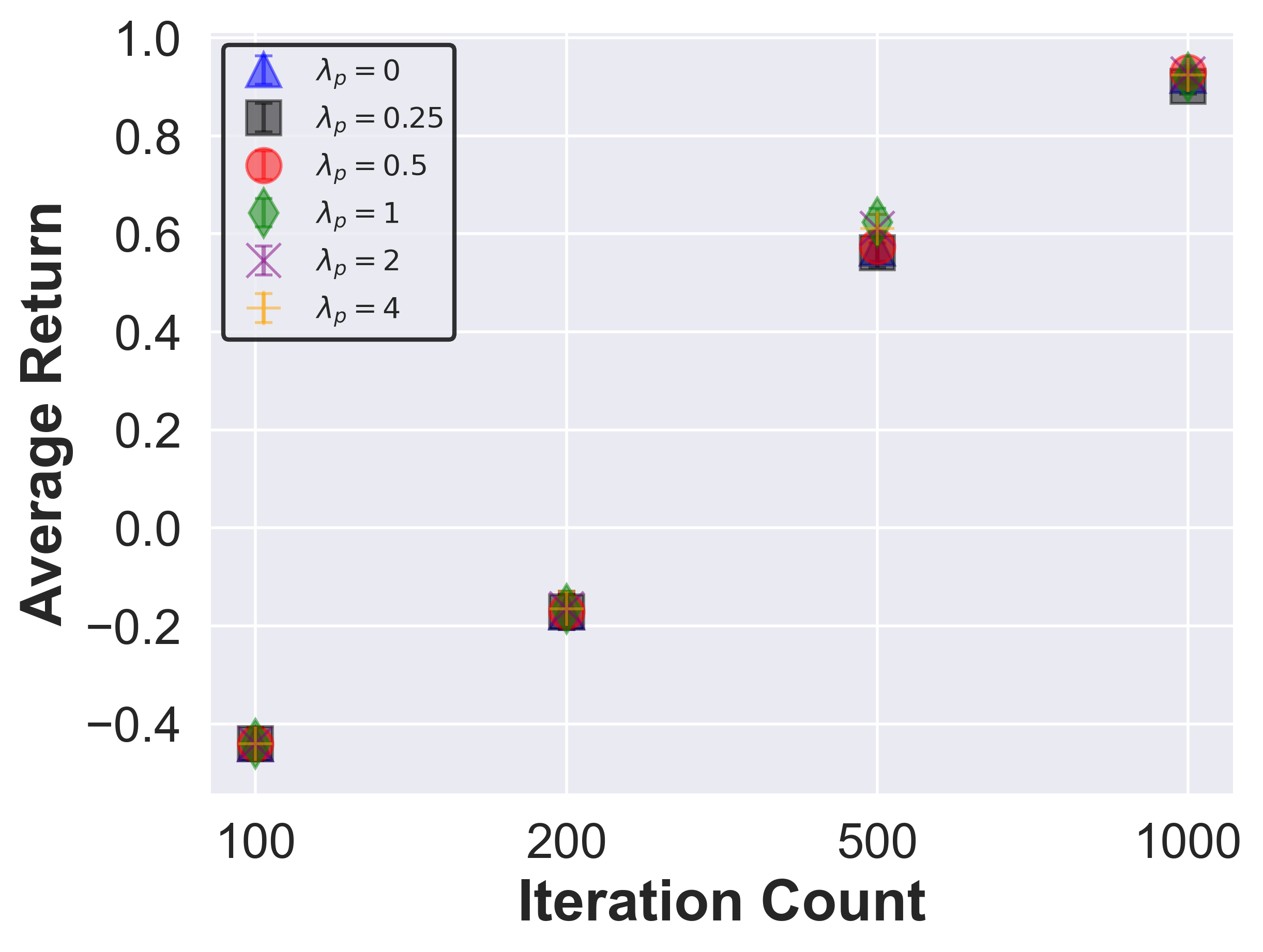}
\caption*{(p) Chess}
\end{minipage}
\hfill
\begin{minipage}{0.3\textwidth}
\centering
\includegraphics[width=\linewidth]{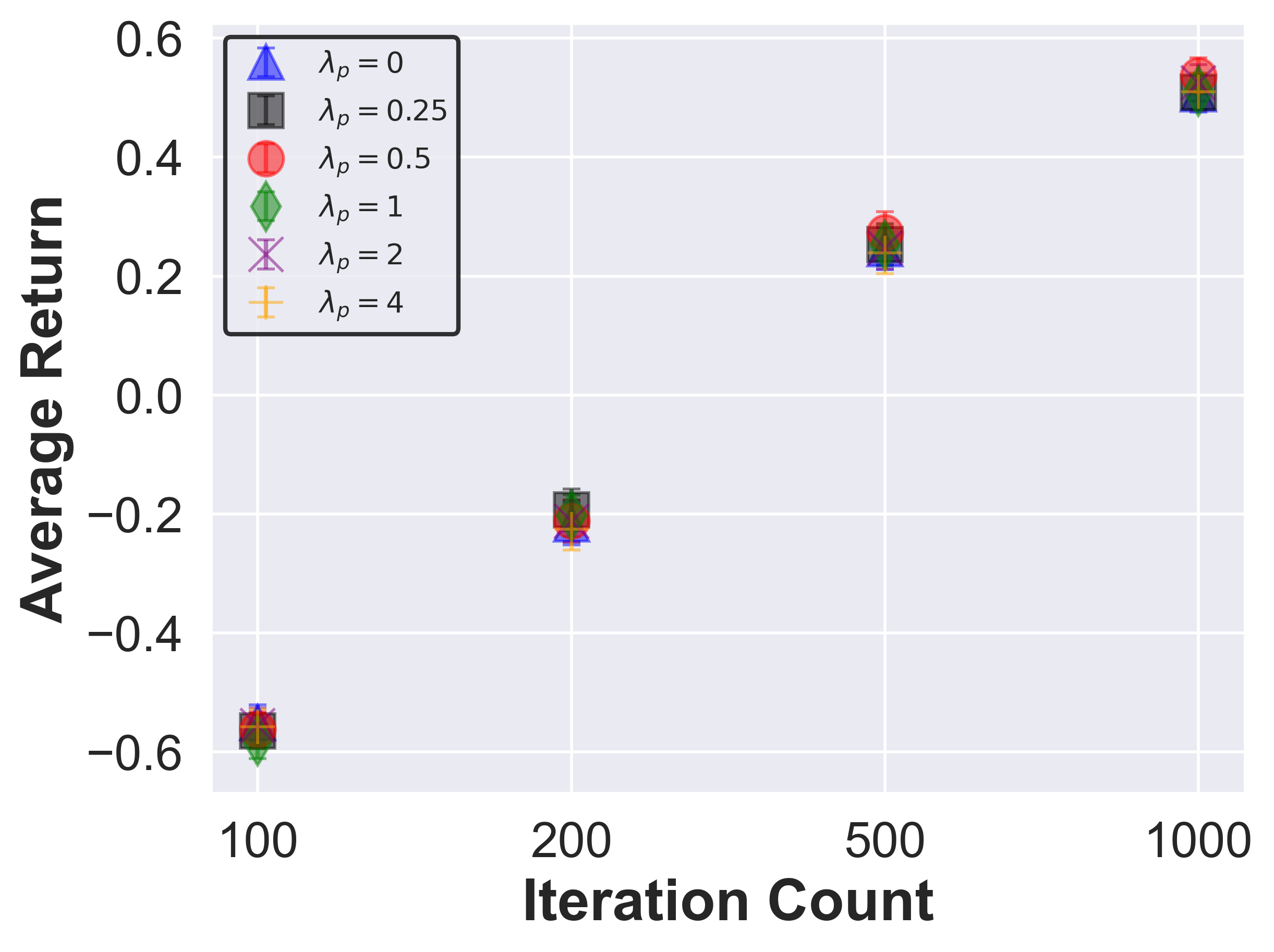}
\caption*{(q) Connect 4}
\end{minipage}
\hfill
\begin{minipage}{0.3\textwidth}
\centering
\includegraphics[width=\linewidth]{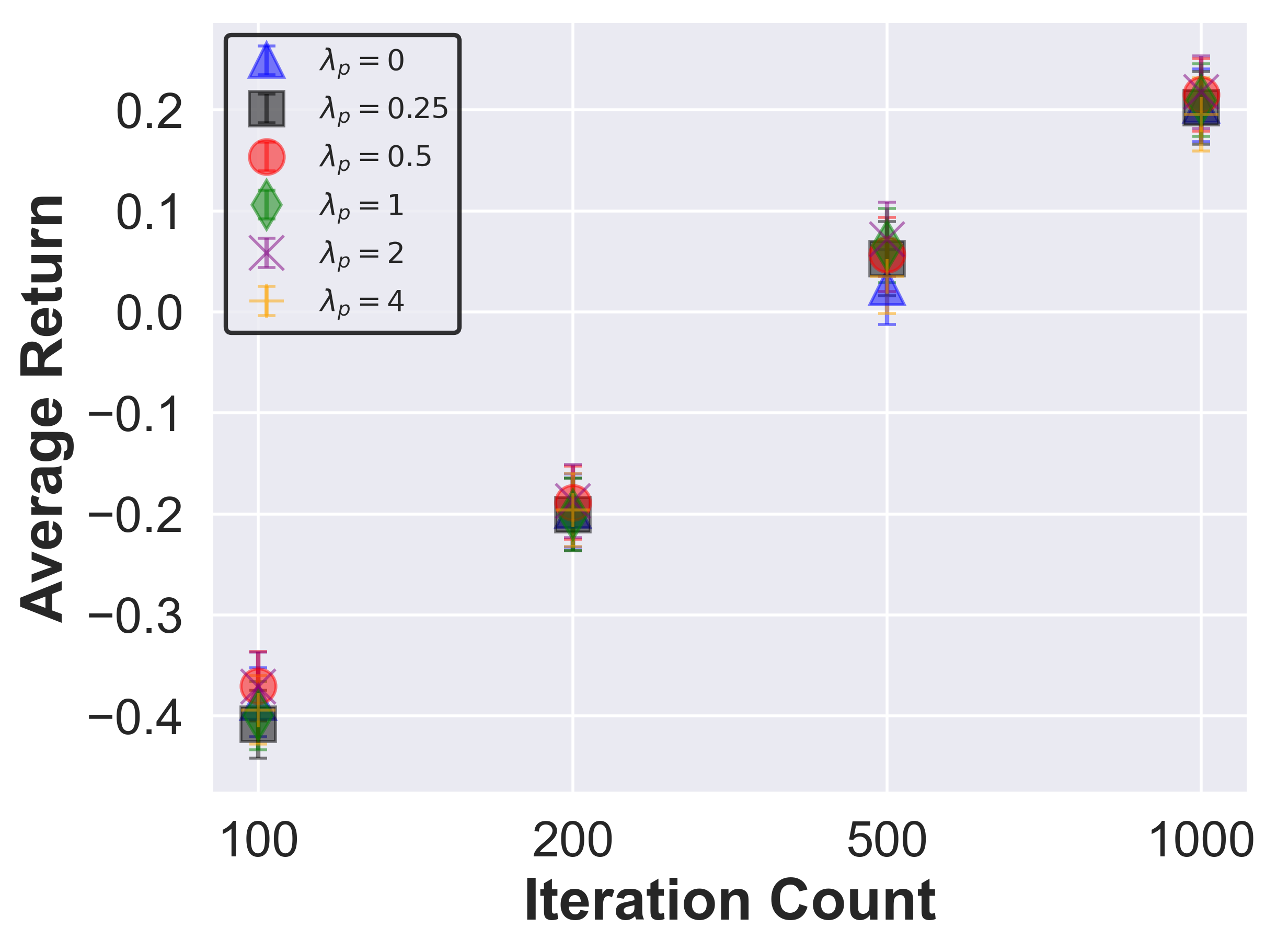}
\caption*{(r) Constrictor}
\end{minipage}
\hfill
\begin{minipage}{0.3\textwidth}
\centering
\includegraphics[width=\linewidth]{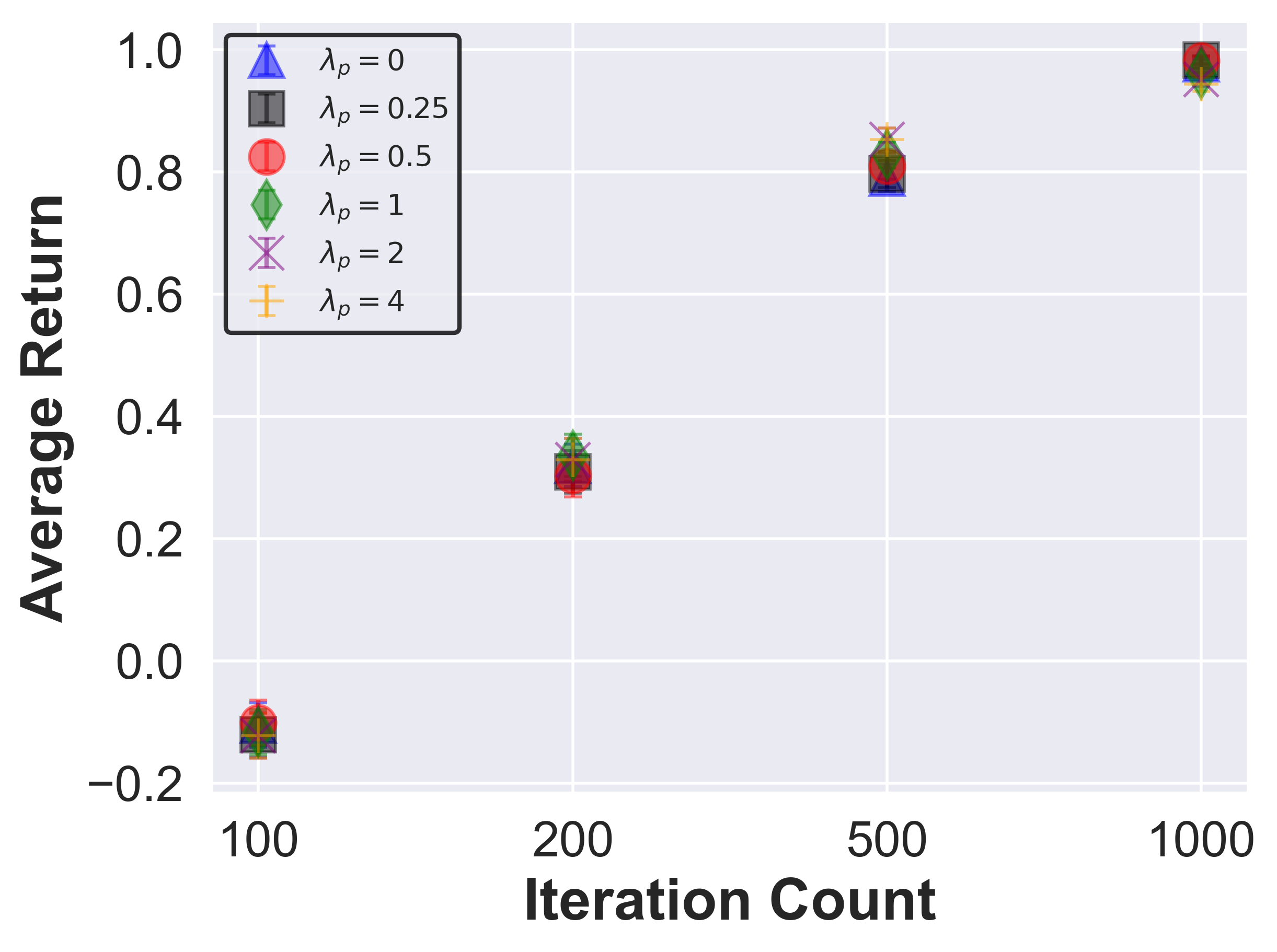}
\caption*{(s) Numbers Race}
\end{minipage}
\hfill
\begin{minipage}{0.3\textwidth}
\centering
\includegraphics[width=\linewidth]{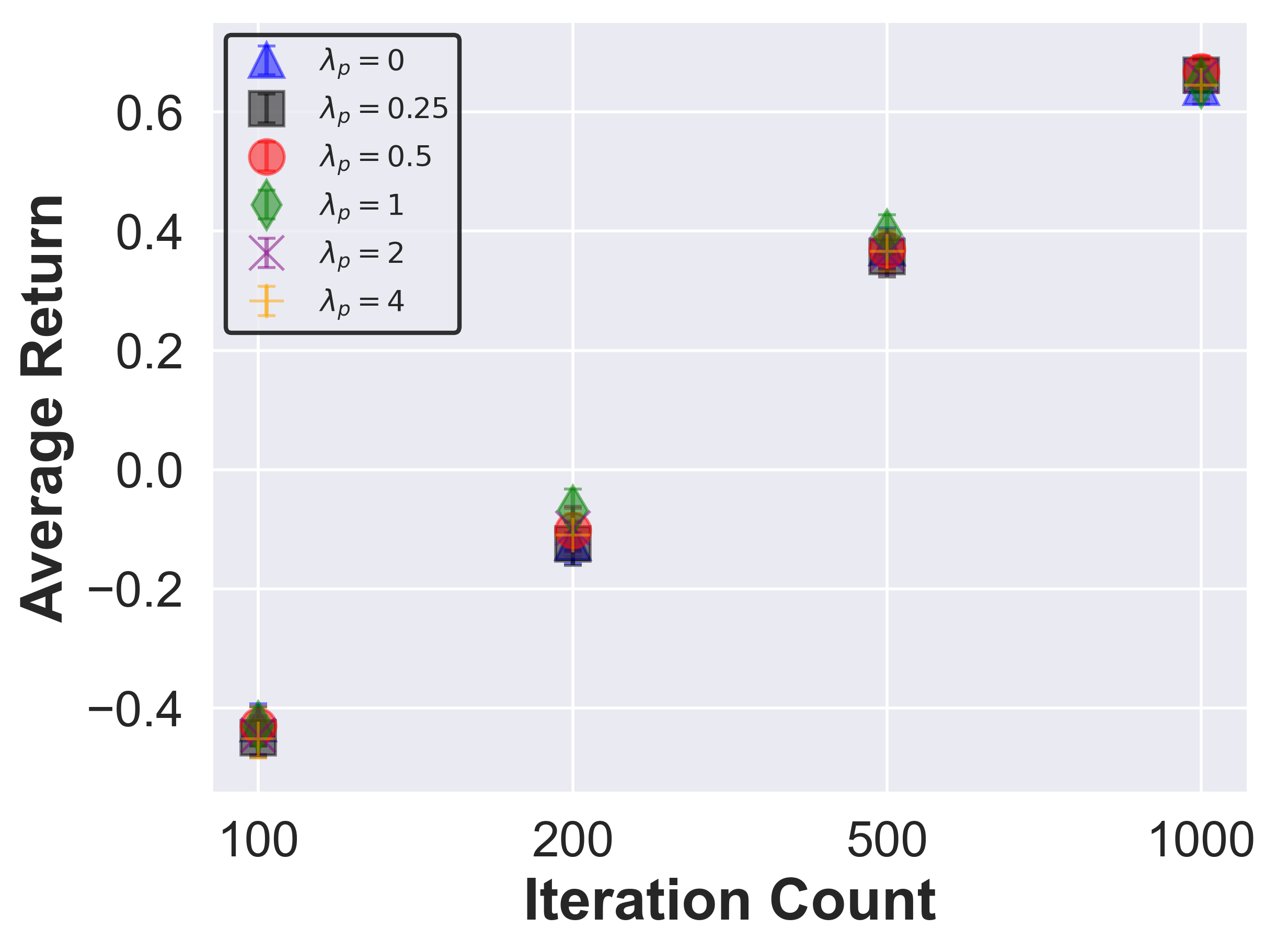}
\caption*{(t) Othello}
\end{minipage}
\hfill
\begin{minipage}{0.3\textwidth}
\centering
\includegraphics[width=\linewidth]{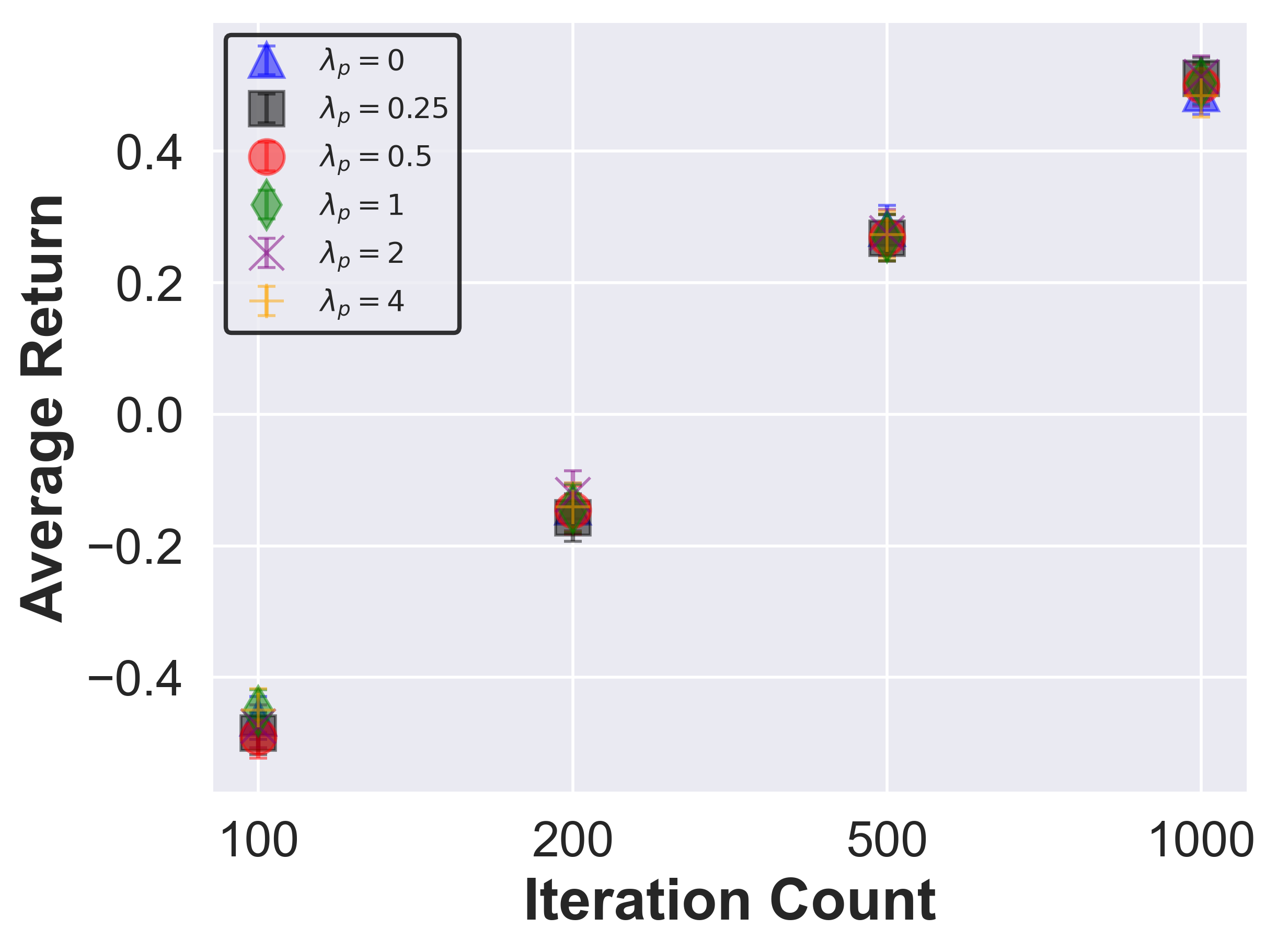}
\caption*{(u) Pylos}
\end{minipage}
\hfill
\begin{minipage}{0.3\textwidth}
\centering
\includegraphics[width=\linewidth]{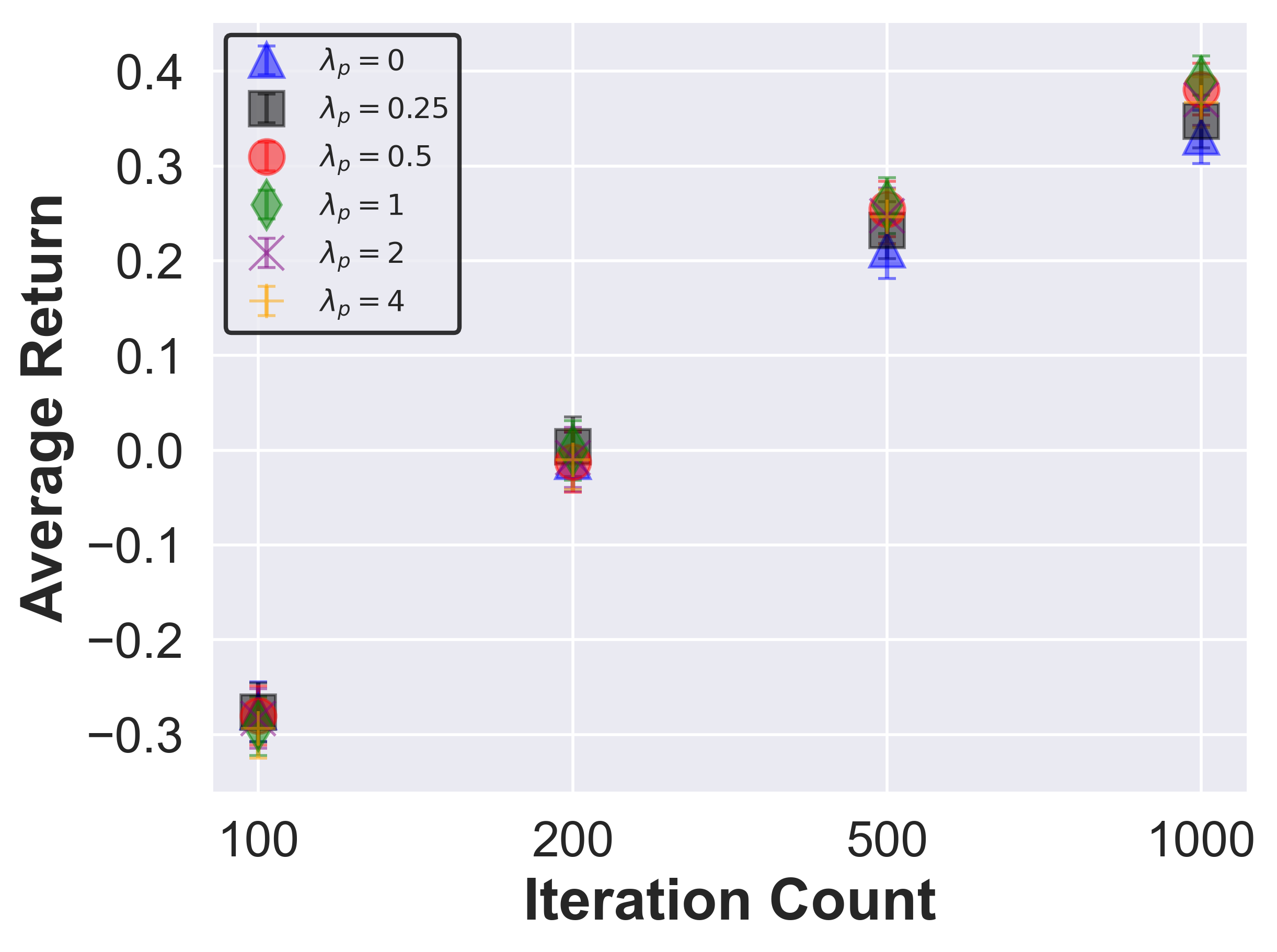}
\caption*{(v) Quarto}
\end{minipage}
\hfill
\begin{minipage}{0.3\textwidth}
\centering
\includegraphics[width=\linewidth]{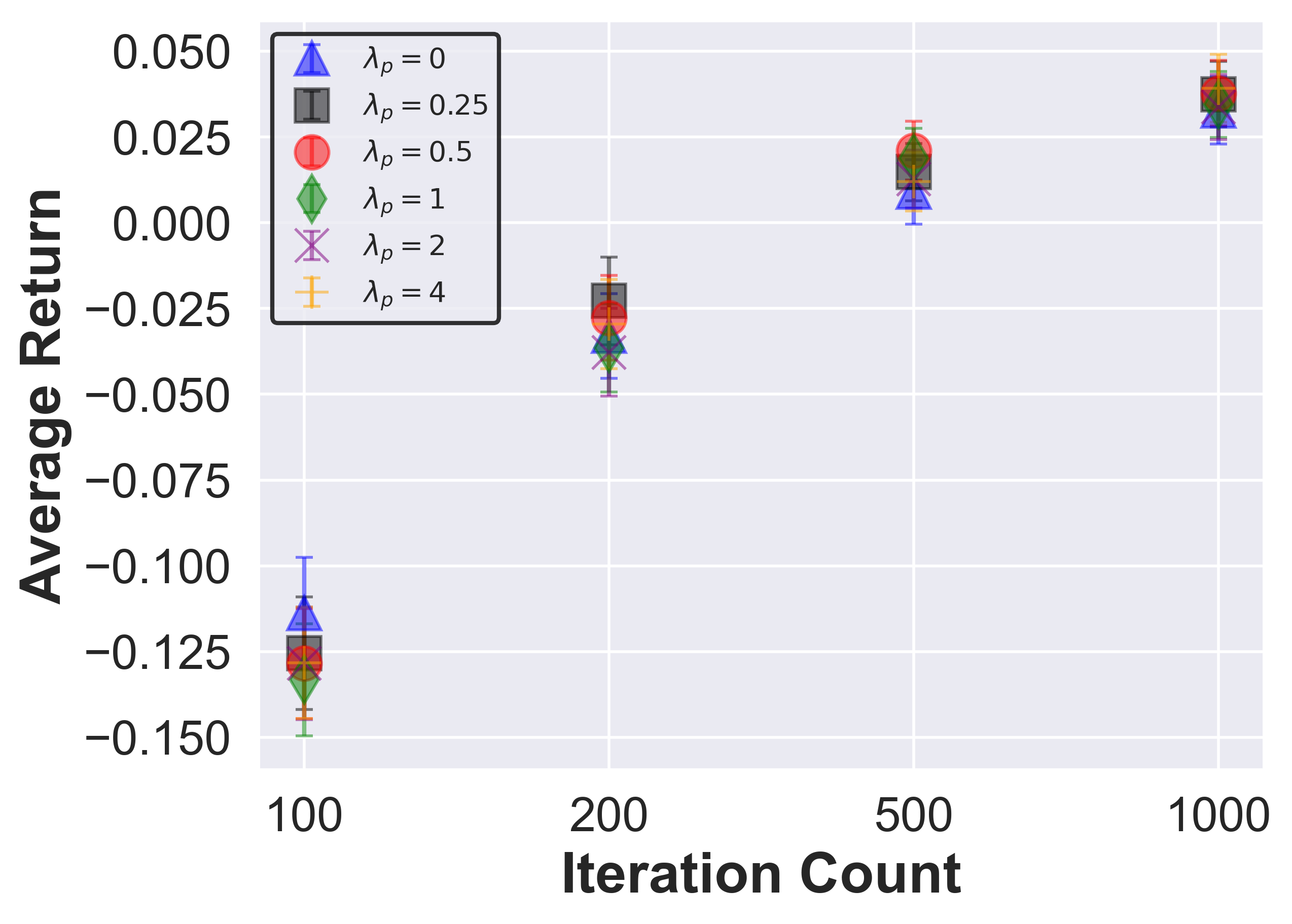}
\caption*{(w) Tic Tac Toe}
\end{minipage}
\hfill
\begin{minipage}{0.15\textwidth}
\centering
\includegraphics[width=\linewidth]{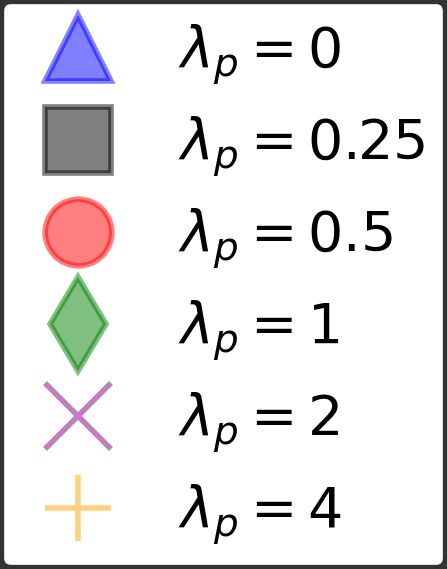}
\caption*{Legend}
\end{minipage}

\caption{The parameter-optimized performance graphs for all problems in dependence on the iteration budget and $\lambda_{\text{p}}$. The parameters over which the agents were optimized are identical to those in Section \ref{sec:experiments} except that we used hand picked $\varepsilon_{\text{a}}$ values for each environment which are listed in Tab.~\ref{tab:epsa_values}.}
\label{fig:ipa:optimized_filter_mp}
\end{figure}

\newpage
\subsection{Definition of the relative improvement and pairings score}
\label{subsec:scors_defs}

In the main experimental section, we evaluated IPA-UCT with respect to the relative improvement and pairings score, which are formalized here. These scores are Borda-like rankings to score the ability of an agent to perform well in a large number of tasks and these were already used in \cite{aupo,intra}. The relative improvement score quantifies the average improvement percentages over other agents and the pairings score simply quantifies the number of tasks some agent outperformed another agent but does not take the magnitude of the improvement into consideration.
 
\noindent\textbf{Definition:}
Let $\{\pi_1,\dots,\pi_n\}$ be $n$ agents (e.g., concrete parameter settings) where each agent was evaluated on $m$ tasks (e.g. a given MCTS iteration budget and an environment) where $p_{i,k} \in \mathbb{R}$ denotes the performance of agent $\pi_i$ on the $k$-th task.  
    The \textit{pairings score matrix} $M^{\text{pairings}} \in \mathbb{R}^{n \times n}$ is defined as 
    \begin{equation}
        M^{\text{pairings}}_{i,j} =  \frac{1}{m-1}\sum\limits_{1 \leq k \leq m}  \text{sgn}(p_{i,k}-p_{j,k})
    \end{equation}
    where sgn is the signum function.
    The \textit{pairings score} $s^{\text{pairings}}_i, i \leq n$ is given by 
    \begin{equation}
        s^{\text{pairings}}_i = \frac{1}{n-1}\sum\limits_{1 \leq l \leq n, l \neq i} M^{\text{pairings}}_{i,l}.
    \end{equation}
    The \textit{relative improvement matrix} $M^{\text{rel}} \in \mathbb{R}^{n \times n}$ is defined as 
    \begin{equation}
        M^{\text{rel}}_{i,j} = \frac{1}{m-1} \sum\limits_{1 \leq k \leq m}  \frac{p_{i,k}-p_{j,k}}{\max(|p_{i,j}|,|p_{j,k}|)}
    \end{equation}
    and the \textit{relative improvement score} $s^{\text{rel}}_i, i \leq n$ is given by 
    \begin{equation}
        s^{\text{rel}}_i = \frac{1}{n-1}\sum\limits_{1 \leq l \leq n, l \neq i} M^{\text{rel}}_{i,l}.
    \end{equation}

\subsection{Runtime measurements}
Tab.~\ref{tab:runtimes} lists the average decision-making times for each environment of IPA-UCT compared to OGA-UCT for 100 and 2000 iterations on states sampled from a distribution induced by random walks. This shows that while UCT adds only a minor overhead, despite having to execute more UCB evaluations. In particular, we are using highly optimized environment implementations that could be the runtime bottleneck in more complex environments.

\begin{table}[H]
\centering

\caption{Average decision-making times of IPA-UCT versus OGA-UCT in milliseconds for 100 and 2000 iterations. This data was obtained using an Intel(R) Core(TM) i5-9600K CPU @ 3.70GHz. The data shows a median runtime overhead of $\approx$5\% for 100 iterations and $\approx$9\% for 2000 iterations.}
\label{tab:runtimes}
\scalebox{1.0}{
\begin{tabular}{l c c c c}
\hline
 Domain & IPA-UCT-100 & OGA-UCT-100 & IPA-UCT-2000 & OGA-UCT-2000 \\\hline

Academic Advising & 2.22 & 2.01 & 164.63 & 125.61 \\ \hline
Cooperative Recon & 4.14 & 3.91 & 267.31 & 232.49 \\ \hline
Crossing Traffic & 2.85 & 2.62 & 382.01 & 378.96\\ \hline
Connect4 & 1.77 & 1.69 & 112.21 & 98.94 \\ \hline
Chess & 18.01 & 18.40 & 454.55 & 421.35 \\ \hline
Constrictor & 4.96 & 4.71 & 347.41 & 316.53 \\ \hline
Earth Observation & 7.61 & 7.92 & 367.06 & 345.02 \\ \hline
Game of Life & 4.14 & 4.02 & 273.46 & 260.22 \\ \hline
Manufacturer & 10.46 & 10.75 & 332.33 & 323.48 \\ \hline
Navigation & 2.57 & 2.34 & 104.53 & 82.99 \\ \hline
NumbersRace & 2.26 & 1.33 & 1012.79 & 876.50 \\ \hline
Othello & 8.18 & 7.77 & 328.30 & 333.46 \\ \hline
Pylos & 4.78 & 4.84 & 229.06 & 206.96 \\ \hline
Quarto & 2.96 & 2.89 & 226.28 & 219.62 \\ \hline
Racetrack & 1.60 & 1.46 & 85.47 & 82.78 \\ \hline
Sailing Wind & 2.23 & 2.15 & 185.44 & 169.06 \\ \hline
Saving & 1.48 & 1.37 & 249.19 & 246.40 \\ \hline
Skills Teaching & 3.95 & 4.08 & 262.31 & 218.11 \\ \hline
SysAdmin  & 1.90 & 1.81 & 173.24 & 156.65 \\ \hline
Tamarisk & 2.94 & 2.87 & 145.57 & 134.80 \\ \hline
Traffic & 3.94 & 3.81 & 171.77 & 167.41\\ \hline
Triangle Tireworld & 4.43 & 3.85 & 143.73 & 125.15 \\ \hline
Tic Tac Toe & 1.06 & 0.98 & 54.46 & 47.35 \\ \hline

\end{tabular}
}
\end{table}

\subsection{Proof of state-abstraction theorem}
\label{sec:proof}
Here, we will prove the following theorem from the main section:

\textbf{Theorem:} Assume $s_1,s_2$ are two states with $n$ and $l$ actions respectively. Furthermore, assume that each of $s_1$'s and $s_2$'s actions is assigned to an abstract Q node from a pool of $m$ abstract Q nodes with uniform probability. The probability $p_{\text{abs}}$ of $s_1$ and $s_2$ being abstracted according to the ASAP framework can be exactly denoted and then upper bounded by
\begin{equation}
    p_{\text{abs}} = \frac{\sum\limits_{k=1}^{c \coloneqq \min \{n,l,m\}}
    \binom{m}{k} f(n,k) f(l,k)
    }{m^{n+l}} \leq \left(\frac{2c}{m}\right)^{n+l}
\end{equation}
where $f(n,k)$ is the number of surjections from a set of $n$ elements to a set of $k \leq n$ elements.

\textbf{Proof:} Let us denote $s_1$'s actions by $A = \{a_1,\dots,a_n\}$ and $s_2$'s actions by $B = \{b_1,\dots,b_l\}$. The set of abstract nodes is denoted as $\mathbb{A}_1,\dots,\mathbb{A}_m$.
We denote the abstraction that has uniformly been assigned to an action $c \in \{a_1,\dots,a_n,b_1,\dots,b_l\}$ by $\text{abs}(c)$. Using the ASAP framework definition, it holds that
\begin{equation}
    p_{\text{abs}} = \mathbb{P}[ \{\text{abs}(a_i)\: | \:  1 \leq i \leq n\} = \{\text{abs}(b_i)\:|\:1 \leq i \leq l\}].
\end{equation}
Since by assumption the abstraction assignment is uniform, $p_{\text{abs}}$ can be denoted as the ratio of abstraction assignments for $s_1$ and $s_2$ that result in the same set of abstract nodes divided by all possible abstraction assignments. Furthermore, assignments that result in the same set of abstract nodes can be split by the size of that abstract node set. Hence, 
\begin{align}
    p_{\text{abs}} &= \frac{\sum\limits_{k=1}^{c \coloneqq \min(n, l, m)} 
    \left| \left\{ 
        f\colon A \mapsto X,\ g\colon B \mapsto X\ \middle|\ 
        f,g \text{ surjective, } X \subseteq \mathbb{A},\ |X| = k 
    \right\} \right|}{m^{n+l}} \notag \\
    &= \frac{\sum\limits_{k=1}^{c}\binom{m}{k}f(n,k)f(l,k)}{m^{n+l}}
\end{align}
where $\mathbb{A} = \{\mathbb{A}_1,\dots,\mathbb{A}_m\}$. This proves the first part of this theorem. Next, using that $f(n,k) \leq k^n$ yields
\begin{equation}
    \sum\limits_{k=1}^{c}\binom{m}{k}f(n,k)f(l,k) \leq \sum\limits_{k=1}^{c}\binom{m}{k}k^{n+l} \leq c^{n+l}\sum\limits_{k=1}^{c}\binom{m}{k} \leq c^{n+l} 2^c \leq (2c)^{n+l}
\end{equation}
from which the theorem directly follows. \qed

\subsection{Number of state abstractions built by OGA and IPA}

\begin{table}[H]
    \centering
    \caption{Comparison of abstraction statistics for different state abstractions and models to show that OGA almost never finds any state abstractions in contrast to our method IPA. Each column denotes the measured ratio of size one state abstractions to the number of total abstractions (excluding trivial abstractions, i.e. those that group all terminal states or size-one abstractions that did not yet receive an update). Hence, the value $1.00$ corresponds to no non-trivial state abstractions, while a value close to $0$ means that almost all states are grouped into node abstract node. The states whose statistics were averaged come from the state distribution of standard OGA-UCT (see Section \ref{sec:experiment_setup}).
    For IPA-UCT, we used $\lambda_{\text{p}} = 0$. The results were averaged from 100 episodes each. The epsilon values $\varepsilon_{\text{t}}$ denote the transition function threshold defined in Section \ref{sec:foundations}.}
    \begin{minipage}[t]{0.48\linewidth}
    \centering
    \scalebox{1}{
      \setlength{\tabcolsep}{1mm}
      \begin{tabular}{c|cc|cc}
        \toprule
        \multirow{2}{*}{Domain} & \multicolumn{2}{c|}{$\varepsilon_{\text{t}}=0$} & \multicolumn{2}{c}{$\varepsilon_{\text{t}}=0.4$}\\
                                & OGA & IPA & OGA & IPA \\
        \midrule
        Academic Advising   & 1.00 & 1.00 & 1.00 & 0.96 \\
         Crossing Traffic    & 0.50 & 0.55 & 0.50 & 0.55 \\
        Cooperative Recon   & 1.00 & 0.88 & 0.99 & 0.82 \\
        Connect4            & 0.99 & 0.95 & 0.99 & 0.95 \\
        Constrictor         & 0.99 & 0.99 & 0.99 & 0.99 \\
         Earth Observation   & 1.00 & 1.00 & 0.94 & 0.92 \\
        Game of Life        & 1.00 & 1.00 & 0.96 & 0.91 \\
        Manufacturer        & 1.00 & 1.00 & 1.00 & 0.93 \\
        Navigation          & 1.00 & 0.95 & 1.00 & 0.90 \\
         NumbersRace         & 1.00 & 0.91 & 1.00 & 0.91 \\
    
        \bottomrule
      \end{tabular}
    }
    \end{minipage}\hfill
    \begin{minipage}[t]{0.48\linewidth}
    \centering
    \scalebox{1}{
      \setlength{\tabcolsep}{1mm}
      \begin{tabular}{c|cc|cc}
        \toprule
        \multirow{2}{*}{Domain} & \multicolumn{2}{c|}{$\varepsilon_{\text{t}}=0$} & \multicolumn{2}{c}{$\varepsilon_{\text{t}}=0.4$}\\
                                & OGA & IPA & OGA & IPA \\
        \midrule
         Othello             & 0.98 & 0.98 & 0.98 & 0.98 \\
        Pylos               & 0.97 & 0.96 & 0.97 & 0.96 \\
     Quarto              & 0.98 & 0.95 & 0.98 & 0.95 \\
        Racetrack           & 1.00 & 0.93 & 1.00 & 0.93 \\
         Sailing Wind        & 1.00 & 0.99 & 1.00 & 0.97 \\
        Skills Teaching     & 0.79 & 0.80 & 0.70 & 0.73 \\
        SysAdmin            & 1.00 & 0.99 & 1.00 & 0.94 \\
        Tamarisk            & 1.00 & 1.00 & 1.00 & 0.97 \\
        Traffic             & 1.00 & 1.00 & 1.00 & 1.00 \\
        Triangle Tireworld  & 0.99 & 0.93 & 0.99 & 0.92 \\
        \bottomrule
      \end{tabular}
    }
    \end{minipage}
    \label{tab:abs_rates}
\end{table}

\subsection{Ratio of value equivalences}
\begin{table}[H]
    \caption{A list of the ratios of local state pairs and action pairs that are value-equivalent to explain why IPA did not improve the performance in Triangle Tireworld or Sailing Wind because these environments contain very few value-equivalent states. These ratios were determined by randomly sampling $10^5$ states. We then applied $i \in \{1,2,3\}$ random actions to each state and a copy of each state. We then counted how many times out of these $10^5$ states, the resulting states after applying $i$ actions, had the same value or the same Q-value for the $i$-th action. We denote these ratios by $V_{\text{abs}}(i)$ and $Q_{\text{abs}}(i)$. Hence, a ratio of $1.00$ would mean that all states in a search tree layer have the same optimal value, while a ratio of $0.00$ means that no two states have the same $V^*$ value.}
    \label{tab:qtable}

   \centering
    \scalebox{0.8}{
    \begin{tabular}{lcccccc}
        \toprule
        Model & $V_{\text{abs}}(0)$ & $Q_{\text{abs}}(0)$ & $V_{\text{abs}}(1)$ & $Q_{\text{abs}}(1)$ & $V_{\text{abs}}(2)$ & $Q_{\text{abs}}(2)$ \\
        \midrule

Crossing Traffic & 0.83 & 0.89 & 0.84 & 0.88 & 0.85 & 0.88  \\

Navigation &  0.05 & 0.05 & 0.05 & 0.13 & 0.05 & 0.12   \\

Racetrack  & 0.38 & 0.53 & 0.37 & 0.42 & 0.34 & 0.37 \\

Sailing Wind &  0.04 & 0.01 & 0.03 & 0.01 & 0.03 & 0.01   \\

Skill Teaching &  0.29 & 0.11 & 0.17 & 0.11 & 0.06 & 0.04 \\

Triangle Tireworld & 0.03 & 0.67 & 0.03 & 0.61 & 0.03 & 0.59   \\

        \bottomrule
    \end{tabular}
    }
    
\end{table}

\subsection{Performances of Alternative Pruning Methods}
\label{subsec:ipa:alternatives_data}
In Section~\ref{sec:experiments} it was mentioned that preliminary experiments on alternative pruning methods were conducted which showed most promise for IPA-UCT compared to TOPN-UCT and CONF-UCT. These results are shown here. Concretely, the following Table~\ref{tab:ipa:alternatives} lists the parameter-optimized performances of IPA-UCT, TOPN-UCT, and CONF-UCT for 500 iterations using the following parameters.
For $(\varepsilon_{\text{a}},\varepsilon_{\text{t}})$-OGA, the environment specific $\varepsilon_{\text{a}}$ values (see Table~\ref{tab:epsa_values}) that include $\{0,\infty\}$ and $\varepsilon_{\text{t}} \in \{0,0.2,0.4,0.8\}$ were tested. For pruned OGA $\alpha \in \{0,0.1,0.2,0.5,0.75,1.0\}$ was tested. The table lists the best performances out of these two techniques. If IPA-UCT is used, then $\lambda_{\text{p}} \in \{0,0.25,0.5,1,2,4,\infty\}$ was varied, for CONF-UCT, we varied $p_{\text{c}} \in \{0.1,0.25,0.5,0.75,0.9\}$, and for TOPN-UCT, $(n_{\text{matches}},n_{\text{min}}) \in \{1,2\} \times \{0,50\}$ was used. All methods were run with $C=2$.

\begin{table}[H]\centering
\caption{The performances various pruning methods as alternatives for the UCB-based pruning in IPA-UCT.}
\label{tab:ipa:alternatives}
\scalebox{1.0}{
\setlength{\tabcolsep}{1mm}\begin{tabular}{ c c c c }
\toprule
&IPA-UCT & CONF-UCT & TOPN-UCT\\
\midrule
Academic Advising & $\boldsymbol { -67.0 \pm 0.9 }$ & $-67.3 \pm 0.9$ & $-68.0 \pm 0.9$\\
Chess & $\boldsymbol { 0.2 \pm 0.0 }$ & $0.1 \pm 0.0$ & $0.1 \pm 0.0$\\
Connect4 & $\boldsymbol { 0.2 \pm 0.0 }$ & $0.2 \pm 0.0$ & $0.1 \pm 0.0$\\
Constrictor & $\boldsymbol { 0.1 \pm 0.0 }$ & $0.1 \pm 0.0$ & $0.1 \pm 0.0$\\
Cooperative Recon & $\boldsymbol { 13.3 \pm 0.3 }$ & $11.2 \pm 0.4$ & $10.7 \pm 0.4$\\
Crossing Traffic & $\boldsymbol { -25.0 \pm 1.2 }$ & $-25.9 \pm 1.3$ & $-25.9 \pm 1.3$\\
Earth Observation & $\boldsymbol { -8.4 \pm 0.2 }$ & $-8.5 \pm 0.3$ & $-8.5 \pm 0.2$\\
Game of Life & $565.1 \pm 2.4$ & $\boldsymbol { 566.0 \pm 2.3 }$ & $565.1 \pm 2.3$\\
Manufacturer & $-1214.4 \pm 12.9$ & $\boldsymbol { -1212.8 \pm 12.9 }$ & $-1303.0 \pm 14.4$\\
Navigation & $-15.9 \pm 0.4$ & $\boldsymbol { -15.9 \pm 0.4 }$ & $-17.5 \pm 0.5$\\
NumbersRace & $\boldsymbol { 0.2 \pm 0.0 }$ & $-0.0 \pm 0.0$ & $0.1 \pm 0.0$\\
Othello & $0.2 \pm 0.0$ & $\boldsymbol { 0.2 \pm 0.0 }$ & $0.2 \pm 0.0$\\
Pylos & $0.2 \pm 0.0$ & $\boldsymbol { 0.2 \pm 0.0 }$ & $0.2 \pm 0.0$\\
Quarto & $\boldsymbol { 0.2 \pm 0.0 }$ & $0.1 \pm 0.0$ & $0.2 \pm 0.0$\\
Racetrack & $-8.7 \pm 0.0$ & $-8.7 \pm 0.0$ & $\boldsymbol { -8.6 \pm 0.0 }$\\
Sailing Wind & $\boldsymbol { -64.0 \pm 1.2 }$ & $-64.3 \pm 1.3$ & $-64.5 \pm 1.3$\\
Saving & $\boldsymbol { 49.4 \pm 0.2 }$ & $49.4 \pm 0.2$ & $49.3 \pm 0.2$\\
Skills Teaching & $\boldsymbol { 65.4 \pm 7.5 }$ & $63.7 \pm 7.6$ & $31.9 \pm 7.5$\\
SysAdmin & $\boldsymbol { 396.2 \pm 2.0 }$ & $395.8 \pm 2.0$ & $395.4 \pm 2.0$\\
Tamarisk & $\boldsymbol { -562.9 \pm 8.5 }$ & $-573.2 \pm 8.6$ & $-604.8 \pm 8.8$\\
TicTacToe & $0.0 \pm 0.0$ & $0.0 \pm 0.0$ & $\boldsymbol { 0.0 \pm 0.0 }$\\
Traffic & $\boldsymbol { -15.1 \pm 0.3 }$ & $-15.1 \pm 0.3$ & $-15.2 \pm 0.3$\\
Triangle Tireworld & $80.5 \pm 1.2$ & $\boldsymbol { 80.8 \pm 1.2 }$ & $80.6 \pm 1.1$\\
\bottomrule
\end{tabular}}
\end{table}

\subsection[How IPA-UCT transforms $J_{\text{UCB}}$ into an equivalence relation]{How IPA-UCT transforms \bm{$J_{\text{UCB}}$} into an equivalence relation}
\label{sec:vea_jhat_to_equiv}Since $J_{\text{UCB}}$ does not induce an equivalence relation, we cannot simply place any two states $s_1,s_2$ such that $s_1 \sim_{J_{\text{UCB}}} s_2$ into the same abstract node. We will handle this similarly to how the epsilon greater than zero case for the state-action-pairs (i.e. $(\varepsilon_{\text{a}},\varepsilon_{\text{t}})$-OGA) case was handled. Again, the aim of the following heuristic is to produce an equivalence relation for states whilst creating as big and stable as possible abstract nodes. The subsequent technique corresponds to the method \textsc{compute\_state\_abstraction} in the Pseudocode~\ref{alg:ipa:pseudocode}.

Each abstract state node now also keeps track of its representative, which is one of its original nodes. Furthermore, it is assigned a unique and constant ID at its creation. At its creation, an abstract node is assigned an ID equal to the total number of abstract state nodes that have been created so far. Whenever a state node $s$ with abstract node $\mathcal{N}$ is updated, the set $J_{\text{UCB}}(s)$ is updated. Then, if either $s$ is the representative of $\mathcal{N}$ or if the equations \ref{eq:vea1}, \ref{eq:vea2} do not hold (with respect to the representative of $\mathcal{N}$), then the abstract node of $s$ is updated by choosing the largest abstract node (tie breaks by using the ID) with a representative $s^{\prime}$ such that $s \sim_{J_{\text{UCB}}} s^{\prime}$. In case this leads to a different abstract node than the current one of $s$, a new representative for the old abstract node is chosen at random. In the case $s^{\prime} = s$, then the equations \ref{eq:vea1}, \ref{eq:vea2} are checked with the old value $J_{\text{UCB}}$ for $s^{\prime}$.

\newpage
\subsection{Monte Carlo Tree Search}
\label{sec:mcts}
All abstraction algorithms presented here rely on Monte Carlo Tree Search (MCTS). In the following we are going to specify the MCTS version used here.
\begin{enumerate}
    \item Since this is a necessary requirement for ASAP and IPA to detect abstractions in the first place, our MCTS version builds a directed acyclic graph, i.e. two state-action pairs have the same successor node if it represents the same MDP state.
    \item The tree policy is the Upper Confidence Bounds (UCB) policy which chooses the action that maximizes the UCB value
\begin{equation}
    \text{UCB}(a) = 
    \underbrace{\frac{V_a}{N_a}}_{\text{Q term}} + 
    \underbrace{\lambda \sqrt{\frac{\log\left(\sum\limits_{a^{\prime} \in \mathbb{A}(s)}N_{a^{\prime}}\right)}{N_a}}}_{\text{Exploration term}}.
\end{equation}
Here, $s$ is the state at which the decision has to be made, $V_a$ is the sum of returns of the action under consideration, and $N_a$ are its visits.
    \item The greedy decision policy is used, i.e. the root action with the maximal Q value is chosen as the final decision.
\end{enumerate}

\subsection{Problem descriptions}
\label{sec:problem_descriptions}

All the problem domains that appeared in this paper are described in \cite{kvda} as well as in \cite{demcts}. 
Furthermore, most environments are parametrizable (e.g., the racetrack choice in Racetrack). The concrete parameter choices used for the experiments can be found in the \textit{ExperimentConfigs} folder of the repository accompanying this paper \cite{repo}.

\subsection{IPA-UCT Pseudocode}

\begin{algorithm}[H]
\scriptsize
\DontPrintSemicolon
\SetKwComment{tcp}{// }{}
\SetKwProg{Fn}{function}{}{}
\SetKwFunction{treePolicy}{treePolicy}
\SetKwFunction{newSingletonQ}{new\_singleton\_Q\_abstraction}
\SetKwFunction{newSingletonS}{new\_singleton\_state\_abstraction}
\SetKwFunction{rollout}{rollout}
\SetKwFunction{backup}{backup}
\SetKwFunction{updateQ}{update\_Q\_abstraction}
\SetKwFunction{computeState}{compute\_state\_abstraction}
\SetKwFunction{chooseRep}{choose\_new\_representative\_randomly}
\SetKwFunction{updateStats}{update\_statistics}
\SetKwFunction{updateASAP}{update\_state\_abstraction}
\SetKwFunction{initTree}{init\_tree}
\SetKwFunction{similar}{similar}
\SetKwFunction{distance}{distance}
\SetKwInOut{Parameters}{Parameters}

\caption{IPA-UCT}
\label{alg:ipa:pseudocode}

\Parameters{$\lambda_{\text{p}}$, \textit{oga\_args}}
\KwIn{\textit{state}}
\textbf{Globals: } $RecencyCount$, $abstractionsQ$, $abstractionsStates$

$max\_id = 0, max\_id\_states = 0$\;
$tree = \initTree(state)$\;

\For{$i = 1$ \KwTo $oga\_args.iterations$}{
  $leaf, path\_to\_leaf, newQnodes = \treePolicy(tree)$ \tcp*{UCB using aggregate abstraction statistics}
  \ForEach{$newQnode \in newQnodes$}{
    $abstractionsQ[newQnode] = \newSingletonQ()$\;

    $abstractionsQ[newQnode].representative = newQnode$\;

    $newQnode.id = max\_id++$\;
  }
  \If{$leaf \notin tree$}{
    $abstractionsStates[leaf] = \newSingletonS()$\;

    \colorbox{blue!7}{
    $leaf.representative = abstractionStates[leaf]$
    }
    
        \colorbox{blue!7}{
    $leaf.id = max\_id\_states++$}
  }
  $rollout\_return = \rollout(leaf)$\;
  \backup($path\_to\_leaf$, $rollout\_return$)\;
  \ForEach{$(Q,s) \in path\_to\_leaf$}{
    \updateQ($Q$)\;

    \blkstart{y}
    \updateASAP($s$)\;
    \blkend{y}
    
  }
}

\Return $\arg\max\limits_{a} \; Q(state,a)$\;

\BlankLine
\Fn{\updateASAP{$state$}}{

\If{$RecencyCount[state]++ < oga\_args.K$}{
    \Return
}

  $RecencyCount[state] = 0$\;
  $new\_abs = \computeState(state)$\;
  \If{$abstractionsStates[state] \neq new\_abs$}{
    \tcp{Transfer original node to new abstraction if needed}

   \blkstart{A}
    \If{$state == abstractionsStates[state].representative$ \textbf{and} $abstractionsStates[state].size > 1$}{
      \chooseRep($abstractionsStates[state]$, excluding $= state$)\;
    }
    \blkend{A}
    
    $abstractionsState[state] = new\_abs$
    \updateQ($state.parents$)\;
  }
}

\BlankLine
\blkstart{computeA}%
    \Fn{\computeState{$state$}}{

        \textit{Update} $J_{\text{UCB}}(s)$
    
        \If{$state$ is fully expanded \textbf{and }($state == abstractionsStates[state].representative$ \textbf{or} $\lnot (state \sim_{J_{\text{UCB}}} abstractionsStates[state].representative)$)  }{

        $\textrm{\textit{targetAbs}} = None$
        
        \ForEach{$absState \in abstractionsState$ with the same layer as $state$}{

        \tcp{Using the old  $J_{\text{UCB}}$ value iff absState.representative $==$ s}
          \If{ \textbf{not} $state \sim_{J_{\text{UCB}}} absState.representative$}{
            \textbf{continue}
          }
          
          \If{ ($\textrm{\textit{targetAbs}} == \textrm{\textit{None}}$ \textbf{or} $\textrm{\textit{targetAbs}}.size < absState.size$ \textbf{or} ($\textrm{\textit{targetAbs}}.size == absState.size$ \textbf{and} $\textrm{\textit{targetAbs}}.representative.id > absState.representative.id$))}{
            $\textrm{\textit{targetAbs}} = absState$\;
          }
        }
        \If{$\textrm{\textit{targetAbs}} == \textrm{\textit{None}}$}{
            \Return \newSingletonS()
        }
        \Else{
            \Return $\textrm{\textit{targetAbs}}$\;
        }
      }\Else{
        \Return $abstractionsState[state]$
      }
}
\blkend{computeA}%
\end{algorithm}

\end{document}